\documentclass{article} 
\usepackage{iclr2026_conference,times}

\usepackage{amsmath,amsfonts,bm}

\def\eqref#1{equation~\ref{#1}}

\def\1{\bm{1}}

\def\vx{{\bm{x}}}

\def\mX{{\bm{X}}}

\DeclareMathAlphabet{\mathsfit}{\encodingdefault}{\sfdefault}{m}{sl}
\SetMathAlphabet{\mathsfit}{bold}{\encodingdefault}{\sfdefault}{bx}{n}

\usepackage{hyperref}
\usepackage{url}

\title{Improving Visual Discriminability of CLIP for Training-Free Open-Vocabulary Semantic Segmentation}

\airxivcopy                   
\author{Jinxin Zhou, Jiachen Jiang \& Zhihui Zhu\thanks{The corresponding author.} \\
Department of Computer Science\\
The Ohio State University\\
\texttt{\{zhou.3820,jiang.2880,zhu.3440\}@osu.edu} \\
}

\usepackage{enumitem}
\usepackage{graphicx}  
\usepackage{subfig}    
\usepackage{cleveref}
\usepackage{booktabs}   
\usepackage{multirow}   
\usepackage{xcolor,colortbl} 
\usepackage{pifont}     
\usepackage{caption}    
\usepackage[table]{xcolor}   
\usepackage{bbm}
\newcommand{\graycell}[1]{\cellcolor{gray!20}{#1}}
\newcommand{\green}[1]{\textcolor{green!60!black}{#1}}

\newcommand{\revise}[1]{{\color{black}{#1}}}

\usepackage{tikz,xcolor}

\newcommand{\graybox}[1]{\colorbox{gray!20}{\textcolor{black}{#1}}}
\begin{document}

\maketitle

\begin{abstract}

Extending CLIP models to semantic segmentation remains challenging due to the misalignment between their image-level pre-training objectives and the pixel-level visual understanding required for dense prediction. While prior efforts have achieved encouraging results by reorganizing the final layer and features, they often inherit the global alignment bias of preceding layers, leading to suboptimal segmentation performance. In this work, we propose LHT-CLIP, a novel training-free framework that systematically exploits the visual discriminability of CLIP across \emph{layer}, \emph{head}, and \emph{token} levels. Through comprehensive analysis, we reveal three key insights: (i) the final layers primarily strengthen image–text alignment with sacrifice of visual discriminability (e.g., last 3 layers in ViT-B/16 and 8 layers in ViT-L/14), partly due to the emergence of anomalous tokens; (ii) a subset of attention heads (e.g., 10 out of 144 in ViT-B/16) display consistently strong visual discriminability across datasets; (iii) abnormal tokens display sparse and consistent activation pattern compared to normal tokens. Based on these findings, we propose three complementary techniques: semantic-spatial reweighting, selective head enhancement, and abnormal token replacement to effectively restore visual discriminability and improve segmentation performance without any additional training, auxiliary pre-trained networks, or extensive hyperparameter tuning. Extensive experiments on 8 common semantic segmentation benchmarks demonstrate that LHT-CLIP achieves state-of-the-art performance across diverse scenarios, highlighting its effectiveness and practicality for real-world deployment.



\end{abstract}

\section{Introduction}\label{sec:intro}
Recent advances in vision-language pretrained models, such as CLIP~\cite{radford2021learning}, have demonstrated remarkable generalization and open-vocabulary recognition capabilities at the image level, thereby opening up possibilities for transferring image-text alignment to pixel-level tasks. Despite this progress, they often underperform in dense prediction tasks like semantic segmentation, primarily due to their limited capacity to localize fine-grained visual details~\cite{rao2022denseclip, wang2024sclip}. To address these limitations, several studies have incorporated trainable modules into CLIP, typically relying on additional forms of supervision such as dense annotations for a restricted set of categories~\cite{xu2022simple, xu2022groupvit, xing2023rewrite, li2024mask} or supplementary image-text pairs~\cite{cha2023learning, luo2023segclip, ren2023viewco, xu2023learning, zhang2023uncovering}. Although these approaches have demonstrated improved segmentation performance, they incur significant computational and annotation costs. Furthermore, the dependence on limited supervision undermines the generalizability of the model, making it prone to overfitting the training distribution. 

These challenges have sparked increasing interest in training-free methods\cite{wang2024sclip, li2023clip, zhou2022extract, lan2024clearclip, hajimiri2025pay, lan2024proxyclip, shao2024explore, bousselham2024grounding, yang2024resclip}, which aim to adapt CLIP’s pre-trained representations for semantic segmentation without additional training, while preserving its generalization capability. A key difficulty in this direction is enhancing \revise{visual representations} for accurate pixel-level predictions. For instance, MaskCLIP\cite{zhou2022extract} computes similarity between key features in the final attention layer to enrich patch embeddings. SCLIP~\cite{wang2024sclip} replaces the standard query-key attention with correlative self-attention (query-query and key-key). ClearCLIP~\cite{lan2024clearclip} further removes residual connections and discards the FFN in the final layer to reduce noise and improve spatial alignment. ResCLIP~\cite{yang2024resclip} incorporates attention maps from earlier layers to refine final-layer attention map. However, these methods largely focus on modifying the final-layer attention, often leading to suboptimal ambiguous local relationships and noisy segmentation. To address spatial limitations, some approaches incorporate features from auxiliary backbones such as DINO~\cite{wysoczanska2024clip, lan2024proxyclip}, SAM~\cite{lan2024proxyclip, zhang2024corrclip}, or diffusion models~\cite{Corradini2024FreeSegDiffTO, sun2024cliper}. While effective, these methods incur significant computational and memory overhead.

\revise{Motivated by these limitations, we begin with a layer-wise analysis of visual discriminability and text-semantic alignment within the CLIP vision encoder (see~\Cref{subsec:analysis-SDSA} for detailed definitions). As shown in~\Cref{fig:layer-auc}, the final layers exhibit a clear trade-off: visual discriminability drops sharply while semantic alignment improves only marginally. To understand the cause of this phenomenon, we further examine internal token interactions and structural patterns across layers. Attention map visualizations reveal that abnormal tokens emerge in deeper layers, attracting disproportionately high attention from nearly all spatial positions. This behavior causes the majority of tokens to converge on a small subset, thus disrupting the spatial coherence. Further analysis reveals that these abnormal tokens have sparse, high-magnitude activations that remain consistent across positions, layers, and samples. Complementary to prior assumptions that such tokens primally encode global semantic content, our findings suggest they act more as bias components offsetting global mean features, thereby facilitating alignment with text embeddings.

Based on the analysis, we propose LHT-CLIP, a training-free framework that leverages the inherent properties of CLIP to enhance the visual discriminability while preserving semantic alignment. LHT-CLIP comprises three complementary strategies: abnormal token replacement (ATR), spatial-semantic reweighting (SSR), and selective head enhancement (SHE). Specifically, ATR identifies abnormal tokens via sparsity thresholding and replaces them with neighboring tokens. SSR mitigates the degradation of visual discriminability in the final layers by upweighting residual pathways, thereby restoring balance between spatial coherence and semantic alignment. Finally, SHE further enhances visual discriminability by selectively aggregating features from high-discriminability attention heads, using them as soft pseudo-masks to refine output features. Experimental results show that LHT-CLIP consistently improves performance when integrated into diverse baselines, achieving new state-of-the-art results on eight benchmark datasets.

}
\textbf{Contributions.} Our contributions can be summarized as follows:
\begin{itemize}[leftmargin=0.13in,topsep=0.2em,itemsep=0.11em]
    \item We conduct a throughout analysis of visual discriminability at the token, head, and layer levels.
    \item We propose, a novel training-free approach, terms LHT-CLIP. To the best of our knowledge, this is the first work to explicitly modify the inference procedure prior to the final layer, enabling improved spatial coherence without compromising semantic alignment. 
    \item The extensive experiment results on open-vocabulary semantic segmentation tasks consistently demonstrate the effectiveness of the proposed method. 
\end{itemize}

\section{Analysis of Visual Discriminability and Semantic Alignment}\label{sec:analysis}
\subsection{Preliminaries}
CLIP employs a Vision Transformer (ViT)~\cite{Dosovitskiy2020AnII} as its image encoder to generate visual representations that are aligned with corresponding textual descriptions. The vision encoder first tokenizes an input image of size $H \times W \times 3$ by dividing it into a grid of non-overlapping patches of size $P \times P$, yielding $h = H / P$ rows and $w = W / P$ columns of patches. Each patch is then linearly projected into a $D$-dimensional embedding space, $\vx_i \in \mathbb{R}^D$, and augmented with positional embeddings. An additional learnable \texttt{[CLS]} token is prepended to the sequence and is later used for image-level prediction. The resulting token sequence is denoted as $\mX^0 = {[\vx^0_\text{cls}, \vx^0_1, \ldots, \vx^0_{hw}]} \in \mathbb{R}^{(1+hw) \times D}$. This sequence is passed through a stack of $L$ Transformer encoder layers, each consisting of a multi-head self-attention (MSA) module followed by a feed-forward network (FFN). Let LN$(\cdot)$ denotes layer normalization, the token representations are updated at each layer $l$ as follow:
\begin{align}
    \hat{\mX}^{l}&=\mX^{l-1}+\text{MSA}(\text{LN}(\mX^{l-1})),\\
    \mX^{l}&=\hat{\mX}^{l}+\text{FFN}(\text{LN}(\hat{\mX}^{l})).
\end{align}
The CLIP model is originally trained on large-scale image–text pairs for open-vocabulary image recognition tasks. To extend it to semantic segmentation, a natural approach is to compute the similarity between the visual tokens $\mX^L = [\vx^L_1, \ldots, \vx^L_{hw}]$ from the final Transformer layer and the textual embeddings of $C$ category names, denoted by $\mathbf{t} \in \mathbb{R}^{C \times D}$. This results in a patch-text similarity map of size $hw \times C$. Denote $\mathbf{t}_c$ as the embedding of the $c$-th class name, the final segmentation prediction is obtained by applying an \texttt{argmax} operation over the class dimension of this similarity map, as follows:
\begin{align}\label{eq:seg}
    \hat c(\vx_i) = \arg\max_c \frac{\langle\vx_{i}^L, \mathbf{t}_c\rangle}{\|\vx_{i}^L\| \cdot\|\mathbf{t}_{c}\|},
\end{align}

Ideally, for effective semantic segmentation, the vision encoder should produce feature representations that satisfy two key properties:
\begin{itemize}[leftmargin=0.13in,topsep=0.2em,itemsep=0.11em]
    \item \textbf{Visual discriminability}: token features should exhibit high internal consistency within the same semantic category while remaining clearly distinguishable from those of other categories, thereby enabling accurate and clean segmentation results.
    \item \textbf{Semantic alignment}: token features should be well-aligned with their corresponding textual embeddings to enable semantically meaningful segmentation results.
\end{itemize}
\revise{Beyond their importance in open-vocabulary semantic segmentation, these two properties are also highly relevant to the development of multimodal large language models. To preserve strong generalization capability, the vision encoder of CLIP is often directly employed to extract visual representations without additional training, which are then used as inputs to downstream language models, as exemplified by LLaVA~\cite{liu2023visual, liu2024improved}. }

\subsection{Analysis of visual discriminability and semantic alignment}\label{subsec:analysis-SDSA}
\begin{figure*}[t]
    \centering
    
    \subfloat[VOC (ViT-B)]{\includegraphics[width=0.24\textwidth]{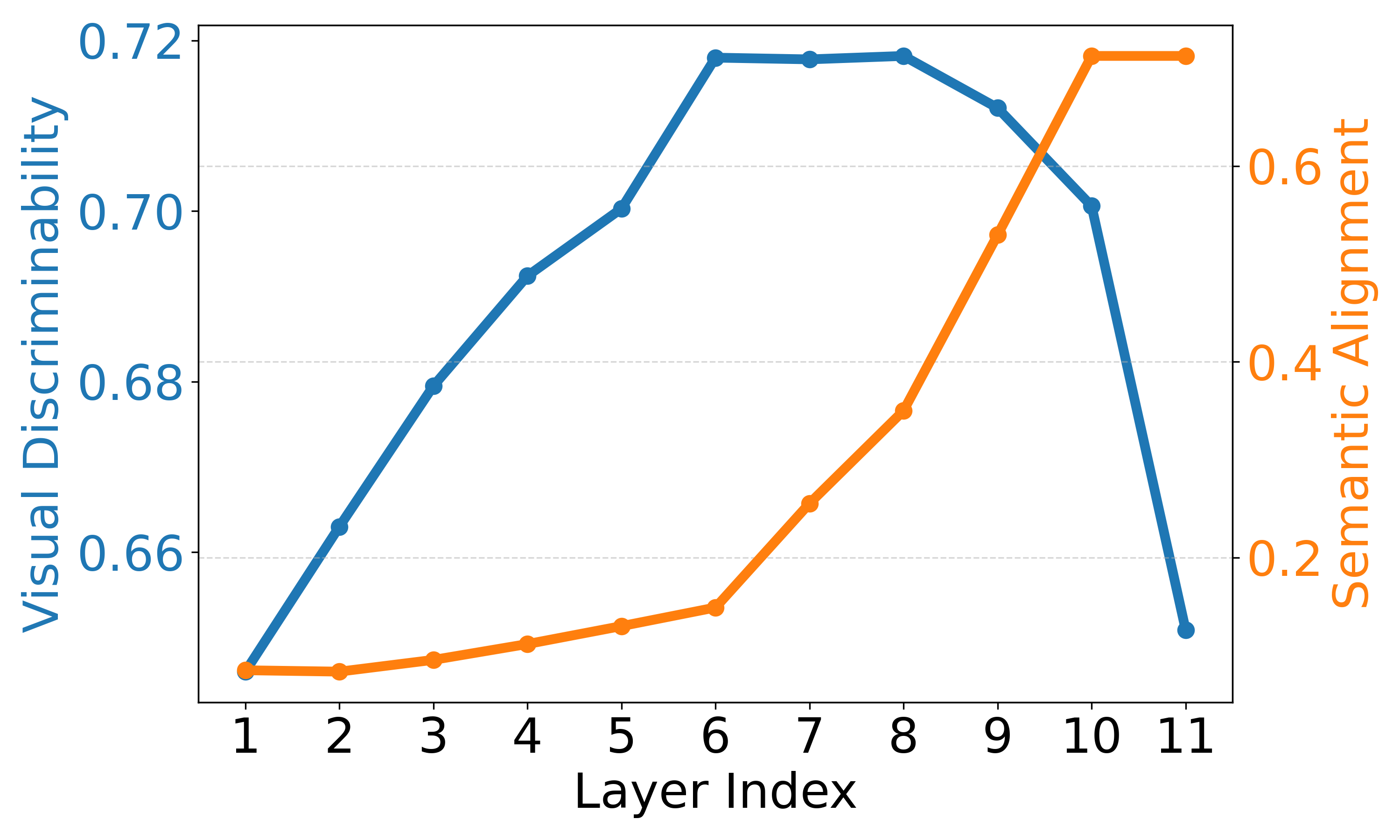}}\ 
    \subfloat[Context (ViT-B)]{\includegraphics[width=0.24\textwidth]{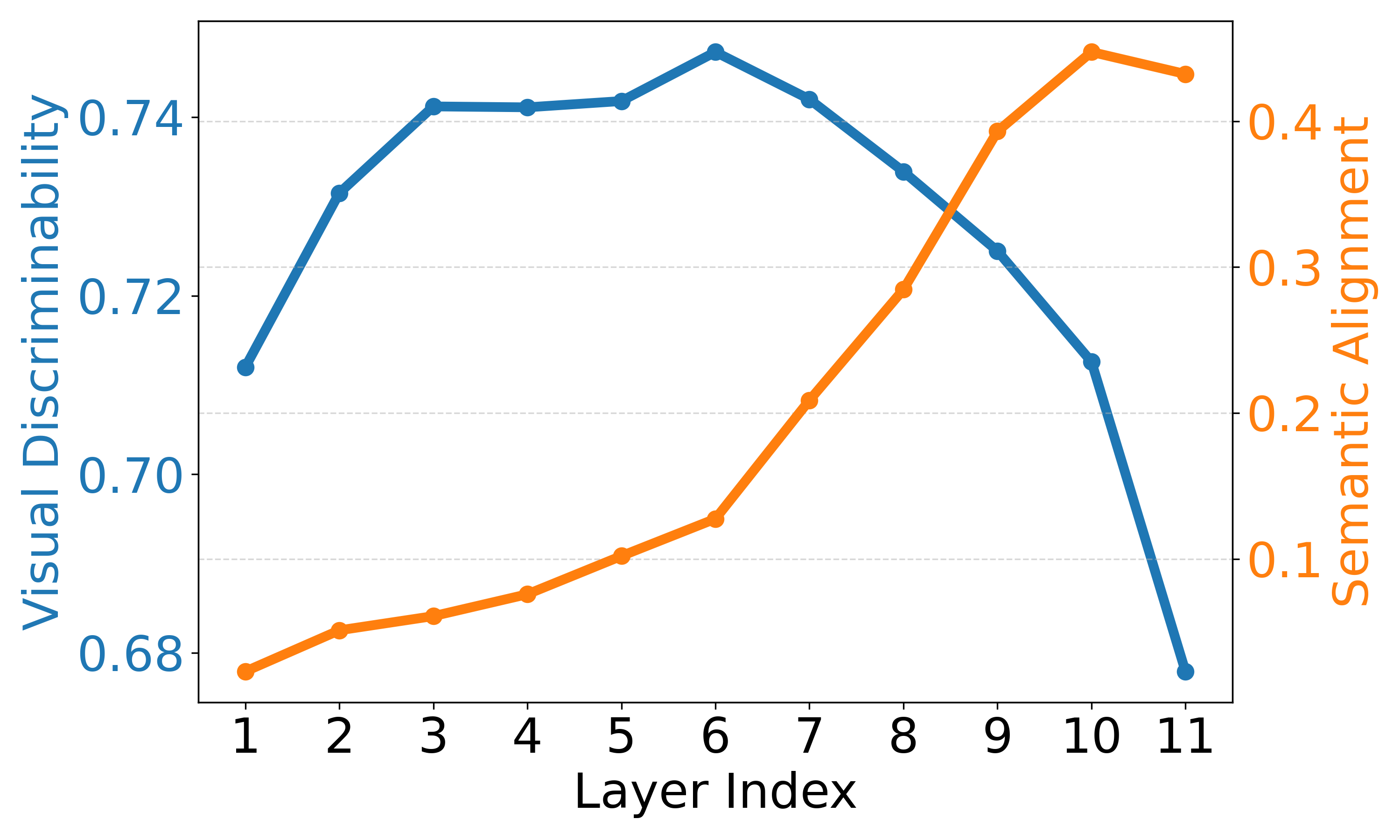}}\ 
    \subfloat[ADE (ViT-B)]{\includegraphics[width=0.24\textwidth]{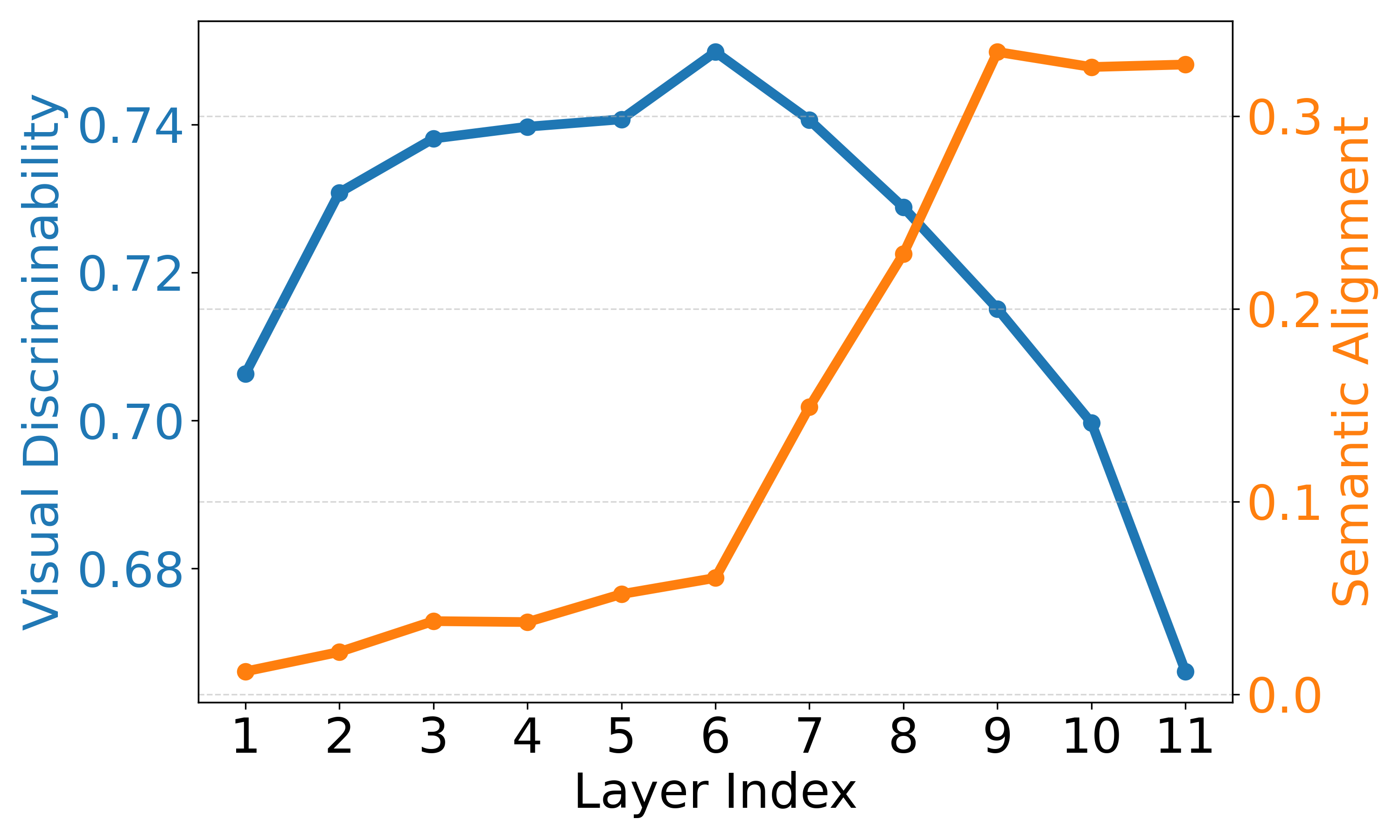}}
    \subfloat[COCO-Stuff (ViT-B)]{\includegraphics[width=0.24\textwidth]{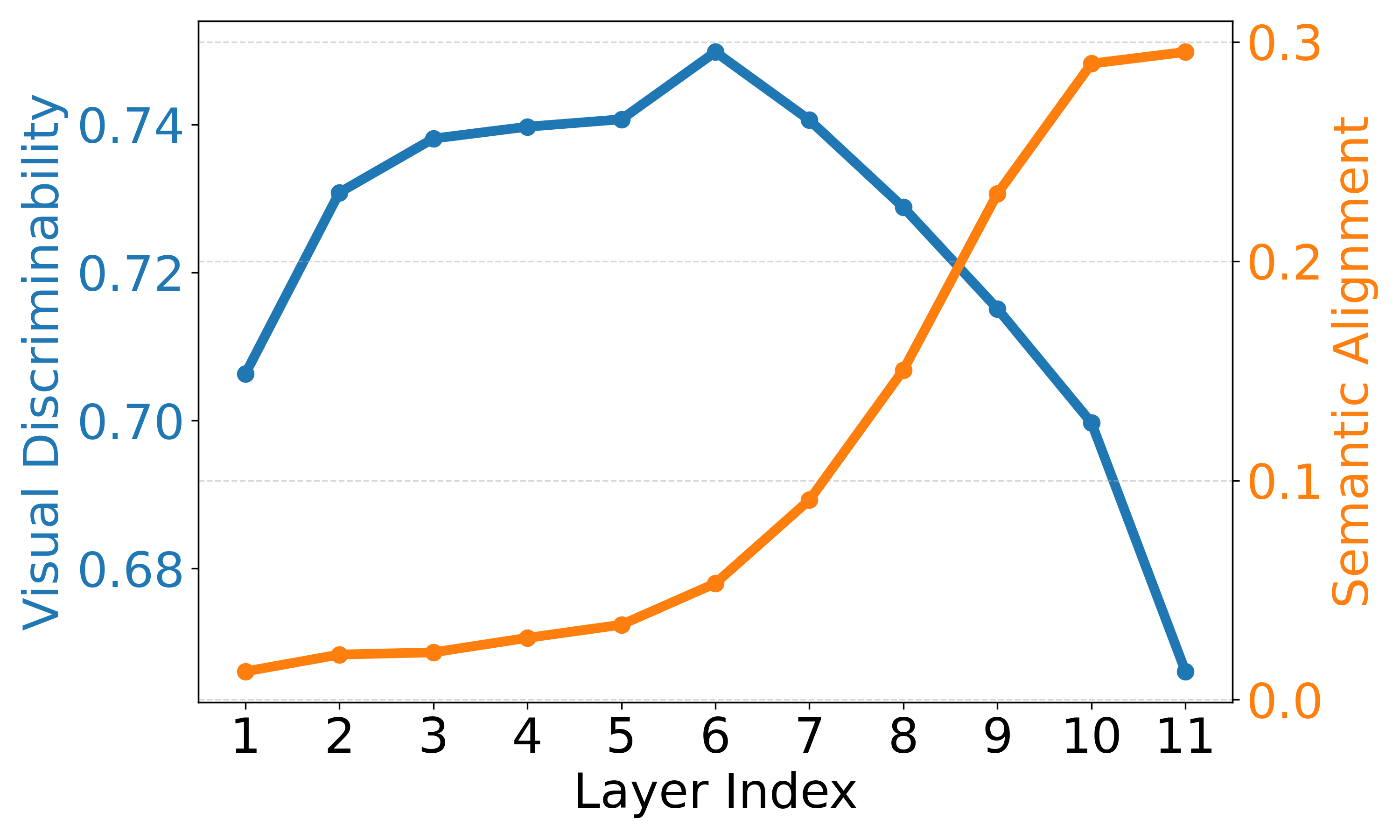}}\\

    \subfloat[VOC (ViT-L)]{\includegraphics[width=0.24\textwidth]{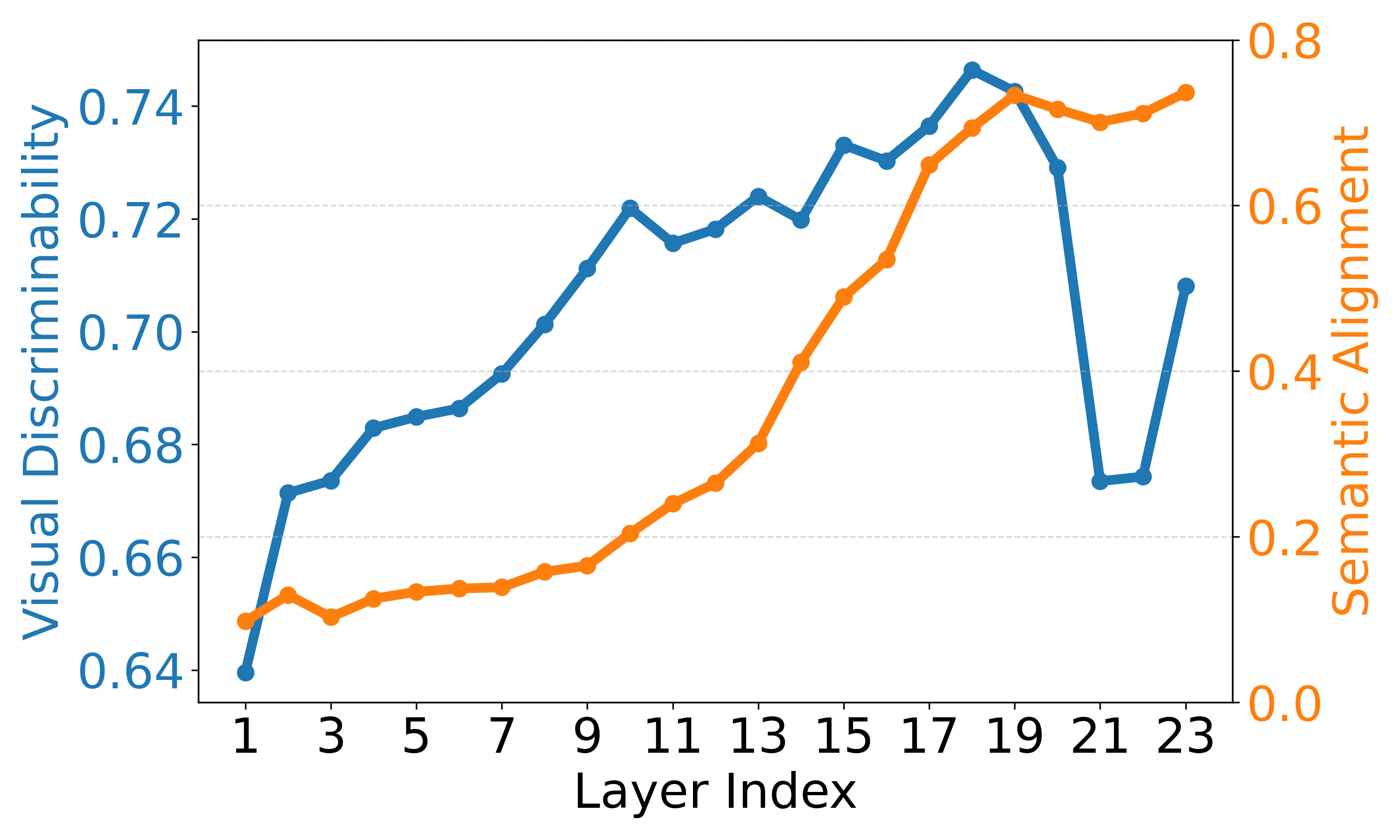}}\ 
    \subfloat[Context (ViT-L)]{\includegraphics[width=0.24\textwidth]{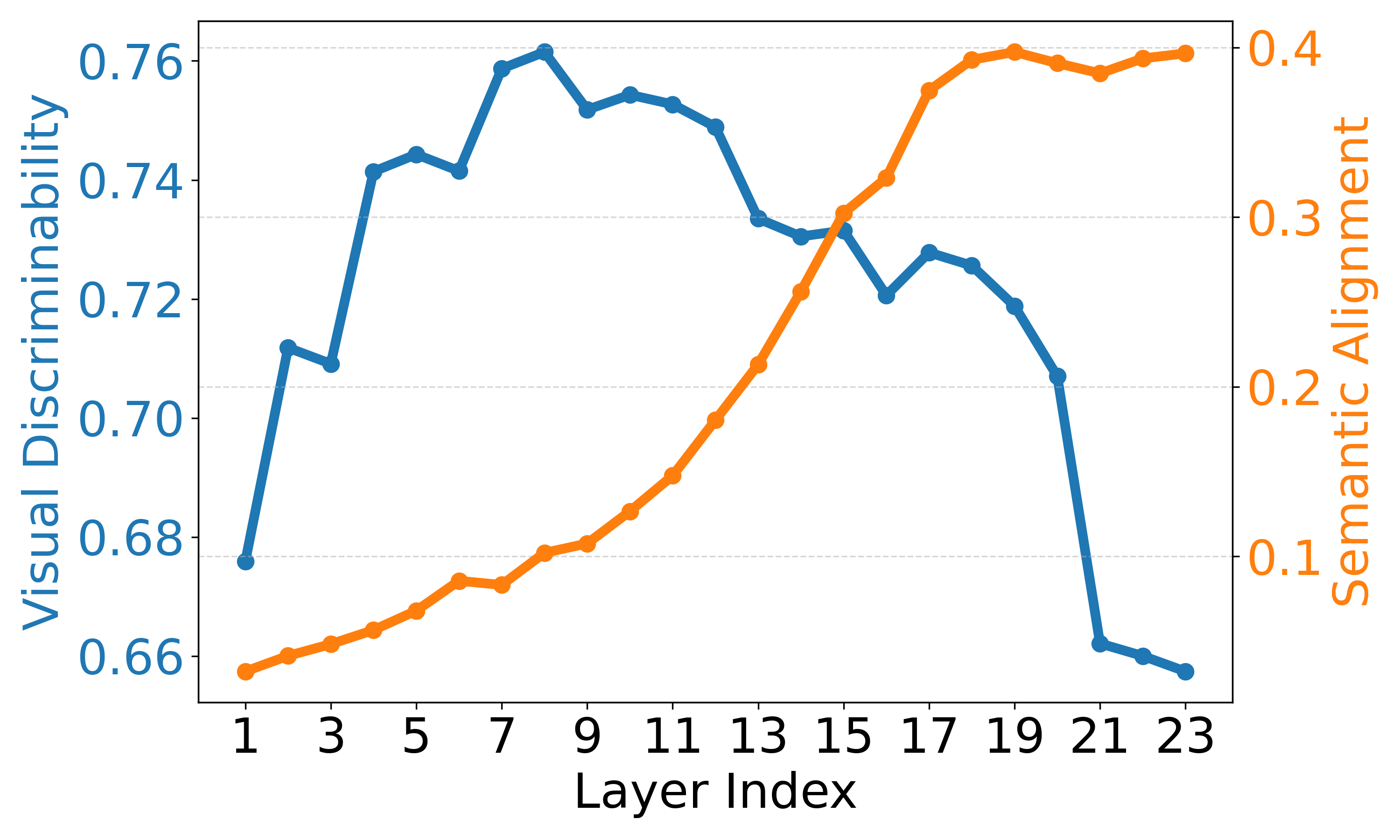}}\ 
    \subfloat[ADE (ViT-L)]{\includegraphics[width=0.24\textwidth]{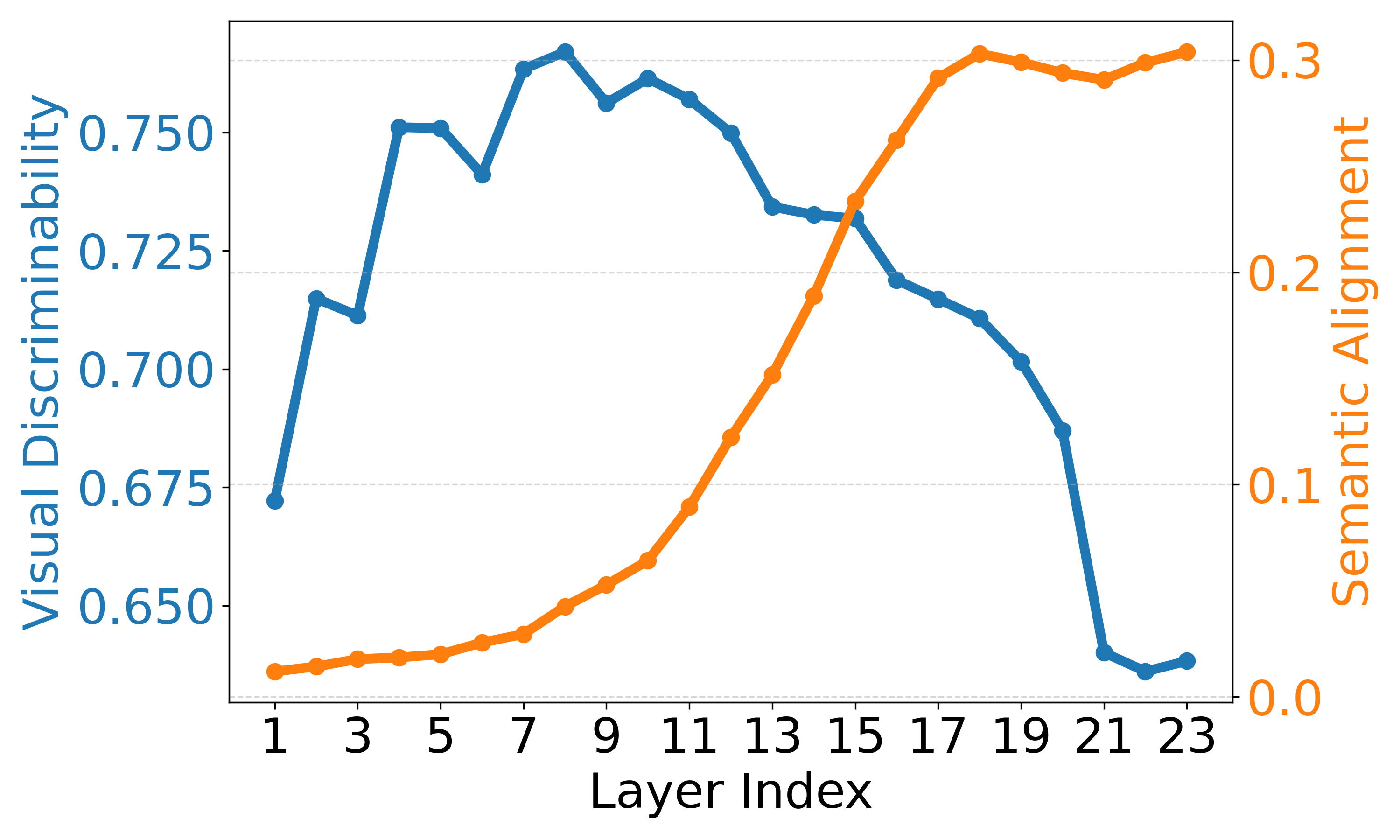}}
    \subfloat[COCO-Stuff (ViT-L)]{\includegraphics[width=0.24\textwidth]{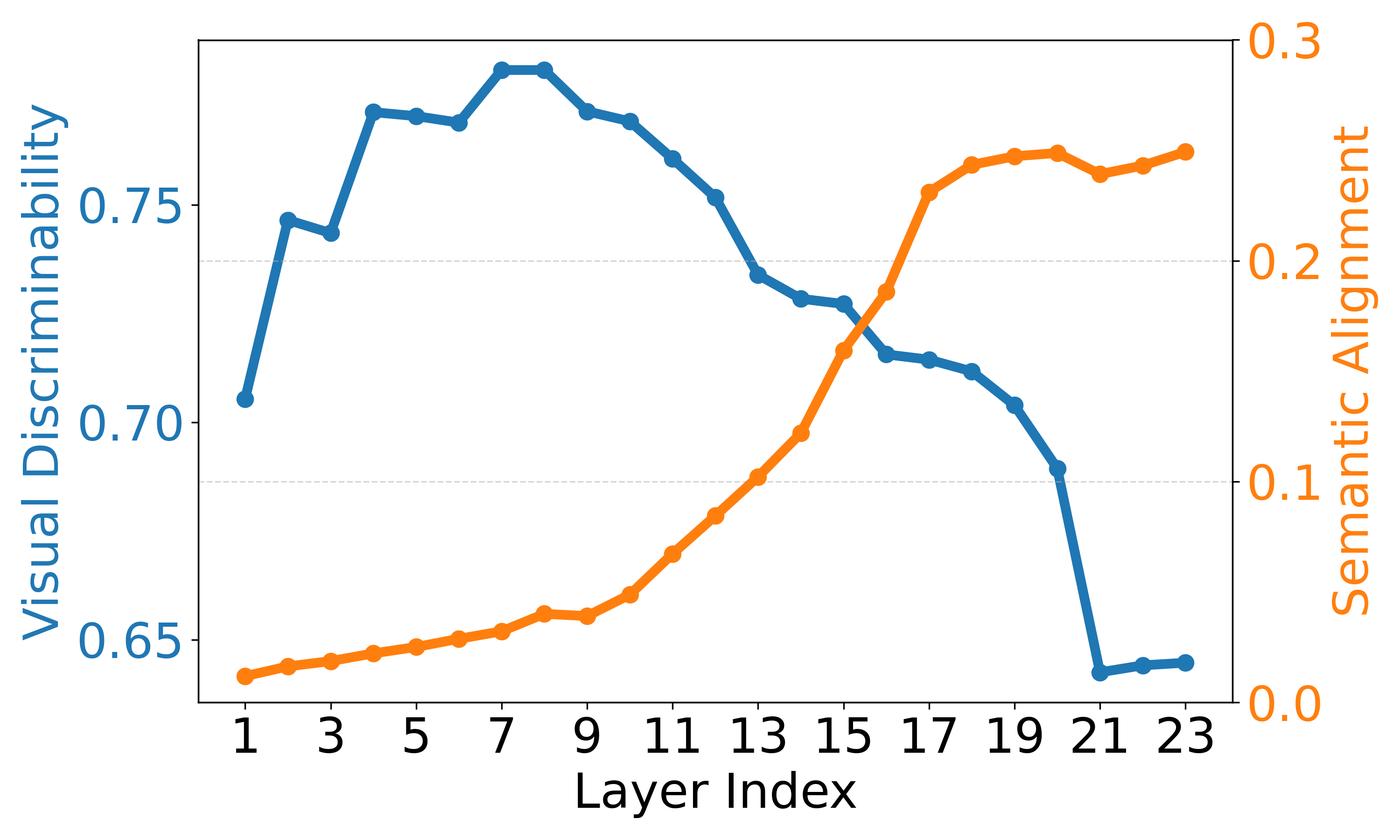}}
    
    \caption{Layer-wise analysis of visual discriminability (blue) and semantic alignment (orange) within the CLIP vision encoders across different datasets. The final layer is excluded from the analysis to avoid discrepancies caused by prior modifications to the last-layer in different methods.}
    \label{fig:layer-auc}
\end{figure*}

\textbf{Measures of visual discriminability and semantic alignment.} 
\revise{To quantify \textbf{visual discriminability}, we adopt the evaluation protocol proposed by~\cite{mukhoti2023open}. Specifically, let $\mathbf{x}_i^l \in \mathbb{R}^{D}$ and $\mathbf{x}_j^l \in \mathbb{R}^{D}$ denote the features of two image patches $i$ and $j$ extracted from the $l$-th layer. Each feature vector is $\ell_2$-normalized, and their cosine similarity is computed to serve as the prediction of a binary classifier that determines whether the two patches belong to the same semantic category. Given the corresponding semantic labels $t(\vx_i)$ and $t(\vx_j)$, the target label for classification is set to 1 if $t(\vx_i) = t(\vx_j)$, and 0 otherwise. Performance on this binary classification task provides a measure of visual discriminability, since effective representations should yield high cosine similarity for patches from the same semantic class and low similarity otherwise. To evaluate \textbf{semantic alignment}, we extract the intermediate representations $\mathbf{x}_i^l \in \mathbb{R}^{D}$ from each individual visual token at layer $l$, and project them into the final visual–text aligned space using the last ViT layer. Based on these projected features, semantic alignment is quantified as the average accuracy between predicted and ground-truth semantic labels, following~\Cref{eq:seg}. To prevent contamination from extraneous semantic information and noisy integration during the final attention computation, we follow prior work~\cite{lan2024clearclip, zhou2022extract} by replacing the attention matrix with an identity matrix and removing both the FFN and residual connections of the last ViT layer.

\textbf{Sharp decline in visual discriminability with marginal gains in semantic alignment in the final layers.} To analyze the layer-wise dynamics of visual discriminability and semantic alignment, we investigate on four datasets: Pascal VOC~\cite{everingham2011pascal}, PASCAL Context~\cite{mottaghi2014role}, ADE20K~\cite{zhou2017scene}, and COCO-Stuff~\cite{caesar2018coco}. For each dataset, we evaluate the ViT-B/16 and ViT-L/14 variants of the CLIP vision encoder using 1,000 randomly selected training samples. As shown in~\Cref{fig:layer-auc}, we observe a consistent pattern across datasets:}
\begin{itemize}[leftmargin=0.13in,topsep=0.2em,itemsep=0.11em]
    \item {\it Visual discriminability follows an inverted U-shaped curve across layers}: it increases in the early stages but declines sharply in the deeper layers. For instance, in the ViT-B/16 model, the last two layers preceding the final layer exhibit a pronounced reduction in visual discriminability, while in the ViT-L/14 model, a similar decline is observed over the last seven layers.
    \item {\it Semantic alignment exhibits an approximately monotonic increase across layers}: it improves substantially in the early stages but gradually saturates in the later layers, yielding only marginal gains thereafter.
\end{itemize}
These observations offer a nuanced understanding of why CLIP has proven effective for open-vocabulary semantic segmentation. In particular, the strong semantic alignment observed in the final layers explains why prior work can leverage last-layer features for aligning visual tokens with textual categories. However, the significant decline in visual discriminability in last layers reveals a key limitation as they may lack the fine-grained visual distinctions necessary for producing accurate and precise segmentation masks. In this work, we aim to mitigate this limitation by proposing methods that jointly improve visual discriminability and preserve semantic alignment. Before introducing our approach, we first investigate the underlying causes of the decline in visual discriminability.

\begin{figure}[t]
    \centering
    \includegraphics[width=\linewidth]{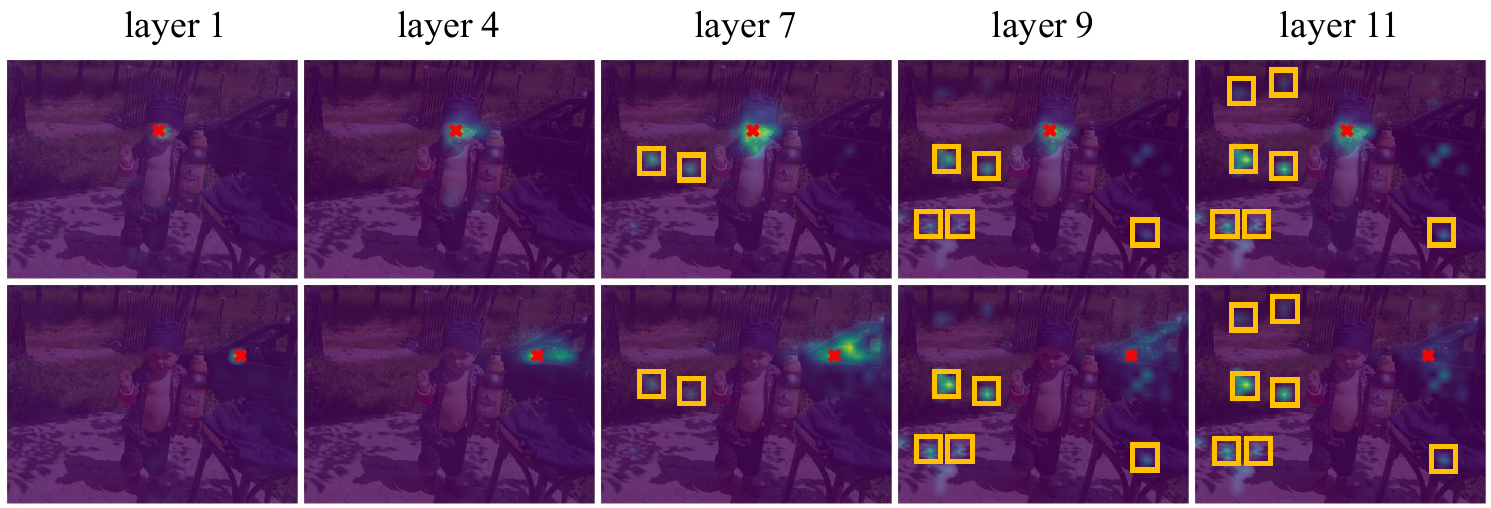}
    \caption{Abnormal token phenomenon in attention maps across different layers of the ViT-B/16 model used as the CLIP vision encoder. Attention maps are computed with respect to specific visual token positions, denoted by $\textcolor{red}{\boldsymbol{\times}}$ (e.g., the “child” token in the top row and the “car” token in the bottom row). Representative abnormal tokens are highlighted with orange boxes.}
    \label{fig:layer-abnormal-token}
    \vspace{-.1in}
\end{figure}

\begin{figure}[t]
    \centering
    \subfloat[Attention Map]{\includegraphics[width=0.195\textwidth]{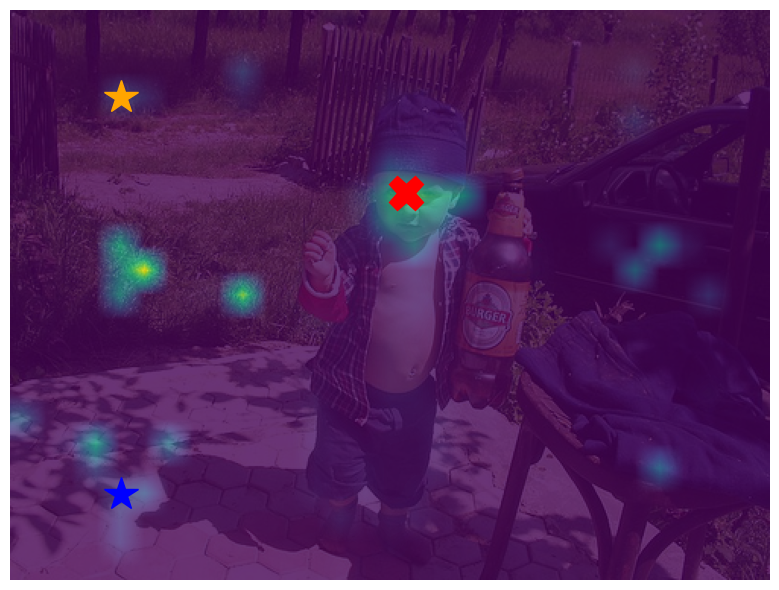}}\  
    \subfloat[Normal Token]{\includegraphics[width=0.195\textwidth]{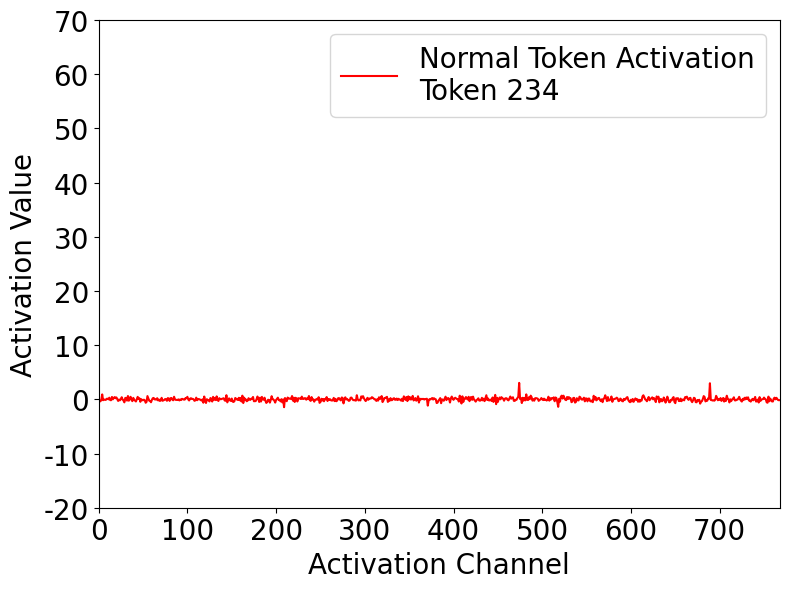}}\
    \subfloat[Abnormal Token]{\includegraphics[width=0.195\textwidth]{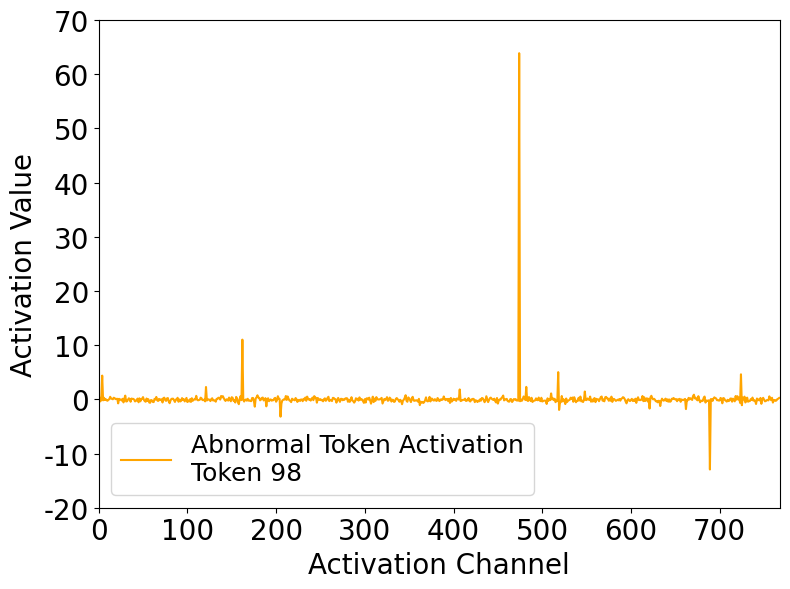}}\
    \subfloat[Abnormal Token]{\includegraphics[width=0.195\textwidth]{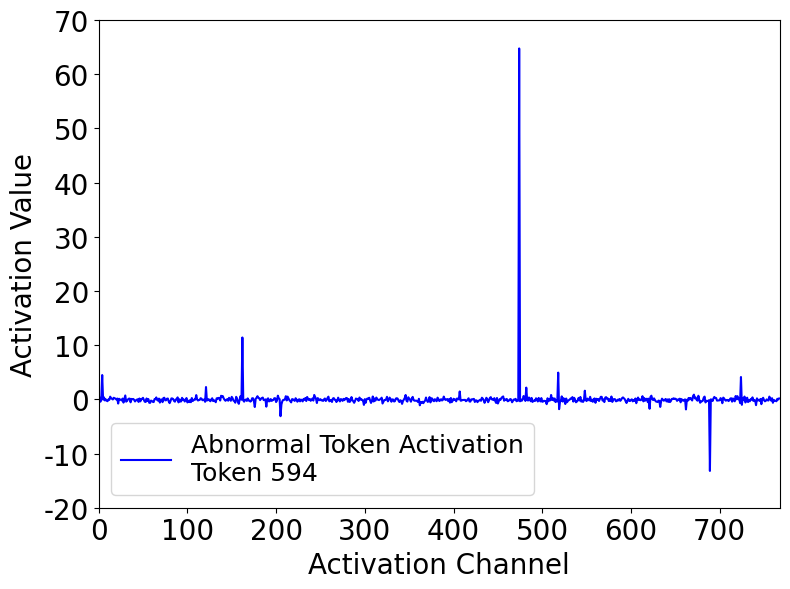}}\
    \subfloat[Hoyer Map]{\includegraphics[width=0.195\textwidth]{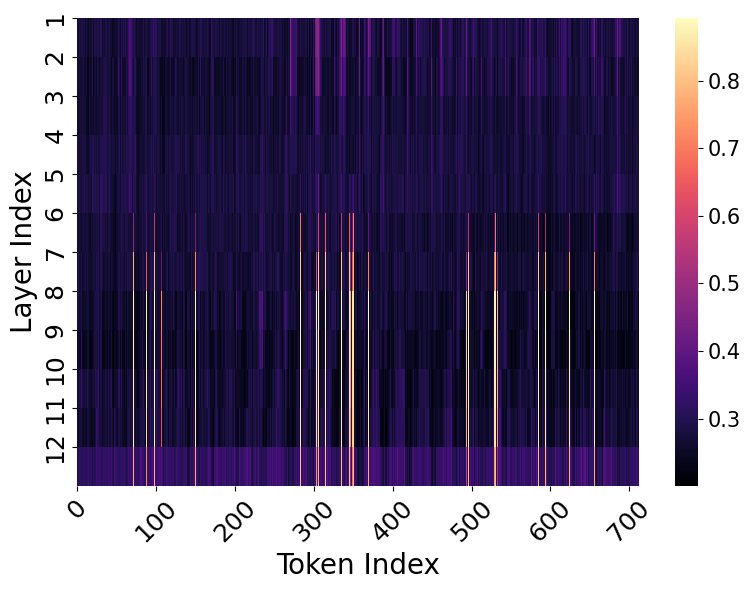}}
    \caption{Illustration of the sparsity and high-norm characteristics of abnormal tokens. Figure (a) shows the attention map of the red anchor token $\textcolor{red}{\boldsymbol{\times}}$. Figures (b)–(d) depict the channel activations of a normal token (red $\textcolor{red}{\boldsymbol{\times}}$) and two abnormal tokens (orange $\textcolor{orange}{\boldsymbol{\star}}$ and blue $\textcolor{blue}{\boldsymbol{\star}}$) highlighted in Figure (a). Figure (e) presents the hoyer score distribution across layers and token positions.}
    \vspace{-.1in}
    \label{fig:token-analysis}
\end{figure}

\revise{\textbf{Emergence of abnormal tokens with high norms and sparse activations.} To understand the cause of the sharp decline observed in the final layers, we analyze the attention maps among visual tokens across different layers. As shown in~\Cref{fig:layer-abnormal-token}, deeper layers exhibit a small set of dominant tokens that receive disproportionately high attention from nearly all visual tokens, causing most tokens to focus on this subset, consistent with prior observations~\cite{darcet2023vision, shao2024explore}. The presence of dominant tokens attracts visual tokens to become similar, thereby reducing visual discriminability and degrading segmentation result.  To further characterize these dominant tokens, we compare their features with those of normal tokens. As illustrated in~\Cref{fig:token-analysis}, dominant tokens exhibit sparse activation patterns, with only a few channels maintaining high activation. To quantify this sparsity, we adopt the hoyer score~\cite{hoyer2004non}}:
\begin{align}
\mathcal{H}(\mathbf{x}_{i}^l) = \frac{\sqrt{D} - \frac{\|\mathbf{x}_{i}^l\|_1}{\|\mathbf{x}_i^l\|_2}}{\sqrt{D} - 1}\in [0, 1],
\end{align}

\revise{Let $\mathbf{x}_i^l \in \mathbb{R}^D$ denote the feature vector of the $i$-th token at layer $l$. A higher score corresponds to sparser activations. We employ this metric to quantify sparsity and to visualize its distribution across layers and token positions in~\Cref{fig:token-analysis}(d). As shown in the figure, sparse high-norm abnormal tokens begin to emerge in the middle layers and persist at the same positions in subsequent layers.}

\begin{figure}[t]
    \centering
    \subfloat[Inter-token Similarity]{\includegraphics[width=0.30\textwidth]{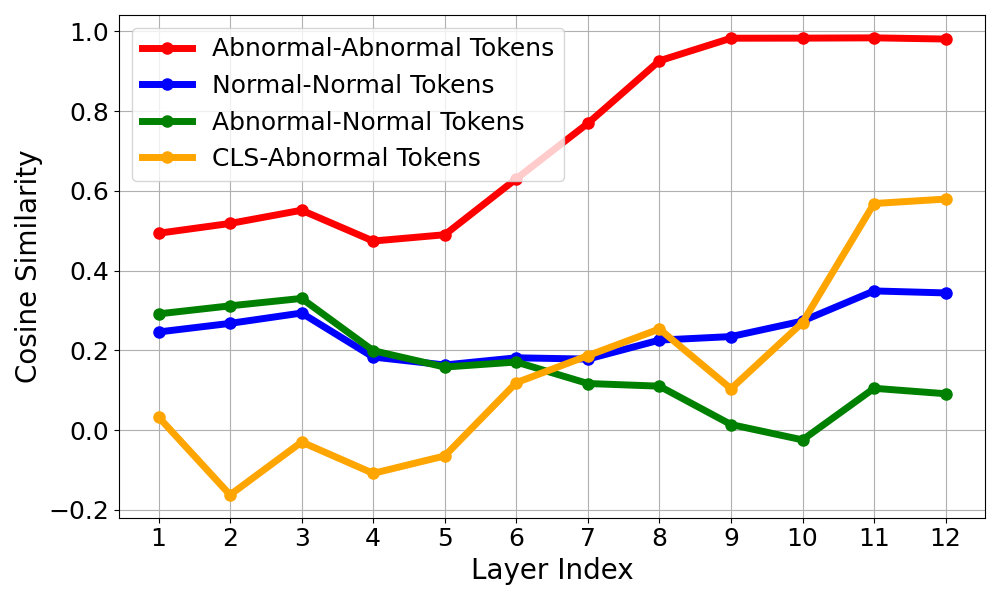}}\ 
    \subfloat[Inter-layer Similarity]{\includegraphics[width=0.30\textwidth]{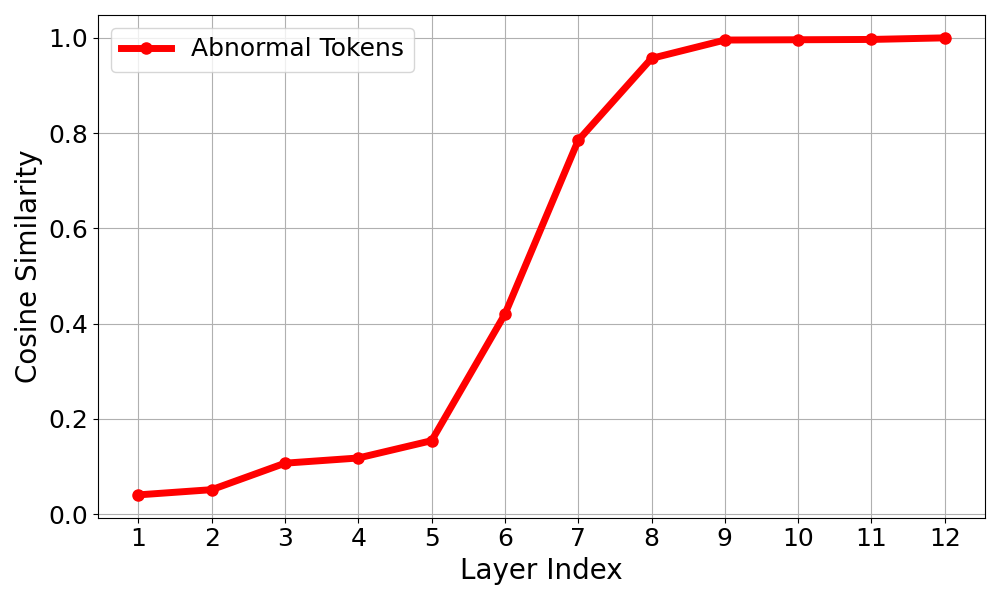}}\ 
    \subfloat[Inter-sample Similarity]{\includegraphics[width=0.30\textwidth]{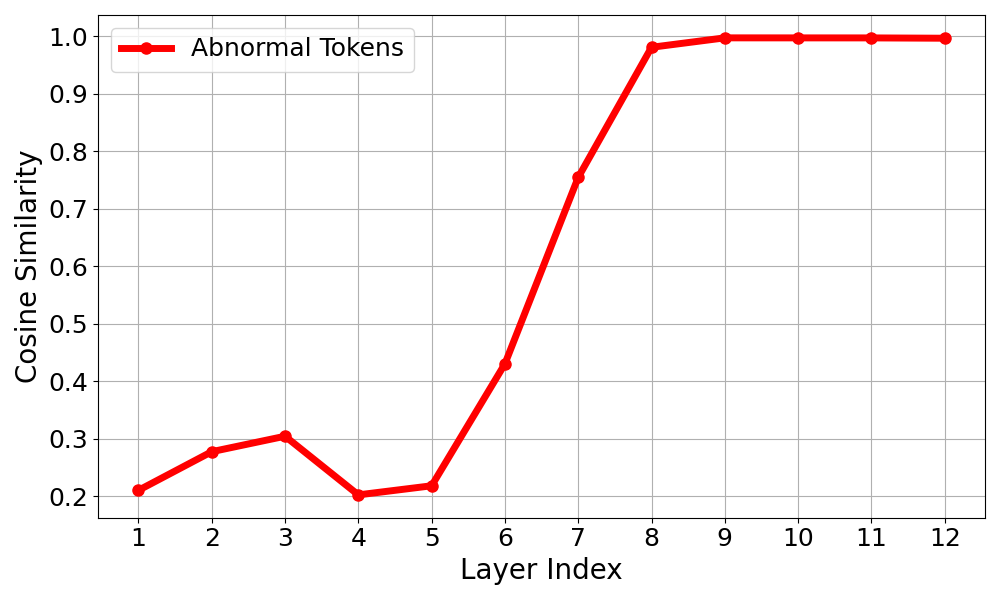}}
    \caption{Layer-wise cosine similarity among abnormal tokens across positions, layers and samples.}
    \vspace{-.1in}
\label{fig:token-property}
\end{figure}

\revise{\textbf{Abnormal tokens encode global information while being dominated by a bias-like component.} To further explore the information encoded in abnormal tokens, we analyze their pairwise cosine similarity with normal tokens and the [CLS] token on the same images from the ImageNet~\cite{deng2009imagenet} validation set. As shown in~\Cref{fig:token-property}(a), the cosine similarity with [CLS] increases while the similarity with normal tokens decreases as the layer depth increases, suggesting that abnormal tokens progressively align with global information and discard local details. Interestingly, when analyzing their pairwise cosine similarity across different positions, layers, and data samples, we find that these tokens exhibit consistently high similarity, with cosine values exceeding 0.98, as shown in~\Cref{fig:token-property}. Complementary to prior discovery, this indicates that abnormal tokens primarily act as bias components that offset global mean features, thereby facilitating text alignment in a manner analogous to the bias term in final-layer classifiers under neural collapse~\cite{zhu2021geometric, zhou2022optimization}. However, both global information and bias-like components in abnormal tokens are detrimental for segmentation tasks, which rely on fine-grained local understanding.}

\begin{figure}[t]
    \centering
    
    \subfloat[Context (ViT-B)]{\includegraphics[width=0.32\textwidth]{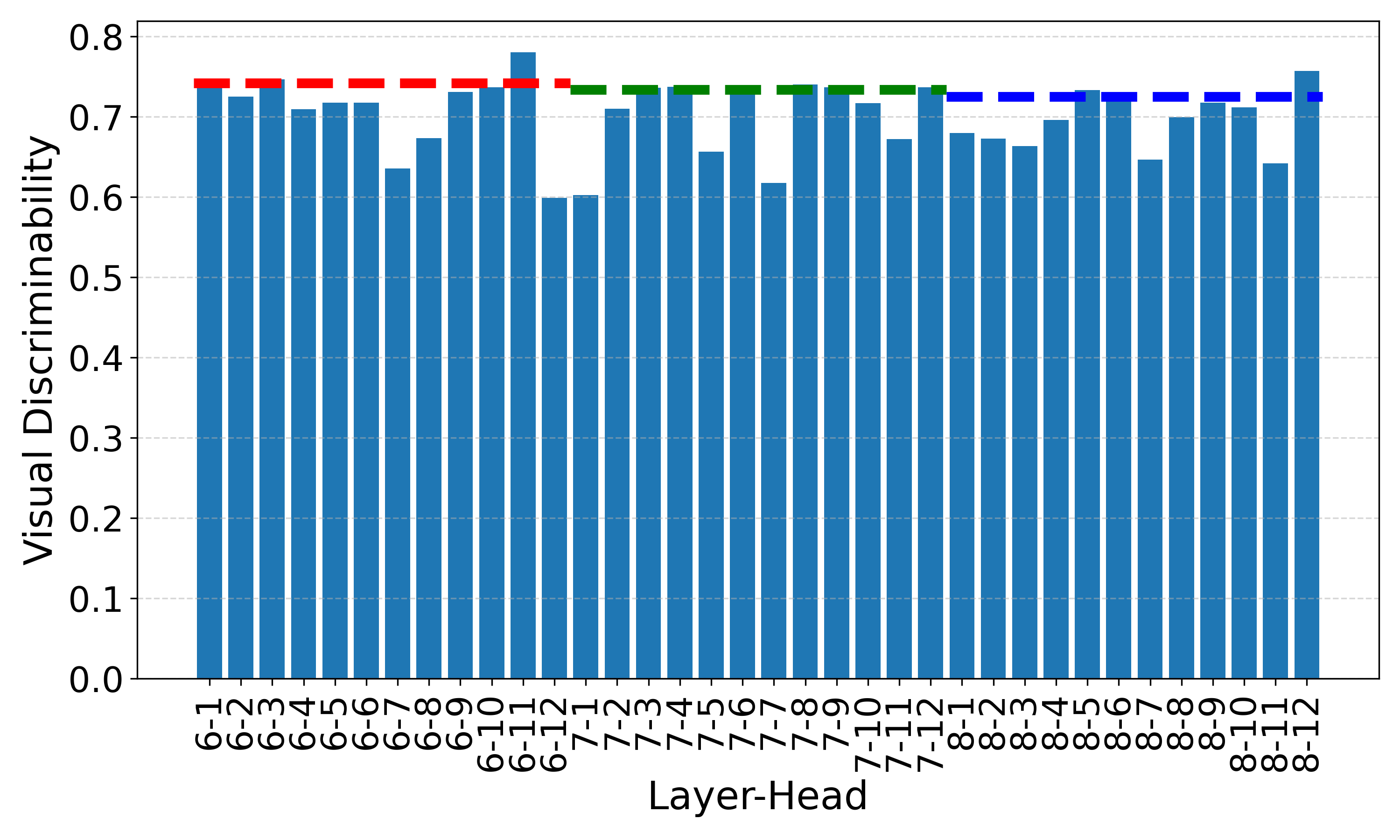}}\ 
    \subfloat[ADE (ViT-B)]{\includegraphics[width=0.32\textwidth]{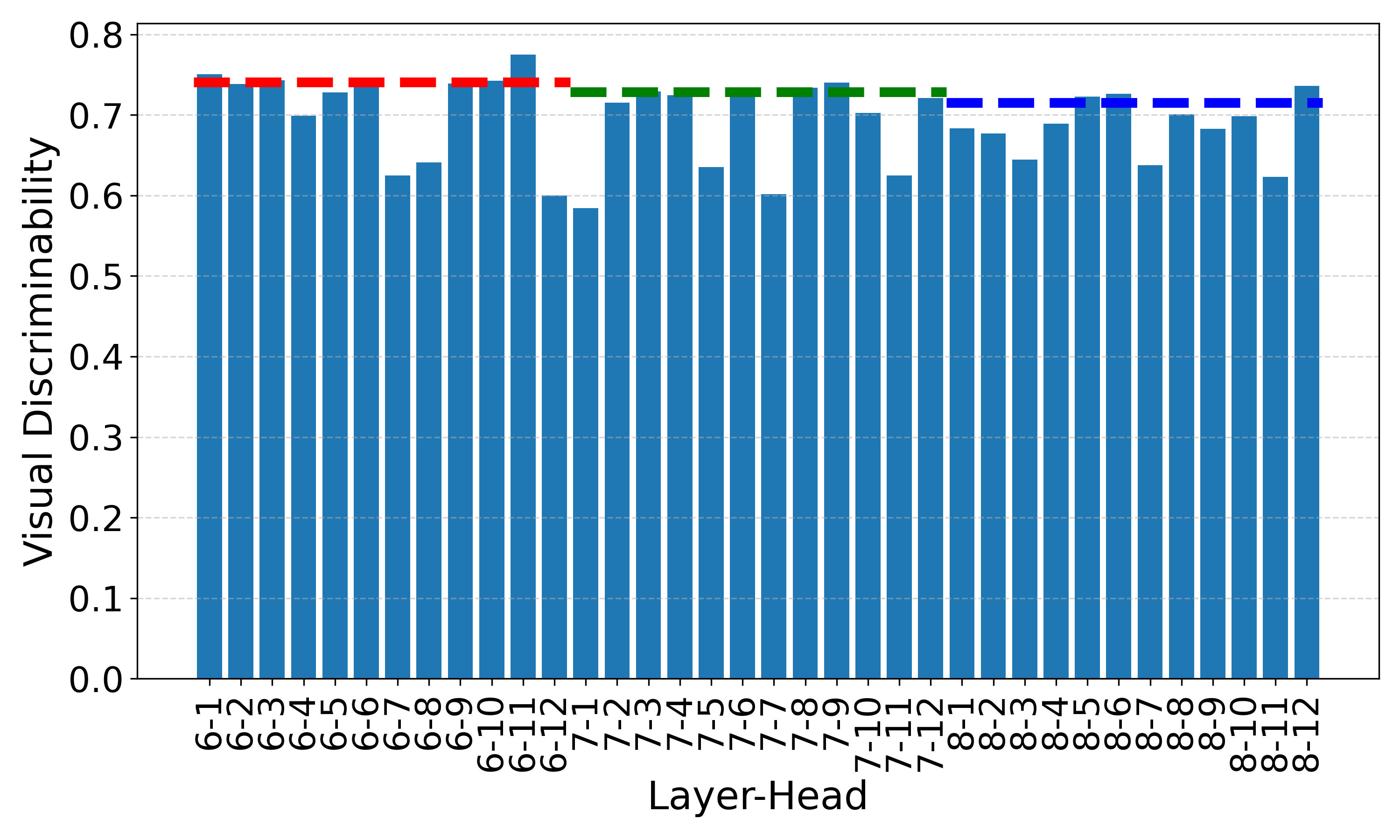}}
    \subfloat[COCO-Stuff (ViT-B)]{\includegraphics[width=0.32\textwidth]{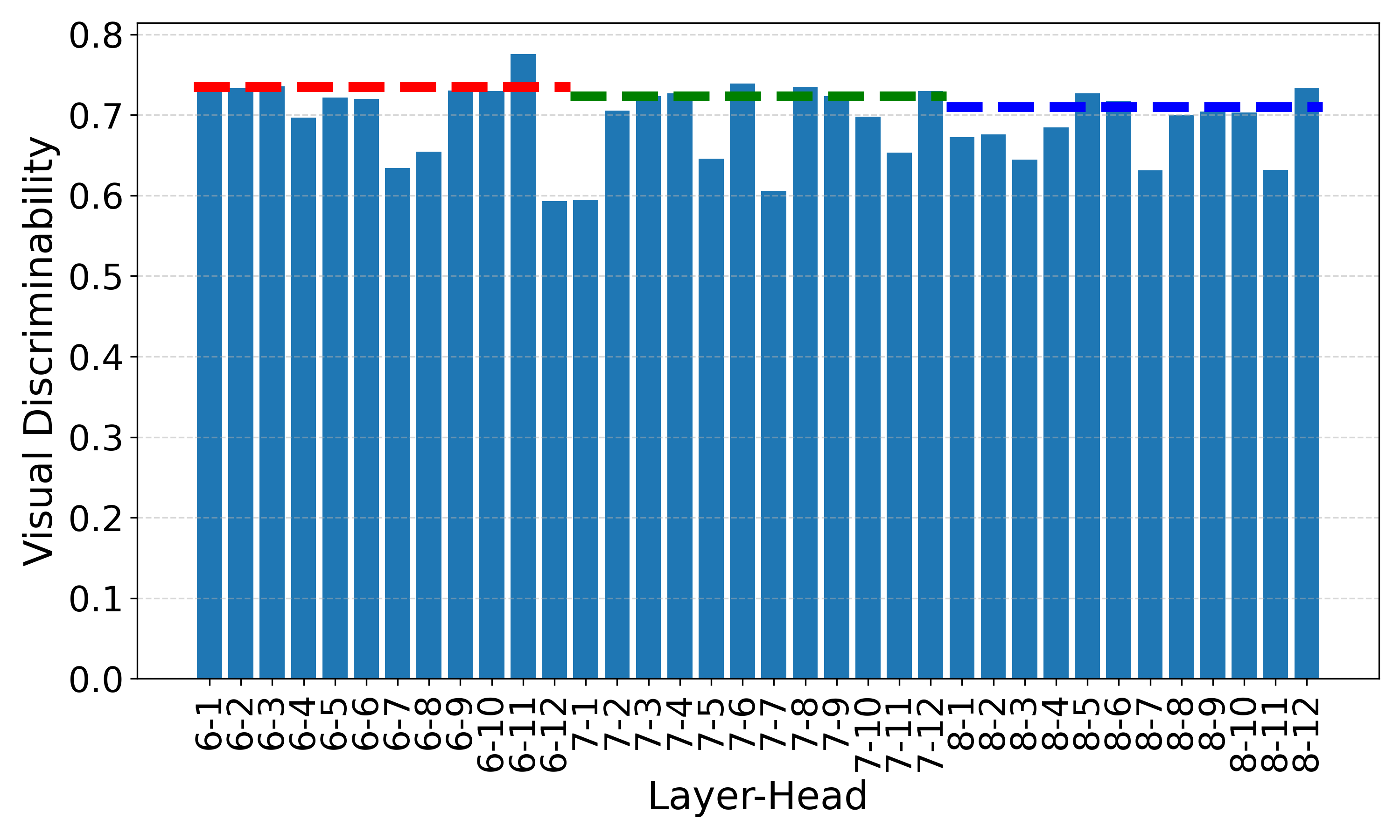}}
    \caption{Head-wise visual discriminability analysis across multiple datasets using the ViT-B/16 backbone. The dashed lines in different colors denote the corresponding layer-wise visual discriminability scores. For clarity, heads from three layers (i.e., the 6th, 7th, and 8th layers) are displayed. }
    \vspace{-.1in}
\label{fig:head_auc}\end{figure}

\textbf{A subset of attention heads exhibits consistently strong visual discriminability.} \revise{To enhance the visual discriminability of last-layer features, a natural strategy is to exploit the more discriminative intermediate features to construct a pseudo mask, guiding the reorganization of the final-layer features, as indicated by the analysis in~\Cref{fig:layer-auc}. Inspired by recent studies~\cite{gandelsman2023interpreting, kang2025your} showing that different attention heads capture distinct visual concepts, such as number, shape and texture, we take a step further by investigating whether specific heads are particularly responsible for encoding visual discriminability. To identify such heads, we follow the formulation introduced in~\cite{elhage2021mathematical, gandelsman2023interpreting}, which rewrites the MSA output as a summation over $H$ independent attention heads: $\text{MSA}(\text{LN}(\mX^{l}))=\sum_{h=1}^H\mathbf{A}^l_h\mathbf{V}^l_h\mathbf{W}^l_o\in\mathbb{R}^{(1+hw)\times D}$, where $\mathbf{A}^l_h$ and $\mathbf{V}^l_h$ denote the attention and value matrices for the $h$-th head at layer $l$, and $\mathbf{W}^l_o$ is the output projection matrix shared across all heads. Accordingly, the features of the $h$-th head at layer $l$ are expressed as:
\begin{align}
\mX^{l,h}=\mathbf{A}^l_h\mathbf{V}^l_h\mathbf{W}^l_o.
\end{align}
In~\Cref{fig:head_auc}, we show the distribution of visual discriminability across attention heads for ViT-B/16. From the figure, we observe that the output features of certain attention heads, such as the 11th head in the 6th layer, consistently exhibit high visual discriminability across different datasets, suggesting that there are a subset of heads which are more effective in capturing locally distinguishable features.}

\section{Method for Improving Visual Discriminability}\label{sec:method}
In this section, we introduce our training-free framework, which comprises three components: Abnormal Token Replacement (ATR) in \Cref{subsec:atr}, Spatial-Semantic Reweighting (SSR) in~\Cref{subsec:ssr}, and Selective Head Enhancement (SHE) in \Cref{subsec:she}. Each component is complementary, and together they work synergistically to enhance the visual discriminability of the CLIP model. 
\subsection{Abnormal token replacement (ATR)}\label{subsec:atr}
\textbf{Identify and replace abnormal tokens with their neighbors.} To mitigate the adverse effects of anomalous tokens, we propose a simple and effective strategy to suppress their impact. As shown in earlier analysis, these tokens are characterized by high norms and sparse activations. \revise{To identify them systematically, we compare a norm-based and a sparsity-based criterion. Empirically, we find that the sparsity-based approach is more robust to hyperparameter selection (see~\Cref{sec:norm-sparse-compare}); therefore, we adopt the hoyer score $\mathcal{H}(\mathbf{x}_i^l)$ defined before as a sparsity-based criterion.} Tokens with scores exceeding a predefined threshold $\tau$ are deemed anomalous and grouped into the set $\mathcal{A}_{l}=\{i|\mathcal{H}(\mathbf{x}_i^l)>\tau\}$. To mitigate their impact, each anomalous token at spatial position $(m,n)\in\mathcal{A}$ is substituted with a weighted aggregation of its eight nearest neighbors: 
\begin{align}
    \vx_{m, n}^{l}&=\frac{\sum_{i=m-1}^{m+1}\sum_{j=n-1}^{n+1} \mathbbm{1}((i,j)\notin\mathcal{A})\vx_{i, j}^{l}}{\sum_{i=m-1}^{m+1}\sum_{j=n-1}^{n+1} \mathbbm{1}((i,j)\notin\mathcal{A})}, \quad \forall (m, n) \in \mathcal{A} 
\end{align}
Here, $\mathbbm{1}(\cdot)$ ensure that only normal tokens contribute to the replacement of anomalous ones. Empirically, we find that applying this strategy before the penultimate layer leads to a performance drop, likely due to the removal of inherent biases encoded in abnormal tokens, which substantially alters the inference process. Therefore, we apply it only at the penultimate layer, i.e., with $l = L - 1$.

\subsection{Spatial-semantic reweighting (SSR)}\label{subsec:ssr} 
\textbf{Upweight residual component to improve visual discriminability.} \revise{After mitigating the impact of anomalous tokens at the penultimate layer, the final features exhibit improved visual discriminability (see~\Cref{fig:last-layer-roc} in~\Cref{sec:improve-vd}). However, a critical challenge remains: anomalous tokens appeared in earlier layers already degrade visual discriminability, limiting the effectiveness of final-layer refinements. To address it, we propose a spatial–semantic reweighting strategy, based on our layer-wise analysis showing that the final layers provide only marginal gains in semantic alignment while substantially reducing visual discriminability. Overall, the proposed strategy aims to enhance the model’s spatial coherence while preserving its semantic alignment.} Specifically, given the feature representation $\mX^{l-1}$ at the $l$-th layer within the final few layers (e.g., layers 10–11 in ViT-B/16 and layers 17–23 in ViT-L/14), we reweight the forward pass by upweighting the residual pathway and downweighting the attention and MLP submodules, as follows:
\begin{align}
    \hat{\mX}^{l}&=(1+\alpha)\mX^{l-1}+(1-\alpha)\text{MSA}(\text{LN}(\mX^{l-1})),\\
    \mX^{l}&=(1+\alpha)\hat{\mX}^{l}+(1-\alpha)\text{FFN}(\text{LN}(\hat{\mX}^{l})),
\end{align}
where $\alpha \in [0, 1]$ is a reweighting coefficient that controls the emphasis on the residual signal. As $\alpha$ increases, the $l$-th block preserves more visually discriminative features from earlier layers through the residual pathway, while reducing the adverse impact of noisy semantic aggregation in the MSA and FFN submodules. To the best of our knowledge, prior work has primarily focused on reforming the final layer or modifying its representations to improve performance. However, such approaches inevitably inherit the global semantic alignment bias of the proceeding layers, leading to suboptimal segmentation due to substantially reduced visual discriminability. In contrast, our SSR strategy explicitly addresses this limitation by rebalancing residual and semantic contributions, thereby improving the visual discriminability of intermediate features, as validated in~\Cref{fig:layer-auc-ssr}.


\subsection{Selective head enhancement (SHE)}\label{subsec:she}
\revise{\textbf{Construct pseudo masks from discriminative heads to further enhance penultimate features. } Based on the head-wise analysis, we leverage high-performing heads to construct soft pseudo masks, which are used to enhance the visual discriminability of the penultimate features. Specifically, let $\text{VD}_{l,h}^{s}$ denote the visual discriminability score of the $h$-th head in the $l$-th layer for dataset $s \in \{\text{Context}, \text{ADE}, \text{Stuff}\}$. To obtain a dataset-agnostic measurement, we compute the average score (described in \Cref{subsec:analysis-SDSA}) across datasets, denoted as $\overline{\text{VD}}_{l,h}$. We then rank heads by their $\overline{\text{VD}}_{l,h}$ values (see~\Cref{fig:head_auc_avg_b} and~\Cref{fig:head_auc_avg_l}) and select the top-$k$ heads to form the set $\mathcal{H}_{k}$. The corresponding features are aggregated as $\overline{\mX}_{k} = \frac{1}{k} \sum_{(l,h) \in \mathcal{H}_{k}} \mX^{l, h}$, and a similarity map is constructed as $S = {\overline{\mX}_{k}  \overline{\mX}_{k}^\top}/{\left\| \overline{\mX}_{k} \right\|^2}$ to capture the pairwise similarity among visual tokens. To suppress the spurious interactions between tokens from different semantic categories, we apply a thresholding operation with a parameter $\beta$, yielding the filtered similarity map $S_\beta$, where
$S_\beta(i,j) = S(i,j)$ if $S(i,j) \geq \beta$, and $S_\beta(i,j) = 0$ otherwise. This filtered similarity map $S_\beta$ serves as a soft pseudo-mask, which is column-normalized and used to refine the final-layer features as $\mX^{L-1}\leftarrow\text{Norm}( S_\beta)\mX^{L-1}$.}

\section{Experiment Results}\label{sec:experiment}

\textbf{Evaluation Protocol.}
We follow the evaluation protocol from prior works~\cite{wang2024sclip, lan2024clearclip, hajimiri2025pay} and assess our method on eight widely used semantic segmentation benchmarks. We group them into two categories and use abbreviated names for clarity. The first category excludes background and includes Pascal VOC~\cite{everingham2011pascal} (VOC20), Pascal Context~\cite{mottaghi2014role} (C59), COCO-Stuff~\cite{caesar2018coco} (Stuff), ADE20K~\cite{zhou2017scene} (ADE), and Cityscapes~\cite{cordts2016cityscapes} (City). The second includes background and consists of VOC21, C60, and COCO-Object~\cite{caesar2018coco} (Object). We use ViT-B/16 and ViT-L/14 as the vision encoder of CLIP~\cite{radford2021learning}, and report results using the mean Intersection-over-Union (mIoU). All hyperparameters are tuned on 1,000 randomly sampled training images from \{Context, ADE, Stuff \} datasets and kept fixed without task-specific tuning during evaluation, based on ablation study in~\Cref{sec:ablation-b}. Additional details are provided in the appendix.

\subsection{Comparison with existing methods.}\label{sec:results-vit-b}
We compare our approach against a comprehensive set of open-vocabulary semantic segmentation methods, including the direct baseline CLIP~\cite{radford2021learning}, as well as several state-of-the-art training-free approaches: MaskCLIP~\cite{zhou2022extract}, CLIPSurgery~\cite{li2023clip}, SCLIP~\cite{wang2024sclip}, NACLIP~\cite{hajimiri2025pay}, ClearCLIP~\cite{lan2024clearclip}, LAVG~\cite{kang2024defense}, and ResCLIP~\cite{yang2024resclip}. We also include several influential weakly supervised methods, such as GroupViT~\cite{xu2022groupvit},  ReCo~\cite{shin2022reco}, and TCL~\cite{cha2023learning}. Unless otherwise specified, all reported results are taken directly from the respective original papers and ResCLIP~\cite{yang2024resclip}. As our method is orthogonal to approaches that primarily target improvements in the final-layer attention, we evaluate its effectiveness when integrated with recent state-of-the-art methods that employ specialized attention mechanisms in the last layer, including SCLIP~\cite{wang2024sclip}, ClearCLIP~\cite{lan2024clearclip}, and ResCLIP~\cite{yang2024resclip}. For fair comparison, we exclude the \emph{Semantic Feedback Refinement} module in ResCLIP, as it relies on the computationally expensive PAMR~\cite{araslanov2020single} post-processing, which is inconsistent with our evaluation setting. 

In \Cref{tab:ovss-results-b}, we summarize the performance of various previous methods on benchmark datasets using the ViT-B/16 backbone. Our proposed LHT-CLIP consistently enhances the performance of state-of-the-art approaches, including SCLIP~\cite{wang2024sclip}, ClearCLIP~\cite{lan2024clearclip}, and ResCLIP~\cite{yang2024resclip}. Notably, when integrated with ResCLIP~\cite{yang2024resclip}, LHT-CLIP achieves state-of-the-art results, outperforming leading weakly supervised methods. As a plug-and-play solution, LHT-CLIP yields consistent improvements across all datasets compared to the respective baselines, demonstrating its strong generalization capability. For comprehensiveness, results on the ViT-L/14 backbone are provided in~\Cref{tab:ovss-results-l}. In line with observations from~\cite{yang2024resclip}, existing methods generally exhibit a performance drop exceeding 2\% mIoU when adapting to a different backbone; for instance, ClearCLIP~\cite{lan2024clearclip} suffers a notable decline of 2.7\% mIoU. In contrast, when augmented with LHT-CLIP, this performance degradation is significantly alleviated, highlighting the robustness of our approach. Across both backbones, LHT-CLIP delivers substantial improvements over baseline methods, validating its effectiveness. 

\begin{table*}[t]
\centering
\caption{Performance comparison of our approach with other methods on eight semantic segmentation benchmarks. Existing methods with our improvement are marked in \graybox{gray}.}
\label{tab:ovss-results-b}
\footnotesize
\tabcolsep=0.22em
\begin{tabular}{llccccccccc}
\toprule
\multirow{2}{*}{\textbf{Methods}} & \multirow{2}{*}{\textbf{Training}} & \multicolumn{3}{c}{With a background class} & \multicolumn{5}{c}{Without background class} & \multirow{2}{*}{Avg.} \\ \cmidrule(lr){3-5}\cmidrule(lr){6-10}
                        &                                & VOC21       & C60       & Object      & VOC20   & City  & C59  & ADE  & Stuff  &                       \\
 \midrule
ReCo & \multicolumn{1}{c}{\ding{51}}  & 25.1 & 19.9 & 15.7 & 57.7 & 21.1 & 22.3 & 11.2 & 14.8 & 23.5 \\
GroupViT &  \multicolumn{1}{c}{\ding{51}} & 52.3 & 18.7 & 27.5 & 79.7 & 18.5 & 23.4 & 10.4 & 15.3 & 30.7 \\
TCL & \multicolumn{1}{c}{\ding{51}}  & 51.2 & 24.3 & 30.4 & 77.5 & 23.1 & 30.3 & 14.9 & 19.6 & 33.9 \\
CLIP                    & \multicolumn{1}{c}{\ding{55}}           & 16.2        & 7.7             & 5.5         & 41.8    & 5.5  & 9.2       & 2.1 & 4.4   & 11.6\\
MaskCLIP & \multicolumn{1}{c}{\ding{55}} & 38.8 & 23.6 & 20.6 & 74.9 & 16.4 &26.4 & 9.8 & 14.8 & 28.2 \\     
CLIPSurgery & \multicolumn{1}{c}{\ding{55}} & 55.2 & 18.7 & 27.5 & 79.7 & 18.5 &23.4 & 10.4 & 15.3 & 31.1 \\ 
LaVG & \multicolumn{1}{c}{\ding{55}} & 62.1 & 31.6 & 34.2 & 82.5 &26.2 & 34.7 &15.8 & 23.2 & 38.8\\
NACLIP & \multicolumn{1}{c}{\ding{55}} &58.9 & 32.2 & 33.2 & 79.7 &35.5 & 35.2 &17.4 & 23.3 & 39.4\\
\midrule
SCLIP & \multicolumn{1}{c}{\ding{55}} &59.7 & 31.7 & 33.5 & 81.5 &32.3 &34.5 &16.5 & 22.7 & 39.1\\
\rowcolor{gray!20}
+LHT-CLIP (ours)  & \multicolumn{1}{c}{\ding{55}} &64.8 & 34.8 & 36.6 & 86.3 &36.1 &37.6 & 18.0 & 24.9&42.4 \green{(+3.3)}\\
ClearCLIP &  \multicolumn{1}{c}{\ding{55}} &57.0 & 32.2 & 32.5 & 82.3 &32.8 &35.8 &17.3 & 24.0 & 39.2\\
\rowcolor{gray!20}
+LHT-CLIP (ours)  & \multicolumn{1}{c}{\ding{55}} &63.8 & 35.2 & 35.6 & 85.7 &37.8 &38.8&19.2 & 25.8 &  42.7 \green{(+3.5)}\\
ResCLIP & \multicolumn{1}{c}{\ding{55}} &60.0 & 32.7 & 34.0 & 85.5 &35.6 & 35.8 &17.7 & 23.8 & 40.6\\
\rowcolor{gray!20}
+LHT-CLIP (ours)  & \multicolumn{1}{c}{\ding{55}} &63.9 & 35.5 & 35.2 & 86.9 &38.2 &38.2&19.1 & 25.5 & 42.8 \green{(+2.2)}\\
\bottomrule 
\end{tabular}
\vspace{-.1in}
\end{table*}

\subsection{Experimental analysis}\label{sec:ablation-b}
In this section, we conduct comprehensive ablation studies to validate the effectiveness of our proposed method. We adopt SCLIP~\cite{wang2024sclip} as the baseline, which enhances spatial correlation by modifying the attention mechanism in the final layer, replacing the standard $QK^\top$ attention with a combination of $QQ^\top + KK^\top$. In addition, following prior work~\cite{lan2024clearclip, yang2024resclip}, we remove the residual connections and FFN from the final transformer layer. For experiments in this part, we highlight the optimal hyperparameter settings in \graybox{gray}.

\textbf{Analysis of the hoyer threshold parameter $\tau$.} Our method relies on hoyer sparsity to identify anomalous tokens, making the sparsity threshold $\tau$ a critical hyperparameter. In~\Cref{tab:hoyer-thres}, We conduct a systematic evaluation. At $\tau = 0.2$, many normal tokens are misclassified, leading to excessive smoothing and degraded performance. As $\tau$ increases to 0.4, performance steadily improves, but plateaus between 0.5 and 0.8, with a decline observed beyond this range. The broad stable region indicates a clear sparsity gap between normal and abnormal tokens, highlighting the robustness of ATR to threshold selection. Based on this analysis, we fix $\tau = 0.5$ for all experiments.

\begin{table}[t]  
\noindent
\begin{minipage}{0.48\linewidth}
\centering
\footnotesize
\captionof{table}{Study of hoyer sparsity threshold $\tau$.}
\label{tab:hoyer-thres}
\begin{tabular}{cccccc}
\toprule
$\tau$ & C60 & Stuff & C59 & ADE & Avg \\
\midrule
$\tau = 0.2$ & 1.3 & 1.2 & 1.5 & 0.6 & 1.2\\
$\tau = 0.4$ & 33.0 & 24.4 & 36.6 & 17.9 & 28.0 \\
\rowcolor{gray!20}
$\tau = 0.5$ &   33.0 & 24.4& 36.7 & 17.9 & 28.0  \\
$\tau = 0.8$ &   33.0 & 24.4 & 36.6 & 17.9 & 28.0   \\
$\tau = 0.9$ &   32.4 & 24.0 & 36.0 & 17.6 & 27.5   \\
baseline & 32.4 & 24.0 & 36.0 & 17.6 & 27.5  \\
\bottomrule
\end{tabular}
\end{minipage}%
\hfill
\begin{minipage}{0.51\linewidth}
\centering
\footnotesize
\captionof{table}{Study of $(l_{\text{start}}, l_{\text{end}}, \alpha)$ in SSR module.}
\label{tab:reweight}
\begin{tabular}{cccccc}
\toprule
$(l_{\text{start}}, l_{\text{end}}, \alpha)$ & C60 & Stuff & C59 & ADE & Avg \\
\midrule
baseline & 32.4 & 24.0 & 36.0 & 17.6 & 27.5 \\
(9, 11, 0.1)  & 32.7 & 23.7 & 36.5 & 17.7 & 27.7 \\
\rowcolor{gray!20}
(10, 11, 0.1) & 33.1 & 24.3 & 36.9 & 18.0 & 28.1 \\
(11, 11, 0.1) & 32.7 & 24.4 & 36.4 & 18.0 & 27.9 \\
(10, 11, 0.05) & 32.8 & 24.3 & 36.4 & 18.0 & 27.9 \\
(10, 11, 0.2) & 32.6 & 23.9 & 36.5 & 17.8 & 27.7 \\
\bottomrule
\end{tabular}
\end{minipage}
\vspace{-.1in}
\end{table}

\begin{table}[t]  
\noindent
\begin{minipage}{0.48\linewidth}
\centering
\footnotesize
\captionof{table}{Study of number of selected heads $k$.}
\label{tab:k-heads}
\begin{tabular}{cccccc}
\toprule
$k$ & C60 & Stuff & C59 & ADE & Avg \\
\midrule
baseline & 32.4 & 24.0 & 36.0 & 17.6 & 27.5
\\
layer($l=8$) & 32.8 & 24.8 & 36.4 & 17.7 & 28.0 \\
$k=5$ &  32.8 & 24.6 & 36.3 & 17.8 & 27.9 \\
\rowcolor{gray!20}
$k=10$ &  33.0 & 25.1 &  37.0 & 18.3 & 28.4 \\
$k=30$ & 33.0 & 25.0 & 37.1 & 18.2 & 28.3 \\
$k=50$ & 32.8 & 25.0 & 37.0 & 18.1 & 28.2 \\
\bottomrule
\end{tabular}
\end{minipage}%
\hfill
\begin{minipage}{0.48\linewidth}
\centering
\footnotesize
\captionof{table}{Combination of three strategies.}
\label{tab:ablation}
\begin{tabular}{lccccc}
\toprule
 & \multicolumn{3}{c}{\textbf{Module}} & \textbf{mIoU} & $\Delta$ \\
\cmidrule(lr){2-4}
 & \textit{ATR} & \textit{SSR} &\textit{SHE} & & \\
\midrule
baseline & -- & -- & -- & 27.5 & -- \\
 & \checkmark & \checkmark & -- & 28.6 & \textcolor{green!60!black}{+1.1} \\
  & -- & \checkmark & \checkmark & 28.9 & \textcolor{green!60!black}{+1.4} \\
 & \checkmark & -- & \checkmark & 28.9& \textcolor{green!60!black}{+1.4} \\
\rowcolor{gray!10} 
 & \checkmark & \checkmark& \checkmark & \textbf{29.2} & \textcolor{green!60!black}{\textbf{+1.7}} \\
\bottomrule
\end{tabular}
\end{minipage}
\vspace{-.1in}
\end{table}
\textbf{Analysis of spatial-semantic reweighting parameters $\alpha$ and number of layers.} To evaluate the impact of the reweighting strength $\alpha$ and the range of layers involved, from $l_{\text{start}}$ to $l_{\text{end}}$, we perform a comprehensive sensitivity analysis. The results are summarized in~\Cref{tab:reweight}. We observe that the best performance is obtained when reweighting is applied to layers 10–11 in the ViT-B/16 backbone. This aligns with our earlier findings that these layers experience a marked decline in visual discriminability while yielding only marginal improvements in semantic alignment. Extending reweighting to include layer 9 results in a slight gain in visual discriminability but introduces noisy semantic signals, ultimately leading to a reduction in segmentation performance. In addition, we examine the effect of varying the reweighting threshold parameter $\alpha$. As $\alpha$ increases from 0 to 0.1, performance improves steadily, indicating a beneficial balance between visual and semantic cues. However, further increasing $\alpha$ leads to a performance drop, as it incorporates inaccurate semantic information from earlier layers and significantly perturbs the input distribution of subsequent layers.

\textbf{Analysis of the number of selected heads $k$.}
In~\Cref{tab:k-heads}, we study the effect of varying the number of top-$k$ attention heads selected for enhancement. For ViT-B/16 backbone, increasing $k$ from 5 to 10 improves segmentation accuracy, as aggregating multiple visual discriminative heads helps suppress spurious correlations. However, performance declines when $k$ becomes too large due to the inclusion of noisy or less informative heads, which introduce noisy cross-category interactions. We also compare head- and layer-level selection (best $l=8$), finding that head-level selection consistently performs better, as discriminative heads are distributed across layers, while entire-layer selection introduces irrelevant heads and degrades performance.

\textbf{Study of combination effect.}
In prior parts, we show the effectiveness of each individual strategy in~\Cref{tab:hoyer-thres}, \Cref{tab:reweight} and \Cref{tab:k-heads}. The~\Cref{tab:ablation} further presents results of different combinations under the same settings. The result highlights the complementary contributions of each strategy to the overall segmentation performance, yielding average 1.7 mIoU improvement on these four datasets.

\section{Conclusion}\label{sec:conclusion}

In this paper, we present a comprehensive analysis of the visual discriminability of pretrained CLIP models at the token, layer, and head levels. Based on the analysis, we introduce LHT-CLIP, a training-free framework that enhances visual discriminability while preserving semantic alignment. LHT-CLIP comprises three complementary components: (1) abnormal token replacement, (2) spatial–semantic reweighting, and (3) selective head enhancement. Our approach provides lightweight, plug-and-play modules compatible with existing architectures. Extensive experiments on multiple segmentation benchmarks show the effectiveness of LHT-CLIP consistently outperforms strong baselines. Furthermore, since CLIP vision encoders are often frozen in MLLM training, our findings offer practical guidance for improving visual understanding in broader MLLMs.
\newpage
\section*{Ethics statement}
This work focuses on improving the visual discriminability of pretrained vision–language models. Our research does not involve the collection of new human or animal data, and all experiments are conducted using publicly available datasets that have been widely adopted in prior work. We acknowledge that vision–language models, including CLIP and its variants, may inherit biases present in their training data. While our method is designed to enhance visual discriminability without additional training, it does not explicitly mitigate such biases. We encourage future research to examine fairness, accountability, and transparency when deploying these models in real-world applications.

\section*{Reproducibility statement}
We have made every effort to ensure the reproducibility of our work. Full implementation details, including model architectures, hyperparameters, and experimental settings, are provided in the main paper and appendix. Our method builds on publicly available pretrained CLIP and SigLIP models and is evaluated on standard benchmark datasets commonly used in prior work. To further support reproducibility, we include our code in the supplementary material and will release the code, configuration files, and detailed instructions upon publication.

\section*{The Use of Large Language Models}
Large language models were used exclusively to assist with writing polish, grammar correction, and improving readability. They were not used for ideation, experiment design, analysis, or generating research content. All technical contributions, experimental implementations, and results reported in this paper are original work conducted by the authors.
\bibliography{reference}

\begin{thebibliography}{63}
\providecommand{\natexlab}[1]{#1}
\providecommand{\url}[1]{\texttt{#1}}
\expandafter\ifx\csname urlstyle\endcsname\relax
  \providecommand{\doi}[1]{doi: #1}\else
  \providecommand{\doi}{doi: \begingroup \urlstyle{rm}\Url}\fi

\bibitem[Araslanov \& Roth(2020)Araslanov and Roth]{araslanov2020single}
Nikita Araslanov and Stefan Roth.
\newblock Single-stage semantic segmentation from image labels.
\newblock In \emph{Proceedings of the IEEE/CVF conference on computer vision and pattern recognition}, pp.\  4253--4262, 2020.

\bibitem[Bolya et~al.(2025)Bolya, Huang, Sun, Cho, Madotto, Wei, Ma, Zhi, Rajasegaran, Rasheed, et~al.]{bolya2025perception}
Daniel Bolya, Po-Yao Huang, Peize Sun, Jang~Hyun Cho, Andrea Madotto, Chen Wei, Tengyu Ma, Jiale Zhi, Jathushan Rajasegaran, Hanoona Rasheed, et~al.
\newblock Perception encoder: The best visual embeddings are not at the output of the network.
\newblock \emph{arXiv preprint arXiv:2504.13181}, 2025.

\bibitem[Bousselham et~al.(2024)Bousselham, Petersen, Ferrari, and Kuehne]{bousselham2024grounding}
Walid Bousselham, Felix Petersen, Vittorio Ferrari, and Hilde Kuehne.
\newblock Grounding everything: Emerging localization properties in vision-language transformers.
\newblock In \emph{Proceedings of the IEEE/CVF Conference on Computer Vision and Pattern Recognition}, pp.\  3828--3837, 2024.

\bibitem[Caesar et~al.(2018)Caesar, Uijlings, and Ferrari]{caesar2018coco}
Holger Caesar, Jasper Uijlings, and Vittorio Ferrari.
\newblock Coco-stuff: Thing and stuff classes in context.
\newblock In \emph{Proceedings of the IEEE conference on computer vision and pattern recognition}, pp.\  1209--1218, 2018.

\bibitem[Cha et~al.(2023)Cha, Mun, and Roh]{cha2023learning}
Junbum Cha, Jonghwan Mun, and Byungseok Roh.
\newblock Learning to generate text-grounded mask for open-world semantic segmentation from only image-text pairs.
\newblock In \emph{Proceedings of the IEEE/CVF Conference on Computer Vision and Pattern Recognition}, pp.\  11165--11174, 2023.

\bibitem[Cherti et~al.(2023)Cherti, Beaumont, Wightman, Wortsman, Ilharco, Gordon, Schuhmann, Schmidt, and Jitsev]{cherti2023reproducible}
Mehdi Cherti, Romain Beaumont, Ross Wightman, Mitchell Wortsman, Gabriel Ilharco, Cade Gordon, Christoph Schuhmann, Ludwig Schmidt, and Jenia Jitsev.
\newblock Reproducible scaling laws for contrastive language-image learning.
\newblock In \emph{Proceedings of the IEEE/CVF conference on computer vision and pattern recognition}, pp.\  2818--2829, 2023.

\bibitem[Contributors(2020)]{mmseg2020}
MMSegmentation Contributors.
\newblock {MMSegmentation}: Openmmlab semantic segmentation toolbox and benchmark.
\newblock \url{https://github.com/open-mmlab/mmsegmentation}, 2020.

\bibitem[Cordts et~al.(2016)Cordts, Omran, Ramos, Rehfeld, Enzweiler, Benenson, Franke, Roth, and Schiele]{cordts2016cityscapes}
Marius Cordts, Mohamed Omran, Sebastian Ramos, Timo Rehfeld, Markus Enzweiler, Rodrigo Benenson, Uwe Franke, Stefan Roth, and Bernt Schiele.
\newblock The cityscapes dataset for semantic urban scene understanding.
\newblock In \emph{Proceedings of the IEEE conference on computer vision and pattern recognition}, pp.\  3213--3223, 2016.

\bibitem[Corradini et~al.(2024)Corradini, Shukor, Couairon, Couairon, Scarselli, and Cord]{Corradini2024FreeSegDiffTO}
Barbara~Toniella Corradini, Mustafa Shukor, Paul Couairon, Guillaume Couairon, Franco Scarselli, and Matthieu Cord.
\newblock Freeseg-diff: Training-free open-vocabulary segmentation with diffusion models.
\newblock \emph{ArXiv}, abs/2403.20105, 2024.
\newblock URL \url{https://api.semanticscholar.org/CorpusID:268793968}.

\bibitem[Darcet et~al.(2023)Darcet, Oquab, Mairal, and Bojanowski]{darcet2023vision}
Timoth{\'e}e Darcet, Maxime Oquab, Julien Mairal, and Piotr Bojanowski.
\newblock Vision transformers need registers.
\newblock \emph{arXiv preprint arXiv:2309.16588}, 2023.

\bibitem[Deng et~al.(2009)Deng, Dong, Socher, Li, Li, and Fei-Fei]{deng2009imagenet}
Jia Deng, Wei Dong, Richard Socher, Li-Jia Li, Kai Li, and Li~Fei-Fei.
\newblock Imagenet: A large-scale hierarchical image database.
\newblock In \emph{2009 IEEE conference on computer vision and pattern recognition}, pp.\  248--255. Ieee, 2009.

\bibitem[Dosovitskiy et~al.(2020)Dosovitskiy, Beyer, Kolesnikov, Weissenborn, Zhai, Unterthiner, Dehghani, Minderer, Heigold, Gelly, Uszkoreit, and Houlsby]{Dosovitskiy2020AnII}
Alexey Dosovitskiy, Lucas Beyer, Alexander Kolesnikov, Dirk Weissenborn, Xiaohua Zhai, Thomas Unterthiner, Mostafa Dehghani, Matthias Minderer, Georg Heigold, Sylvain Gelly, Jakob Uszkoreit, and Neil Houlsby.
\newblock An image is worth 16x16 words: Transformers for image recognition at scale.
\newblock \emph{ArXiv}, abs/2010.11929, 2020.
\newblock URL \url{https://api.semanticscholar.org/CorpusID:225039882}.

\bibitem[Elhage et~al.(2021)Elhage, Nanda, Olsson, Henighan, Joseph, Mann, Askell, Bai, Chen, Conerly, et~al.]{elhage2021mathematical}
Nelson Elhage, Neel Nanda, Catherine Olsson, Tom Henighan, Nicholas Joseph, Ben Mann, Amanda Askell, Yuntao Bai, Anna Chen, Tom Conerly, et~al.
\newblock A mathematical framework for transformer circuits.
\newblock \emph{Transformer Circuits Thread}, 1\penalty0 (1):\penalty0 12, 2021.

\bibitem[Everingham \& Winn(2011)Everingham and Winn]{everingham2011pascal}
Mark Everingham and John Winn.
\newblock The pascal visual object classes challenge 2012 (voc2012) development kit.
\newblock \emph{Pattern Analysis, Statistical Modelling and Computational Learning, Tech. Rep}, 8\penalty0 (5):\penalty0 2--5, 2011.

\bibitem[Gandelsman et~al.(2023)Gandelsman, Efros, and Steinhardt]{gandelsman2023interpreting}
Yossi Gandelsman, Alexei~A Efros, and Jacob Steinhardt.
\newblock Interpreting clip's image representation via text-based decomposition.
\newblock \emph{arXiv preprint arXiv:2310.05916}, 2023.

\bibitem[Gao et~al.(2024)Gao, Geng, Zhang, Ma, Fang, Zhang, Li, and Qiao]{gao2024clip}
Peng Gao, Shijie Geng, Renrui Zhang, Teli Ma, Rongyao Fang, Yongfeng Zhang, Hongsheng Li, and Yu~Qiao.
\newblock Clip-adapter: Better vision-language models with feature adapters.
\newblock \emph{International Journal of Computer Vision}, 132\penalty0 (2):\penalty0 581--595, 2024.

\bibitem[Hajimiri et~al.(2025)Hajimiri, Ayed, and Dolz]{hajimiri2025pay}
Sina Hajimiri, Ismail~Ben Ayed, and Jose Dolz.
\newblock Pay attention to your neighbours: Training-free open-vocabulary semantic segmentation.
\newblock In \emph{2025 IEEE/CVF Winter Conference on Applications of Computer Vision (WACV)}, pp.\  5061--5071. IEEE, 2025.

\bibitem[Hoyer(2004)]{hoyer2004non}
Patrik~O Hoyer.
\newblock Non-negative matrix factorization with sparseness constraints.
\newblock \emph{Journal of machine learning research}, 5\penalty0 (Nov):\penalty0 1457--1469, 2004.

\bibitem[Jia et~al.(2021)Jia, Yang, Xia, Chen, Parekh, Pham, Le, Sung, Li, and Duerig]{jia2021scaling}
Chao Jia, Yinfei Yang, Ye~Xia, Yi-Ting Chen, Zarana Parekh, Hieu Pham, Quoc Le, Yun-Hsuan Sung, Zhen Li, and Tom Duerig.
\newblock Scaling up visual and vision-language representation learning with noisy text supervision.
\newblock In \emph{International conference on machine learning}, pp.\  4904--4916. PMLR, 2021.

\bibitem[Kang \& Cho(2024)Kang and Cho]{kang2024defense}
Dahyun Kang and Minsu Cho.
\newblock In defense of lazy visual grounding for open-vocabulary semantic segmentation.
\newblock In \emph{European Conference on Computer Vision}, pp.\  143--164. Springer, 2024.

\bibitem[Kang et~al.(2025)Kang, Kim, Kim, and Hwang]{kang2025your}
Seil Kang, Jinyeong Kim, Junhyeok Kim, and Seong~Jae Hwang.
\newblock Your large vision-language model only needs a few attention heads for visual grounding.
\newblock \emph{arXiv preprint arXiv:2503.06287}, 2025.

\bibitem[Kr{\"a}henb{\"u}hl \& Koltun(2011)Kr{\"a}henb{\"u}hl and Koltun]{krahenbuhl2011efficient}
Philipp Kr{\"a}henb{\"u}hl and Vladlen Koltun.
\newblock Efficient inference in fully connected crfs with gaussian edge potentials.
\newblock \emph{Advances in neural information processing systems}, 24, 2011.

\bibitem[Lan et~al.(2024{\natexlab{a}})Lan, Chen, Ke, Wang, Feng, and Zhang]{lan2024clearclip}
Mengcheng Lan, Chaofeng Chen, Yiping Ke, Xinjiang Wang, Litong Feng, and Wayne Zhang.
\newblock Clearclip: Decomposing clip representations for dense vision-language inference.
\newblock In \emph{European Conference on Computer Vision}, pp.\  143--160. Springer, 2024{\natexlab{a}}.

\bibitem[Lan et~al.(2024{\natexlab{b}})Lan, Chen, Ke, Wang, Feng, and Zhang]{lan2024proxyclip}
Mengcheng Lan, Chaofeng Chen, Yiping Ke, Xinjiang Wang, Litong Feng, and Wayne Zhang.
\newblock Proxyclip: Proxy attention improves clip for open-vocabulary segmentation.
\newblock In \emph{European Conference on Computer Vision}, pp.\  70--88. Springer, 2024{\natexlab{b}}.

\bibitem[Li et~al.(2023)Li, Wang, Duan, and Li]{li2023clip}
Yi~Li, Hualiang Wang, Yiqun Duan, and Xiaomeng Li.
\newblock Clip surgery for better explainability with enhancement in open-vocabulary tasks.
\newblock \emph{arXiv e-prints}, pp.\  arXiv--2304, 2023.

\bibitem[Li et~al.(2024)Li, Cheng, Feng, Liu, and Wang]{li2024mask}
Yongkang Li, Tianheng Cheng, Bin Feng, Wenyu Liu, and Xinggang Wang.
\newblock Mask-adapter: The devil is in the masks for open-vocabulary segmentation.
\newblock \emph{arXiv preprint arXiv:2412.04533}, 2024.

\bibitem[Liang et~al.(2023)Liang, Wu, Dai, Li, Zhao, Zhang, Zhang, Vajda, and Marculescu]{liang2023open}
Feng Liang, Bichen Wu, Xiaoliang Dai, Kunpeng Li, Yinan Zhao, Hang Zhang, Peizhao Zhang, Peter Vajda, and Diana Marculescu.
\newblock Open-vocabulary semantic segmentation with mask-adapted clip.
\newblock In \emph{Proceedings of the IEEE/CVF Conference on Computer Vision and Pattern Recognition (CVPR)}, pp.\  7061--7070, June 2023.

\bibitem[Liu et~al.(2023)Liu, Li, Wu, and Lee]{liu2023visual}
Haotian Liu, Chunyuan Li, Qingyang Wu, and Yong~Jae Lee.
\newblock Visual instruction tuning.
\newblock \emph{Advances in neural information processing systems}, 36:\penalty0 34892--34916, 2023.

\bibitem[Liu et~al.(2024)Liu, Li, Li, and Lee]{liu2024improved}
Haotian Liu, Chunyuan Li, Yuheng Li, and Yong~Jae Lee.
\newblock Improved baselines with visual instruction tuning.
\newblock In \emph{Proceedings of the IEEE/CVF Conference on Computer Vision and Pattern Recognition}, pp.\  26296--26306, 2024.

\bibitem[Luo et~al.(2022)Luo, Bao, Wu, He, and Li]{Luo2022SegCLIPPA}
Huaishao Luo, Junwei Bao, Youzheng Wu, Xiaodong He, and Tianrui Li.
\newblock Segclip: Patch aggregation with learnable centers for open-vocabulary semantic segmentation.
\newblock In \emph{International Conference on Machine Learning}, 2022.
\newblock URL \url{https://api.semanticscholar.org/CorpusID:254043520}.

\bibitem[Luo et~al.(2023{\natexlab{a}})Luo, Bao, Wu, He, and Li]{luo2023segclip}
Huaishao Luo, Junwei Bao, Youzheng Wu, Xiaodong He, and Tianrui Li.
\newblock Segclip: Patch aggregation with learnable centers for open-vocabulary semantic segmentation.
\newblock In \emph{International Conference on Machine Learning}, pp.\  23033--23044. PMLR, 2023{\natexlab{a}}.

\bibitem[Luo et~al.(2023{\natexlab{b}})Luo, Khandelwal, Sigal, and Li]{Luo2023EmergentOS}
Jiayun Luo, Siddhesh Khandelwal, Leonid Sigal, and Boyang~Albert Li.
\newblock Emergent open-vocabulary semantic segmentation from off-the-shelf vision-language models.
\newblock \emph{2024 IEEE/CVF Conference on Computer Vision and Pattern Recognition (CVPR)}, pp.\  4029--4040, 2023{\natexlab{b}}.
\newblock URL \url{https://api.semanticscholar.org/CorpusID:265498822}.

\bibitem[Mottaghi et~al.(2014)Mottaghi, Chen, Liu, Cho, Lee, Fidler, Urtasun, and Yuille]{mottaghi2014role}
Roozbeh Mottaghi, Xianjie Chen, Xiaobai Liu, Nam-Gyu Cho, Seong-Whan Lee, Sanja Fidler, Raquel Urtasun, and Alan Yuille.
\newblock The role of context for object detection and semantic segmentation in the wild.
\newblock In \emph{Proceedings of the IEEE conference on computer vision and pattern recognition}, pp.\  891--898, 2014.

\bibitem[Mukhoti et~al.(2023)Mukhoti, Lin, Poursaeed, Wang, Shah, Torr, and Lim]{mukhoti2023open}
Jishnu Mukhoti, Tsung-Yu Lin, Omid Poursaeed, Rui Wang, Ashish Shah, Philip~HS Torr, and Ser-Nam Lim.
\newblock Open vocabulary semantic segmentation with patch aligned contrastive learning.
\newblock In \emph{Proceedings of the IEEE/CVF Conference on Computer Vision and Pattern Recognition}, pp.\  19413--19423, 2023.

\bibitem[Qorbani et~al.(2025)Qorbani, Villani, Panagiotakopoulos, Colomer, Harenstam-Nielsen, Segu, Dovesi, Karlgren, Cremers, Tombari, and Poggi]{Qorbani2025SemanticLA}
Reza Qorbani, Gianluca Villani, Theodoros Panagiotakopoulos, Marc~Botet Colomer, Linus Harenstam-Nielsen, Mattia Segu, Pier~Luigi Dovesi, Jussi Karlgren, Daniel Cremers, Federico Tombari, and Matteo Poggi.
\newblock Semantic library adaptation: Lora retrieval and fusion for open-vocabulary semantic segmentation.
\newblock \emph{ArXiv}, abs/2503.21780, 2025.
\newblock URL \url{https://api.semanticscholar.org/CorpusID:277349890}.

\bibitem[Radford et~al.(2021)Radford, Kim, Hallacy, Ramesh, Goh, Agarwal, Sastry, Askell, Mishkin, Clark, et~al.]{radford2021learning}
Alec Radford, Jong~Wook Kim, Chris Hallacy, Aditya Ramesh, Gabriel Goh, Sandhini Agarwal, Girish Sastry, Amanda Askell, Pamela Mishkin, Jack Clark, et~al.
\newblock Learning transferable visual models from natural language supervision.
\newblock In \emph{International conference on machine learning}, pp.\  8748--8763. PmLR, 2021.

\bibitem[Rao et~al.(2022)Rao, Zhao, Chen, Tang, Zhu, Huang, Zhou, and Lu]{rao2022denseclip}
Yongming Rao, Wenliang Zhao, Guangyi Chen, Yansong Tang, Zheng Zhu, Guan Huang, Jie Zhou, and Jiwen Lu.
\newblock Denseclip: Language-guided dense prediction with context-aware prompting.
\newblock In \emph{Proceedings of the IEEE/CVF conference on computer vision and pattern recognition}, pp.\  18082--18091, 2022.

\bibitem[Ren et~al.(2023)Ren, Li, Xu, Zhu, Wang, Liu, Chang, and Liang]{ren2023viewco}
Pengzhen Ren, Changlin Li, Hang Xu, Yi~Zhu, Guangrun Wang, Jianzhuang Liu, Xiaojun Chang, and Xiaodan Liang.
\newblock Viewco: Discovering text-supervised segmentation masks via multi-view semantic consistency.
\newblock \emph{arXiv preprint arXiv:2302.10307}, 2023.

\bibitem[Schuhmann et~al.(2022)Schuhmann, Beaumont, Vencu, Gordon, Wightman, Cherti, Coombes, Katta, Mullis, Wortsman, et~al.]{schuhmann2022laion}
Christoph Schuhmann, Romain Beaumont, Richard Vencu, Cade Gordon, Ross Wightman, Mehdi Cherti, Theo Coombes, Aarush Katta, Clayton Mullis, Mitchell Wortsman, et~al.
\newblock Laion-5b: An open large-scale dataset for training next generation image-text models.
\newblock \emph{Advances in neural information processing systems}, 35:\penalty0 25278--25294, 2022.

\bibitem[Shao et~al.(2024)Shao, Tian, Zhao, and Su]{shao2024explore}
Tong Shao, Zhuotao Tian, Hang Zhao, and Jingyong Su.
\newblock Explore the potential of clip for training-free open vocabulary semantic segmentation.
\newblock In \emph{European Conference on Computer Vision}, pp.\  139--156. Springer, 2024.

\bibitem[Shin et~al.(2022{\natexlab{a}})Shin, Xie, and Albanie]{Shin2022ReCoRA}
Gyungin Shin, Weidi Xie, and Samuel Albanie.
\newblock Reco: Retrieve and co-segment for zero-shot transfer.
\newblock \emph{ArXiv}, abs/2206.07045, 2022{\natexlab{a}}.
\newblock URL \url{https://api.semanticscholar.org/CorpusID:249642384}.

\bibitem[Shin et~al.(2022{\natexlab{b}})Shin, Xie, and Albanie]{shin2022reco}
Gyungin Shin, Weidi Xie, and Samuel Albanie.
\newblock Reco: Retrieve and co-segment for zero-shot transfer.
\newblock \emph{Advances in Neural Information Processing Systems}, 35:\penalty0 33754--33767, 2022{\natexlab{b}}.

\bibitem[Silva-Rodriguez et~al.(2024)Silva-Rodriguez, Hajimiri, Ben~Ayed, and Dolz]{silva2024closer}
Julio Silva-Rodriguez, Sina Hajimiri, Ismail Ben~Ayed, and Jose Dolz.
\newblock A closer look at the few-shot adaptation of large vision-language models.
\newblock In \emph{Proceedings of the IEEE/CVF Conference on Computer Vision and Pattern Recognition}, pp.\  23681--23690, 2024.

\bibitem[Sun et~al.(2024)Sun, Cao, Xie, Jiang, and Pang]{sun2024cliper}
Lin Sun, Jiale Cao, Jin Xie, Xiaoheng Jiang, and Yanwei Pang.
\newblock Cliper: Hierarchically improving spatial representation of clip for open-vocabulary semantic segmentation.
\newblock \emph{arXiv preprint arXiv:2411.13836}, 2024.

\bibitem[Sung et~al.(2022)Sung, Cho, and Bansal]{sung2022vl}
Yi-Lin Sung, Jaemin Cho, and Mohit Bansal.
\newblock Vl-adapter: Parameter-efficient transfer learning for vision-and-language tasks.
\newblock In \emph{Proceedings of the IEEE/CVF conference on computer vision and pattern recognition}, pp.\  5227--5237, 2022.

\bibitem[Wang et~al.(2024)Wang, Mei, and Yuille]{wang2024sclip}
Feng Wang, Jieru Mei, and Alan Yuille.
\newblock Sclip: Rethinking self-attention for dense vision-language inference.
\newblock In \emph{European Conference on Computer Vision}, pp.\  315--332. Springer, 2024.

\bibitem[Wysocza{\'n}ska et~al.(2024)Wysocza{\'n}ska, Sim{\'e}oni, Ramamonjisoa, Bursuc, Trzci{\'n}ski, and P{\'e}rez]{wysoczanska2024clip}
Monika Wysocza{\'n}ska, Oriane Sim{\'e}oni, Micha{\"e}l Ramamonjisoa, Andrei Bursuc, Tomasz Trzci{\'n}ski, and Patrick P{\'e}rez.
\newblock Clip-dinoiser: Teaching clip a few dino tricks for open-vocabulary semantic segmentation.
\newblock In \emph{European Conference on Computer Vision}, pp.\  320--337. Springer, 2024.

\bibitem[Xing et~al.(2023)Xing, Kang, Xiao, Nie, Shao, and Lu]{xing2023rewrite}
Yun Xing, Jian Kang, Aoran Xiao, Jiahao Nie, Ling Shao, and Shijian Lu.
\newblock Rewrite caption semantics: Bridging semantic gaps for language-supervised semantic segmentation.
\newblock \emph{Advances in Neural Information Processing Systems}, 36:\penalty0 68798--68809, 2023.

\bibitem[Xu et~al.(2022{\natexlab{a}})Xu, De~Mello, Liu, Byeon, Breuel, Kautz, and Wang]{xu2022groupvit}
Jiarui Xu, Shalini De~Mello, Sifei Liu, Wonmin Byeon, Thomas Breuel, Jan Kautz, and Xiaolong Wang.
\newblock Groupvit: Semantic segmentation emerges from text supervision.
\newblock In \emph{Proceedings of the IEEE/CVF conference on computer vision and pattern recognition}, pp.\  18134--18144, 2022{\natexlab{a}}.

\bibitem[Xu et~al.(2023{\natexlab{a}})Xu, Liu, Vahdat, Byeon, Wang, and Mello]{Xu2023OpenVocabularyPS}
Jiarui Xu, Sifei Liu, Arash Vahdat, Wonmin Byeon, Xiaolong Wang, and Shalini~De Mello.
\newblock Open-vocabulary panoptic segmentation with text-to-image diffusion models.
\newblock \emph{2023 IEEE/CVF Conference on Computer Vision and Pattern Recognition (CVPR)}, pp.\  2955--2966, 2023{\natexlab{a}}.
\newblock URL \url{https://api.semanticscholar.org/CorpusID:257405338}.

\bibitem[Xu et~al.(2023{\natexlab{b}})Xu, Hou, Zhang, Feng, Wang, Qiao, and Xie]{Xu2023LearningOS}
Jilan Xu, Junlin Hou, Yuejie Zhang, Rui Feng, Yi~Wang, Yu~Qiao, and Weidi Xie.
\newblock Learning open-vocabulary semantic segmentation models from natural language supervision.
\newblock \emph{2023 IEEE/CVF Conference on Computer Vision and Pattern Recognition (CVPR)}, pp.\  2935--2944, 2023{\natexlab{b}}.
\newblock URL \url{https://api.semanticscholar.org/CorpusID:256105322}.

\bibitem[Xu et~al.(2023{\natexlab{c}})Xu, Hou, Zhang, Feng, Wang, Qiao, and Xie]{xu2023learning}
Jilan Xu, Junlin Hou, Yuejie Zhang, Rui Feng, Yi~Wang, Yu~Qiao, and Weidi Xie.
\newblock Learning open-vocabulary semantic segmentation models from natural language supervision.
\newblock In \emph{Proceedings of the IEEE/CVF conference on computer vision and pattern recognition}, pp.\  2935--2944, 2023{\natexlab{c}}.

\bibitem[Xu et~al.(2022{\natexlab{b}})Xu, Zhang, Wei, Lin, Cao, Hu, and Bai]{xu2022simple}
Mengde Xu, Zheng Zhang, Fangyun Wei, Yutong Lin, Yue Cao, Han Hu, and Xiang Bai.
\newblock A simple baseline for open-vocabulary semantic segmentation with pre-trained vision-language model.
\newblock In \emph{European Conference on Computer Vision}, pp.\  736--753. Springer, 2022{\natexlab{b}}.

\bibitem[Yang et~al.(2024)Yang, Deng, Li, and Duan]{yang2024resclip}
Yuhang Yang, Jinhong Deng, Wen Li, and Lixin Duan.
\newblock Resclip: Residual attention for training-free dense vision-language inference.
\newblock \emph{arXiv preprint arXiv:2411.15851}, 2024.

\bibitem[Yu et~al.(2023)Yu, Lu, Jin, Chen, and Wang]{yu2023task}
Tao Yu, Zhihe Lu, Xin Jin, Zhibo Chen, and Xinchao Wang.
\newblock Task residual for tuning vision-language models.
\newblock In \emph{Proceedings of the IEEE/CVF Conference on Computer Vision and Pattern Recognition}, pp.\  10899--10909, 2023.

\bibitem[Zhai et~al.(2023)Zhai, Mustafa, Kolesnikov, and Beyer]{zhai2023sigmoid}
Xiaohua Zhai, Basil Mustafa, Alexander Kolesnikov, and Lucas Beyer.
\newblock Sigmoid loss for language image pre-training.
\newblock In \emph{Proceedings of the IEEE/CVF international conference on computer vision}, pp.\  11975--11986, 2023.

\bibitem[Zhang et~al.(2024)Zhang, Liu, and Tang]{zhang2024corrclip}
Dengke Zhang, Fagui Liu, and Quan Tang.
\newblock Corrclip: Reconstructing correlations in clip with off-the-shelf foundation models for open-vocabulary semantic segmentation.
\newblock \emph{arXiv preprint arXiv:2411.10086}, 2024.

\bibitem[Zhang et~al.(2023)Zhang, Zhou, Li, He, Ma, Zhang, Yao, Zhang, and Wang]{zhang2023uncovering}
Fei Zhang, Tianfei Zhou, Boyang Li, Hao He, Chaofan Ma, Tianjiao Zhang, Jiangchao Yao, Ya~Zhang, and Yanfeng Wang.
\newblock Uncovering prototypical knowledge for weakly open-vocabulary semantic segmentation.
\newblock \emph{Advances in Neural Information Processing Systems}, 36:\penalty0 73652--73665, 2023.

\bibitem[Zhang et~al.(2021)Zhang, Fang, Zhang, Gao, Li, Dai, Qiao, and Li]{zhang2021tip}
Renrui Zhang, Rongyao Fang, Wei Zhang, Peng Gao, Kunchang Li, Jifeng Dai, Yu~Qiao, and Hongsheng Li.
\newblock Tip-adapter: Training-free clip-adapter for better vision-language modeling.
\newblock \emph{arXiv preprint arXiv:2111.03930}, 2021.

\bibitem[Zhou et~al.(2017)Zhou, Zhao, Puig, Fidler, Barriuso, and Torralba]{zhou2017scene}
Bolei Zhou, Hang Zhao, Xavier Puig, Sanja Fidler, Adela Barriuso, and Antonio Torralba.
\newblock Scene parsing through ade20k dataset.
\newblock In \emph{Proceedings of the IEEE conference on computer vision and pattern recognition}, pp.\  633--641, 2017.

\bibitem[Zhou et~al.(2022{\natexlab{a}})Zhou, Loy, and Dai]{zhou2022extract}
Chong Zhou, Chen~Change Loy, and Bo~Dai.
\newblock Extract free dense labels from clip.
\newblock In \emph{European Conference on Computer Vision}, pp.\  696--712. Springer, 2022{\natexlab{a}}.

\bibitem[Zhou et~al.(2022{\natexlab{b}})Zhou, Li, Ding, You, Qu, and Zhu]{zhou2022optimization}
Jinxin Zhou, Xiao Li, Tianyu Ding, Chong You, Qing Qu, and Zhihui Zhu.
\newblock On the optimization landscape of neural collapse under mse loss: Global optimality with unconstrained features.
\newblock In \emph{International Conference on Machine Learning}, pp.\  27179--27202. PMLR, 2022{\natexlab{b}}.

\bibitem[Zhu et~al.(2021)Zhu, Ding, Zhou, Li, You, Sulam, and Qu]{zhu2021geometric}
Zhihui Zhu, Tianyu Ding, Jinxin Zhou, Xiao Li, Chong You, Jeremias Sulam, and Qing Qu.
\newblock A geometric analysis of neural collapse with unconstrained features.
\newblock \emph{Advances in Neural Information Processing Systems}, 34:\penalty0 29820--29834, 2021.

\end{thebibliography}
\bibliographystyle{iclr2026_conference}

\newpage
\appendix
\section{Appendix}
In the appendix, we first provide additional related literature in~\Cref{subsec: literature}. We then present extended experimental details and results for both the ViT-B/16 and ViT-L/14 vision encoders in~\Cref{app:subsec-exp}. For each model, we report the distribution of average head visual discriminability scores across multiple datasets, conduct a component-wise analysis of their contributions to visual discriminability, and provide qualitative comparisons along with additional supporting results and visualizations. In~\Cref{sec:discussion}, we address key concerns, including hyperparameter selection, applicability beyond CLIP, and a comparison between SSR and direct skip methods. 

\subsection{Additional literature}\label{subsec: literature}
\paragraph{Vision-language pre-training models.} Deep learning is experiencing a significant paradigm shift driven with the emergence of large-scale vision-language models. Among these, the Contrastive Language-Image Pre-Training (CLIP) framework~\cite{radford2021learning} has achieved remarkable success, largely attributed to its strong zero-shot and few-shot generalization capabilities in visual recognition tasks, particularly image classification. Building on these strengths, an expanding body of research has sought to further improve CLIP’s zero-shot performance through enhanced training on large-scale image-text pairs~\cite{jia2021scaling, cherti2023reproducible, schuhmann2022laion, zhai2023sigmoid}, or by enabling its efficient adaptation to novel downstream tasks using limited labeled data~\cite{gao2024clip, silva2024closer, sung2022vl, yu2023task, zhang2021tip}. Nevertheless, as CLIP is pre-trained predominantly at the image level, its learned representations, particularly the [CLS] token, are optimized to capture global semantics. This coarse-grained supervision inherently limits their utility in dense prediction tasks, where fine-grained spatial localization is critical.

\paragraph{Open-vocabulary Semantic Segmentation.} Driven by the remarkable generalization capabilities of large-scale vision-language models~\cite{radford2021learning, jia2021scaling, cherti2023reproducible, schuhmann2022laion}, a growing line of research has focused on open-vocabulary semantic segmentation~\cite{ren2023viewco, liang2023open, Luo2022SegCLIPPA, Luo2023EmergentOS, Shin2022ReCoRA, Xu2023LearningOS, Xu2023OpenVocabularyPS, li2024mask, Qorbani2025SemanticLA}, which seeks to extend global cross-modal alignment to fine-grained, pixel-level predictions. Existing approaches can be broadly categorized into three groups: \emph{fully supervised}, \emph{weakly supervised}, and \emph{training-free} methods, based on the level of auxiliary supervision required for adaptation. \emph{Fully-supervised} approaches~\cite{liang2023open, Luo2023EmergentOS, li2024mask, Qorbani2025SemanticLA} adapt pretrained CLIP models to semantic segmentation by leveraging large-scale datasets with dense pixel-wise annotations for a predefined, yet limited, set of categories. Although such supervision facilitates the learning of fine-grained spatial representations, these methods often struggle to generalize to novel categories. Furthermore, their reliance on labor-intensive, densely annotated datasets poses significant challenges to scalability, limiting their applicability in real-world scenarios. Instead of requiring pixel-wise annotations, \emph{weakly-supervised} approaches  leverage auxiliary datasets with image-level annotations to adapt pre-trained CLIP models for semantic segmentation. These approaches commonly employ large-scale image-text corpora, wherein textual descriptions explicitly reference the object categories present in the corresponding images. Adaptation is achieved by enforcing cross-modal alignment through a contrastive loss~\cite{xu2022groupvit}, analogous to the original CLIP pre-training objective. Although such strategies alleviate the burden of dense supervision, they still rely on access to large-scale annotated datasets. However, the auxiliary datasets are substantially smaller than those employed during CLIP pre-training, which limits the  generalization capability. Additionally, these methods presuppose prior knowledge of the categories present in each image, thereby constraining their applicability in genuinely open-world segmentation scenarios.

\paragraph{Training-free open-vocabulary semantic segmentation.} \emph{Training-free} methods explore the feasibility of employing frozen CLIP models to generate segmentation masks in the absence of additional data for adaptation. Some approaches rely on auxiliary models pre-trained on large-scale datasets such as DINO~\cite{wysoczanska2024clip, lan2024proxyclip}, SAM~\cite{zhang2024corrclip}, or Stable Diffusion~\cite{Corradini2024FreeSegDiffTO, sun2024cliper}, which incur significant computational and memory overhead. An alternative line of research seeks to enhance CLIP's capability for dense visual representation by modifying the inference pipeline of its visual encoder. For instance, MaskCLIP~\cite{zhou2022extract} removes the self-attention module in the final Transformer layer and demonstrates that the resulting value embeddings encode local visual features that align effectively with textual prompts. CLIP-Surgery~\cite{li2023clip} introduces a dual-path structure that replaces the original self-attention with value-value attention to better preserve semantic consistency, mitigating the tendency to attend to unrelated regions. Building on this idea, GEM~\cite{bousselham2024grounding} and SCLIP~\cite{wang2024sclip} generalize the approach by incorporating correlative self-attention mechanisms, such as query-query and key-key interactions. ClearCLIP~\cite{lan2024clearclip} further simplifies the architecture by removing the final feed-forward layer and associated residual connections, which were found to contribute to noisy segmentation outputs. Additionally, NACLIP~\cite{hajimiri2025pay} imposes explicit spatial regularization, encouraging each token to primarily attend to its neighbors to improve the spatial coherence.

Taking into account both generalization ability and computational costs, this work aims to advance existing training-free methods for open-vocabulary semantic segmentation. While prior training-free approaches predominantly rely on features extracted from the final layer of the visual encoder, we instead conduct a comprehensive analysis across token, head, and layer levels. Building on this multi-level analysis, we propose recipe at each level to enhance the spatial representation capacity of CLIP, thereby improving segmentation performance within a training-free framework.

\subsection{Additional experimental results}\label{app:subsec-exp}
\subsubsection{Additional Implementation Details.} We adopt CLIP~\cite{radford2021learning} with ViT-B/16 and ViT-L/14 backbones, implemented in MMSegmentation~\cite{mmseg2020}. Following the protocol of~\cite{yang2024resclip}, input images are resized to 336 pixels on the shorter side, except for Cityscapes, which is resized to 560 pixels to accommodate higher resolution. Inference is conducted with a sliding-window strategy using $224 \times 224$ crops and a stride of 112 pixels. Consistent with TCL~\cite{cha2023learning}, we avoid computationally intensive post-processing methods that hinder fair comparison, such as PAMR~\cite{araslanov2020single} (used in TCL~\cite{cha2023learning}, NACLIP~\cite{hajimiri2025pay}) and DenseCRF~\cite{krahenbuhl2011efficient} (used in ReCo~\cite{shin2022reco}). For textual inputs, we employ the standard ImageNet prompts~\cite{radford2021learning} without additional prompting strategies. Based on the ablation study in~\Cref{sec:experiment} and~\Cref{sec:results-vit-l}, the following configuration is fixed across all dataset evaluations without task-specific tuning
\begin{itemize}[leftmargin=0.13in,topsep=0.2em,itemsep=0.11em]
    \item \textbf{ViT-B/16}: The hoyer threshold is set to $\tau = 0.5$, and the reweighting operation is applied to layers $[10, 11]$ (i.e., the two layers preceding the final one), with a reweighting coefficient of 0.1. The top-10 attention heads with the highest visual discriminability are selected using a filter threshold of $\beta = 0.7$, corresponding to the following (layer, head) pairs: (8,9), (8,8), (7,10), (9,12), (7,3), (9,4), (5,1), (9,6), (4,11), and (8,6).
    \item \textbf{ViT-L/14}: The hoyer threshold is set to $\tau = 0.4$ and the reweighting is applied to layers $[17, 23]$ with a coefficient of 0.1. Top-performing heads are selected using the same threshold $\beta = 0.7$, with the top-30 (layer, head) pairs identified as: (11, 3), (9, 3), (7, 9), (11, 6), (10, 10), (9, 13), (3, 10), (4, 14), (10, 6), (6, 9), (7, 12), (14, 16), (11, 8), (10, 13), (8, 4), (8, 8), (10, 8), (9, 4), (2, 11), (9, 6), (8, 1), (14, 1), (16, 2), (4, 13), (13, 11), (11, 14), (7, 4), (14, 11), (13, 13), and (3, 13).
\end{itemize}
All experiments are conducted using eight NVIDIA RTX A5000 GPUs, each with 24 GB of memory.

\subsubsection{Effects of individual components effects on visual discriminability}\label{sec:improve-vd}
\begin{figure}[t]
    \centering
    
    \subfloat[VOC (ViT-B)]{\includegraphics[width=0.24\textwidth]{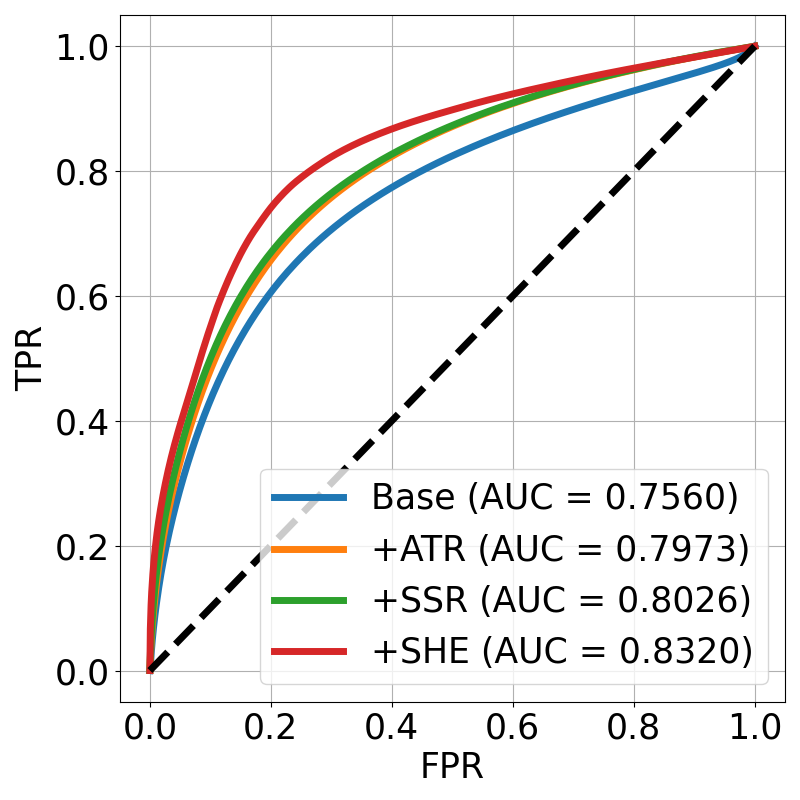}}\ 
    \subfloat[Context (ViT-B)]{\includegraphics[width=0.24\textwidth]{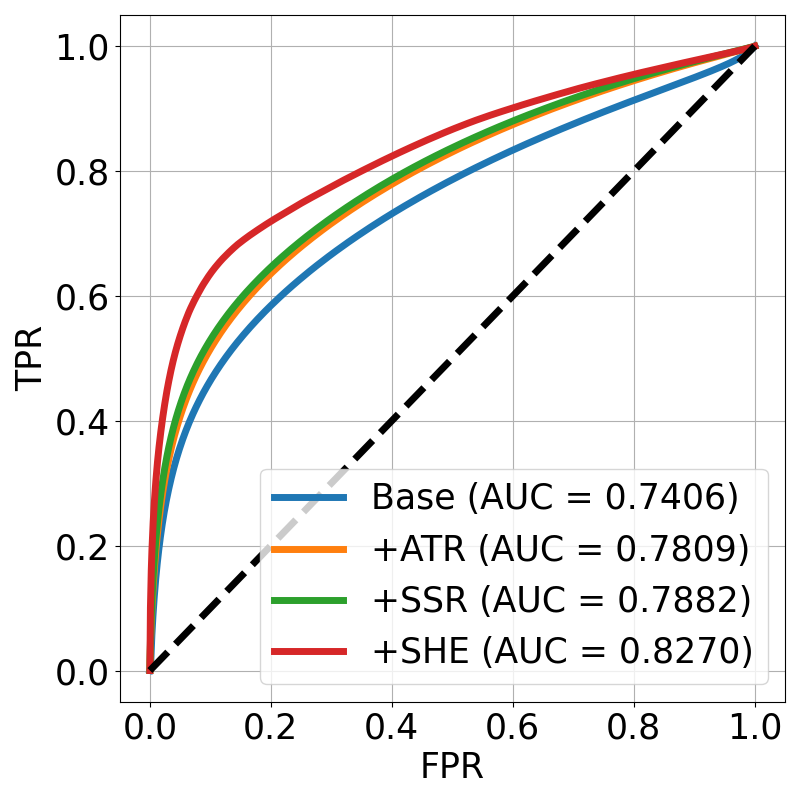}}\ 
    \subfloat[ADE (ViT-B)]{\includegraphics[width=0.24\textwidth]{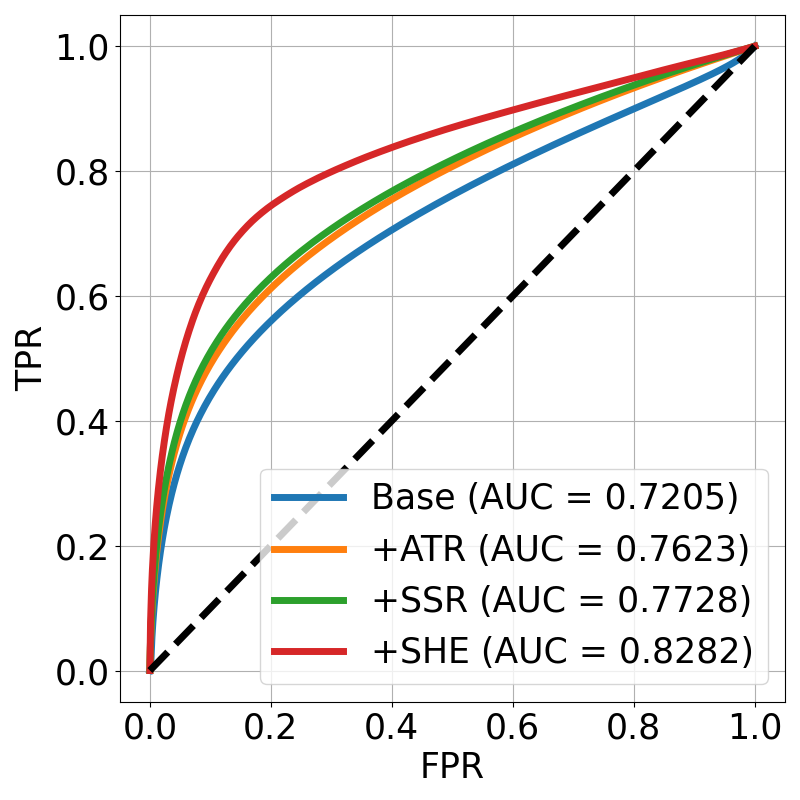}}
    \subfloat[Stuff (ViT-B)]{\includegraphics[width=0.24\textwidth]{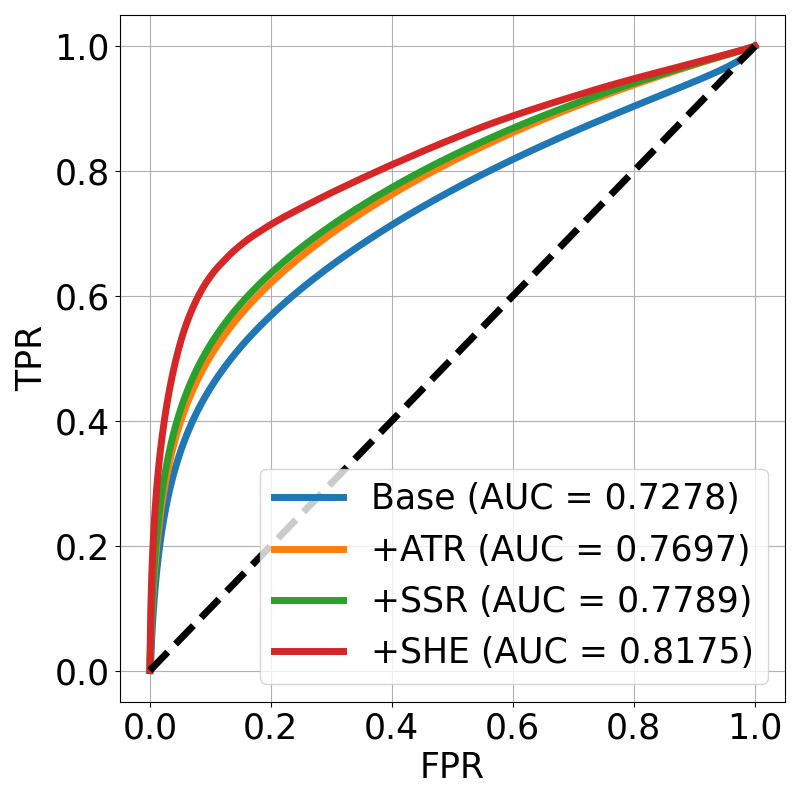}}\\

    \subfloat[VOC (ViT-L)]{\includegraphics[width=0.24\textwidth]{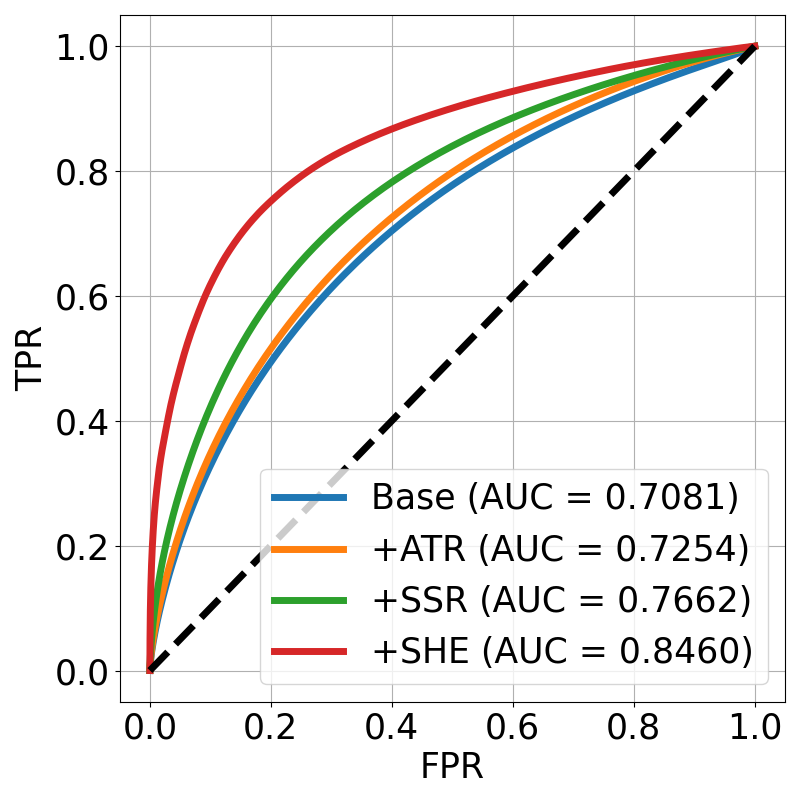}}\ 
    \subfloat[Context (ViT-L)]{\includegraphics[width=0.24\textwidth]{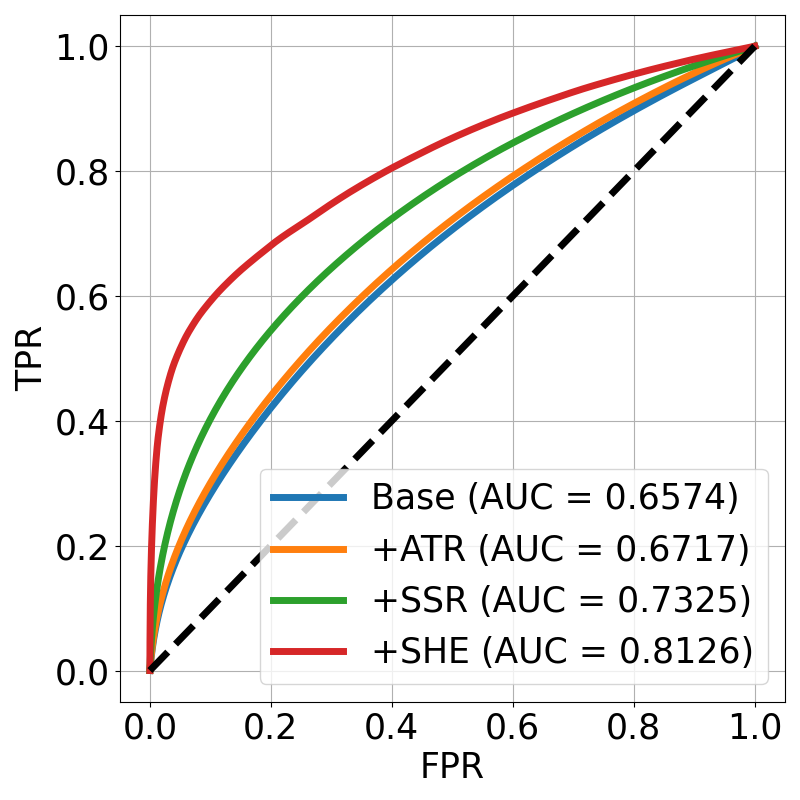}}\ 
    \subfloat[ADE (ViT-L)]{\includegraphics[width=0.24\textwidth]{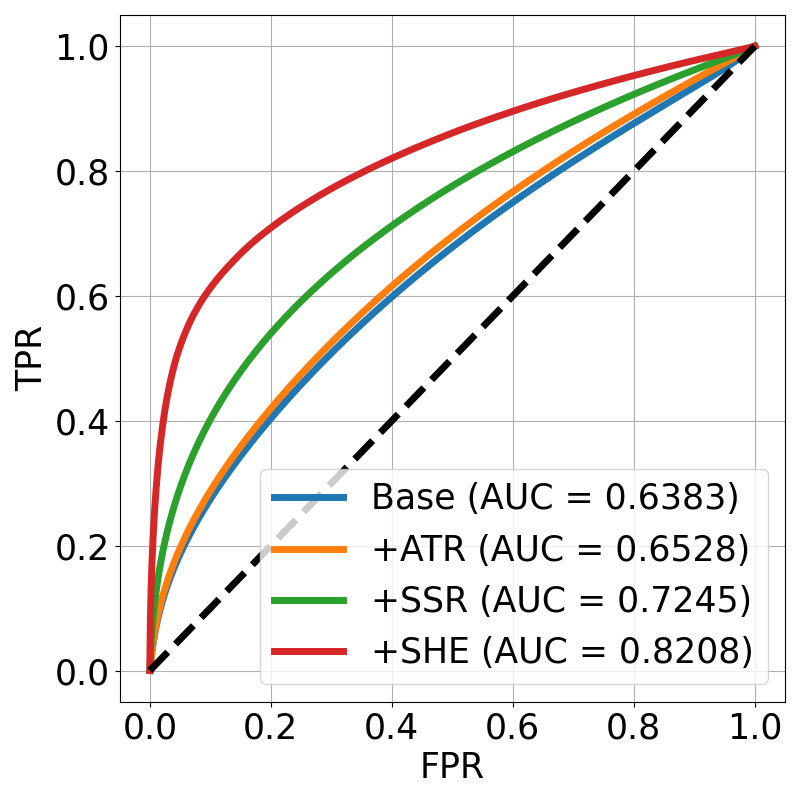}}
    \subfloat[Stuff (ViT-L)]{\includegraphics[width=0.24\textwidth]{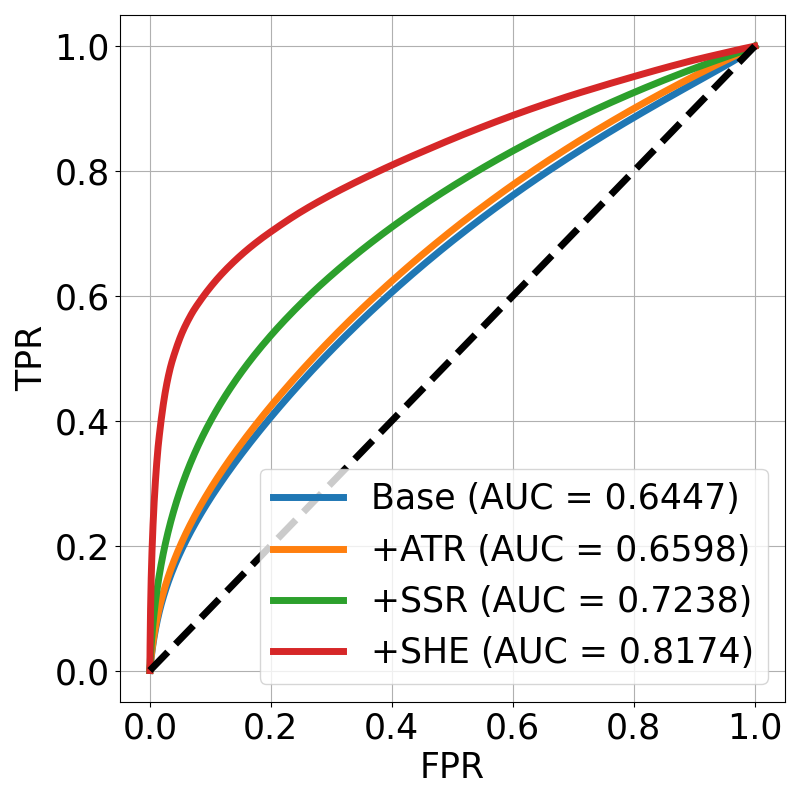}}
    \caption{ROC curves of the penultimate-layer output features for ViT-B/16 and ViT-L/14 backbones across four datasets. Each curve corresponds to an incremental combination of strategies. The results demonstrate consistent improvements in AUC with each added strategy, culminating in significantly higher visual discriminability when all are applied. }
    \vspace{-.1in}
\label{fig:last-layer-roc}\end{figure}

\begin{figure*}[t]
    \centering

    \subfloat[VOC (ViT-L)]{\includegraphics[width=0.24\textwidth]{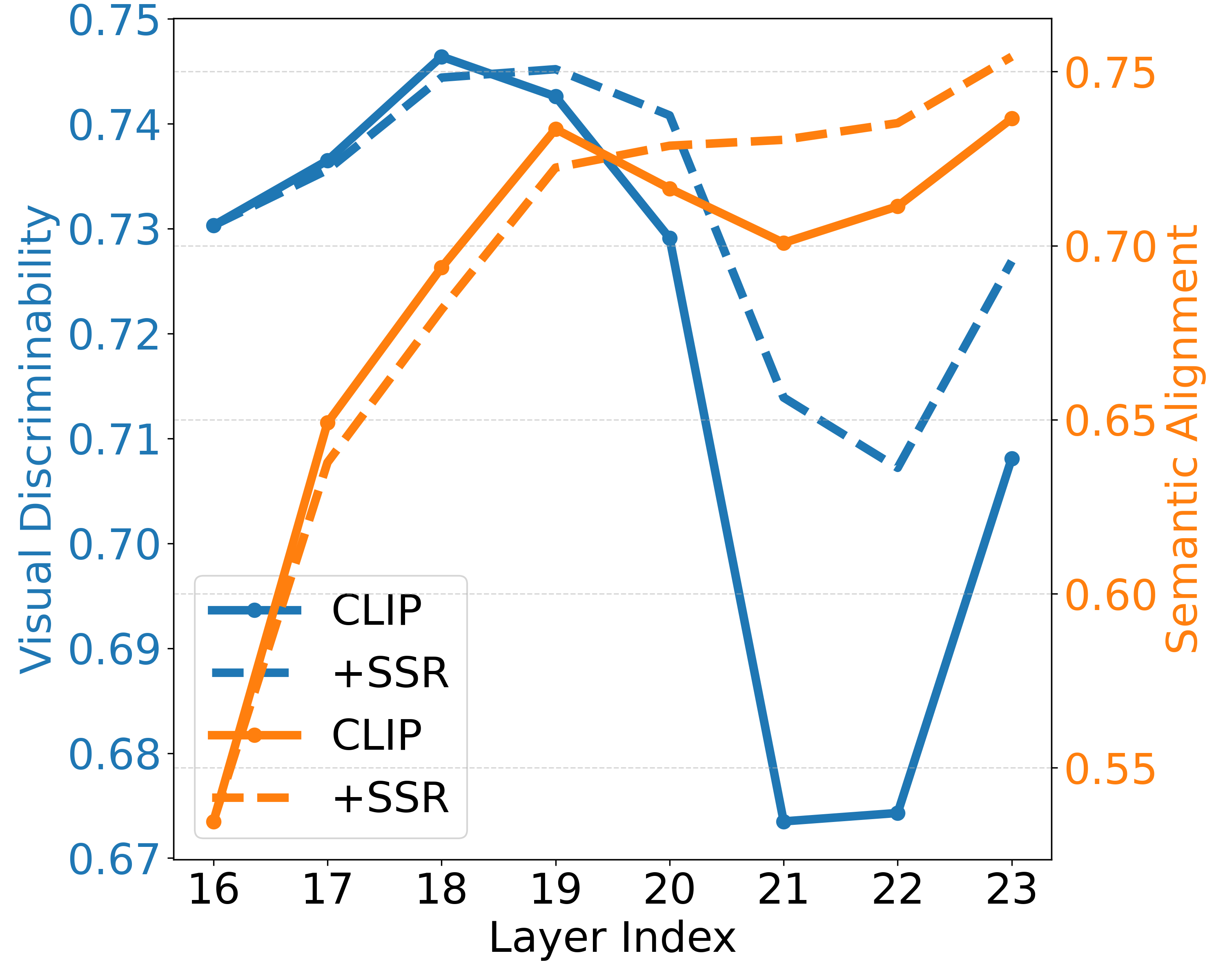}}\ 
    \subfloat[Context (ViT-L)]{\includegraphics[width=0.24\textwidth]{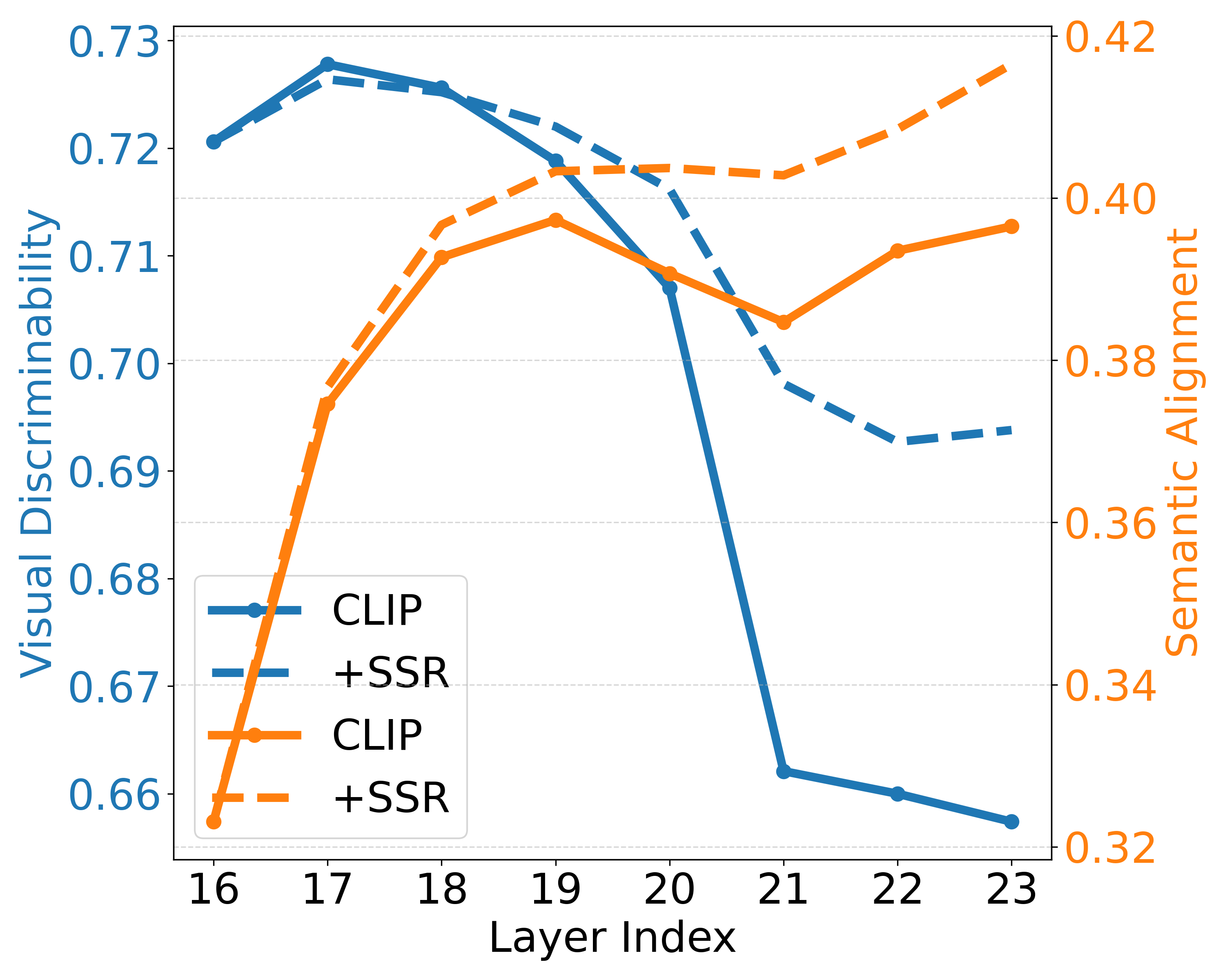}}\ 
    \subfloat[ADE (ViT-L)]{\includegraphics[width=0.24\textwidth]{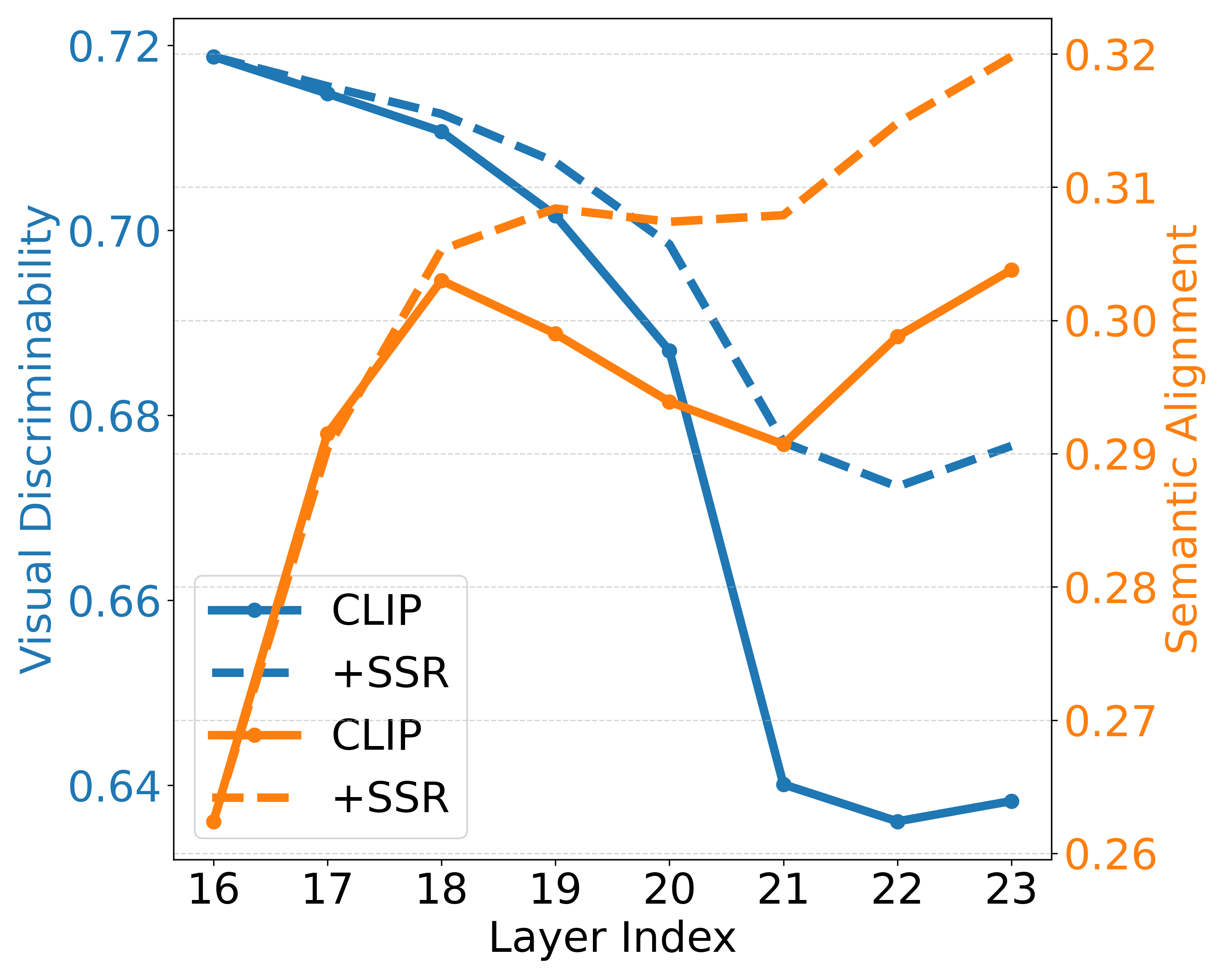}}
    \subfloat[COCO-Stuff (ViT-L)]{\includegraphics[width=0.24\textwidth]{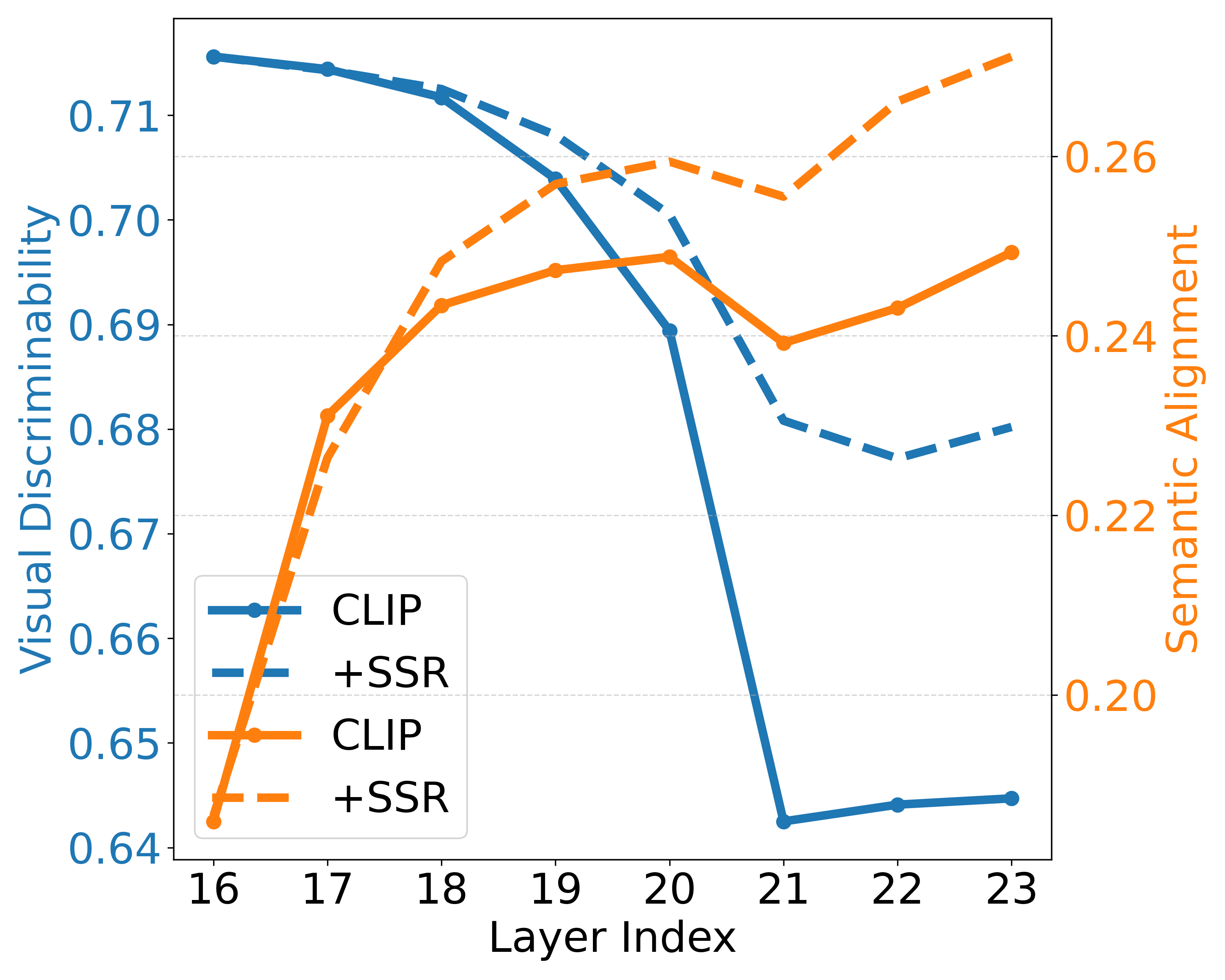}}
    
    \caption{Layer-wise visual discriminability (blue) and semantic alignment (orange) for ViT-L/14 on four benchmarks. Solid lines denote the baseline CLIP (Raw) performance, while dashed lines indicate results after applying the proposed SSR strategy. The application of SSR notably improves visual discriminability in final layers and consistently enhances semantic alignment across all datasets.}
    \label{fig:layer-auc-ssr}
\end{figure*}

\textbf{Complementary and effective for visual discriminability.} The ablation study in~\Cref{sec:experiment} confirms the effectiveness of each individual strategy for segmentation performance. As shown in~\Cref{fig:last-layer-roc}, we further analyze their contributions to visual discriminability by examining how progressively integrating these strategies improves the quality of the final representations. Similar to the definition of visual discriminability, we treat the token-wise similarity score as the output of a binary classifier predicting whether two tokens belong to the same category, and plot the ROC curves of penultimate-layer features. A higher area under the curve (AUC) indicates stronger visual discriminability. The results show that each strategy individually improves AUC, while their combination yields substantial gains in visual discriminability. For the ViT-B/16 backbone, AUC increases from 0.7560 to 0.8320 on VOC, 0.7406 to 0.8270 on Context, 0.7205 to 0.8282 on ADE, and 0.7278 to 0.8175 on COCO-Stuff. Similarly, for ViT-L/14, AUC rises from 0.7081 to 0.8460 on VOC, 0.6574 to 0.8126 on Context, 0.6383 to 0.8208 on ADE, and 0.6447 to 0.8174 on COCO-Stuff. These improvements highlight the complementary nature of the proposed strategies and their collective effectiveness in enhancing visual discriminability across diverse benchmarks.

\textbf{Effectiveness of SSR in enhancing visual discriminability of preceding layers.} Since our SSR strategy is applied not only to the penultimate layer but also to earlier layers, we further present the layer-wise curves of visual discriminability and semantic alignment after applying SSR in~\Cref{fig:layer-auc-ssr}. As illustrated in the figure, the visual discriminability of the final few layers is significantly enhanced, demonstrating the effectiveness of the proposed SSR strategy for visual discriminability improvement. Moreover, semantic alignment also exhibits consistent improvements, as improved visual discriminability reduces spurious semantic aggregation from other noisy tokens.

    

\subsubsection{Head AUC distributions}

\begin{figure}[t]
    \centering
    \subfloat[Heads 1-66]{\includegraphics[width=0.98\textwidth]{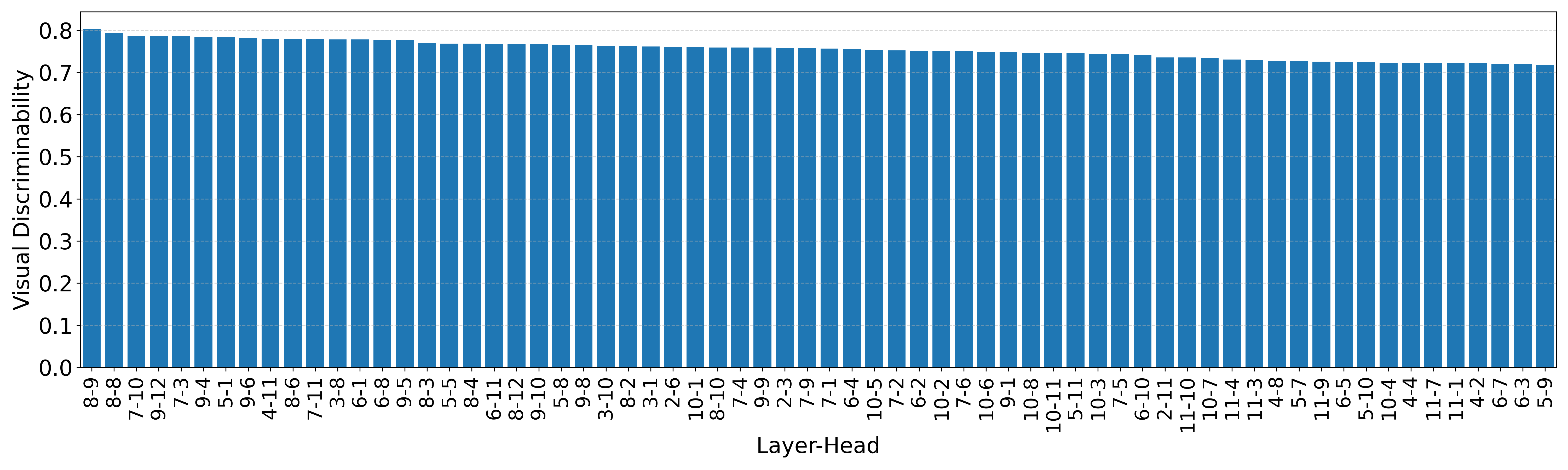}}\\ 
    \subfloat[Heads 67-172]{\includegraphics[width=0.98\textwidth]{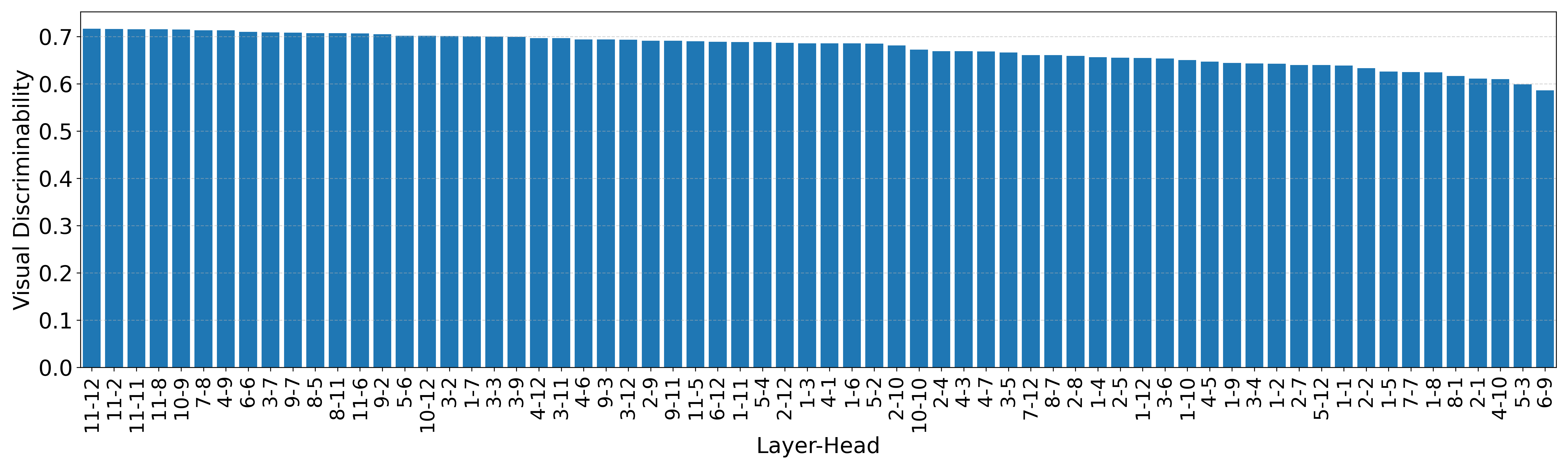}}
    \caption{Distribution of average visual discriminability score for individual attention heads across three datasets (Context, ADE, and Stuff) using the ViT-B/16 model, with the heads from final layer excluded. Bars are sorted in descending order of average visual discriminability.}
    \label{fig:head_auc_avg_b}
\end{figure}

\begin{figure}[ht]
    \centering
    
    \subfloat[Heads 1-88]{\includegraphics[width=0.98\textwidth]{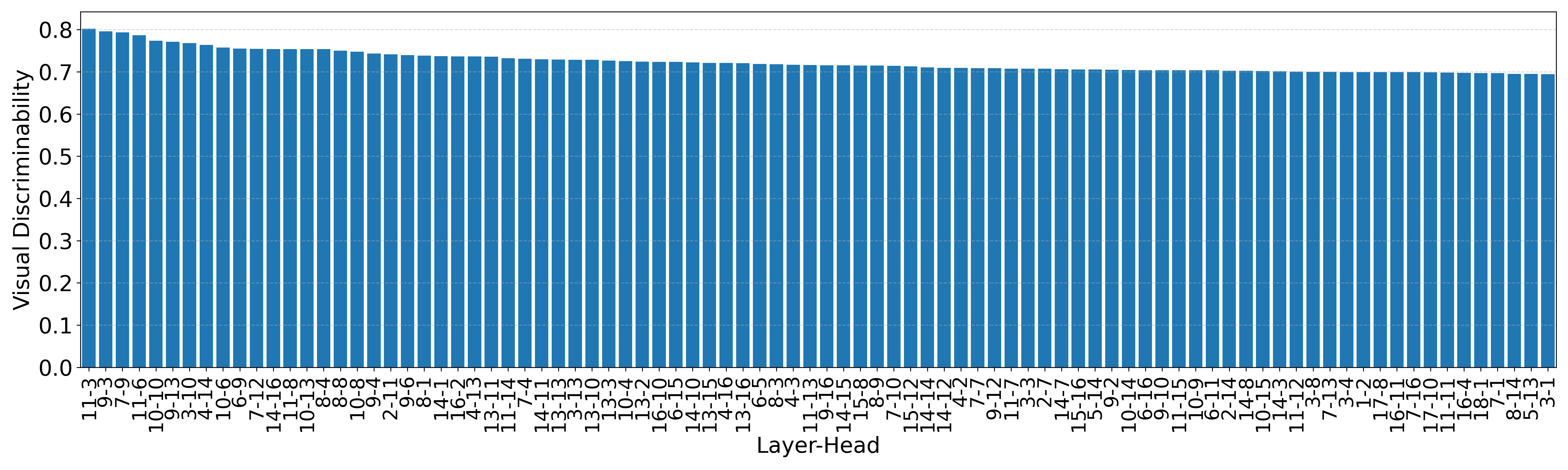}}\\ 
    \subfloat[Heads 89-176]{\includegraphics[width=0.98\textwidth]{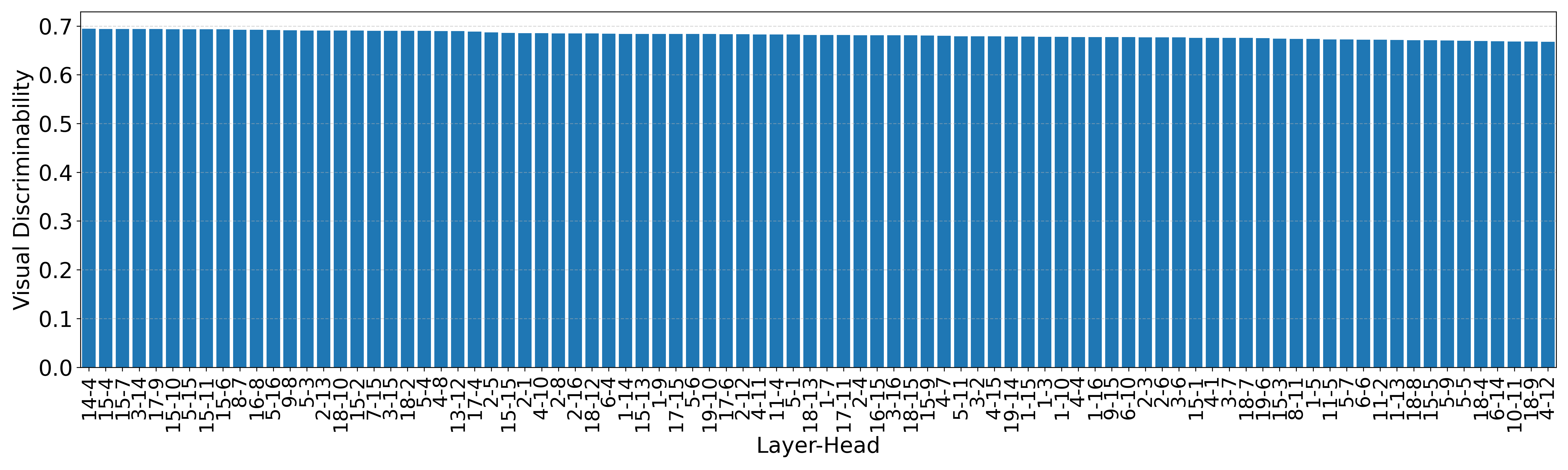}}\\
    \subfloat[Heads 177-264]{\includegraphics[width=0.98\textwidth]{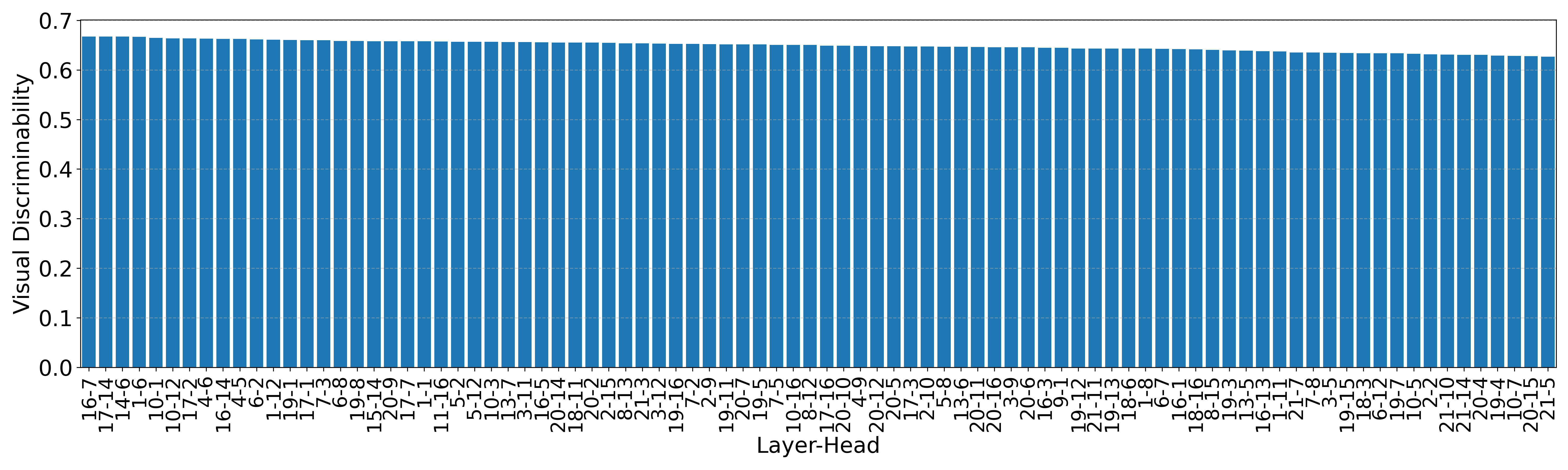}}\\
    \subfloat[Heads 265-352]{\includegraphics[width=0.98\textwidth]{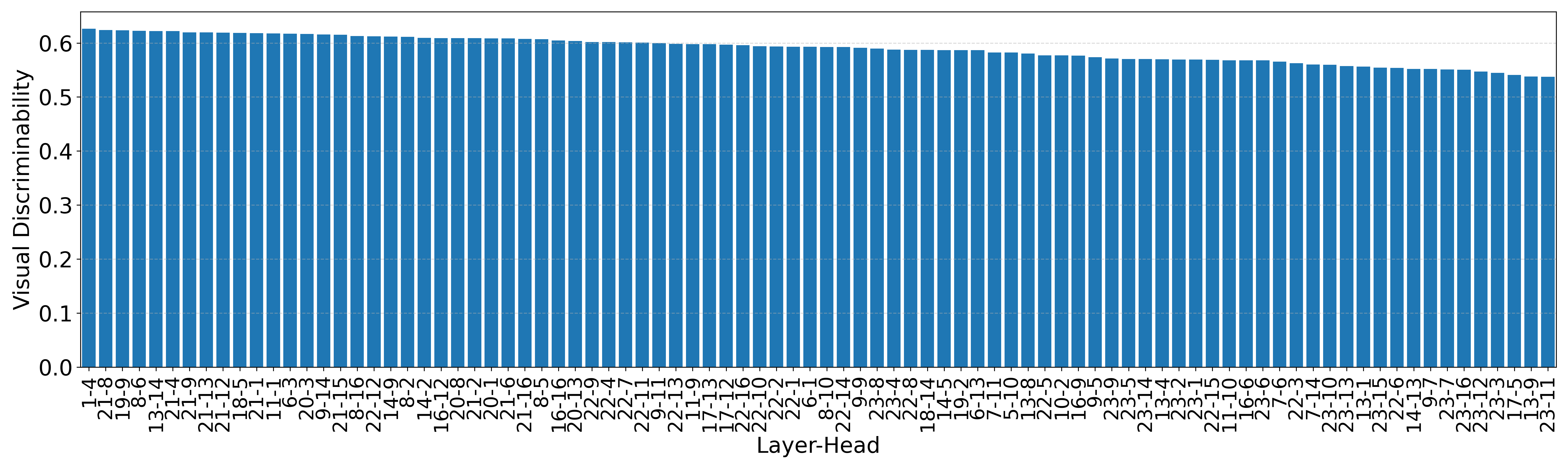}}
    \caption{Distribution of average visual discriminability score for individual attention heads across three datasets (Context, ADE, and Stuff) using the ViT-L/14 model, with the heads from final layer excluded. Bars are sorted in descending order of average visual discriminability.}
    \vspace{-.1in}
\label{fig:head_auc_avg_l}\end{figure}

In~\Cref{sec:method}, we presented the distribution of a subset of head-level visual discriminability scores across multiple datasets, employing ViT-B/16 as the visual encoder. In practice, we observed that the discriminability of head features is also degraded by the presence of abnormal tokens in the final layers. To fully realize the potential of head features, therefore, we apply abnormal token replacement to head features. Similarly, we consistently find that certain heads exhibit significantly higher visual discriminability than others. Since we select the top-$k$ heads based on their average visual discriminability across multiple datasets, we additionally present the distribution of average visual discriminability scores for all heads in descending order in~\Cref{fig:head_auc_avg_b} for ViT-B/16 and~\Cref{fig:head_auc_avg_l} for ViT-L/14, to facilitate reference and reuse in future research.

\subsubsection{Additional experiments on ViT-L/14}\label{sec:results-vit-l}
In~\Cref{tab:ovss-results-l}, we present a detailed comparison of open-vocabulary semantic segmentation models on multiple benchmarks using the ViT-L/14 backbone. LHT-CLIP consistently improves the performance of leading methods, including ClearCLIP~\cite{lan2024clearclip}, NACLIP~\cite{hajimiri2025pay}, and ResCLIP~\cite{yang2024resclip}. In particular, integrating LHT-CLIP with ResCLIP achieves state-of-the-art results. As a plug-and-play module, LHT-CLIP provides consistent gains across all datasets, highlighting its robustness and generalization. As noted in~\cite{yang2024resclip}, many methods suffer performance drops exceeding 2\% mIoU when switching from ViT-B/16 to ViT-L/14; for example, ClearCLIP declines by 2.7\%. In contrast, LHT-CLIP largely mitigates this degradation, demonstrating strong cross-backbone robustness. These results further validate its effectiveness in enhancing open-vocabulary segmentation across both ViT-B/16 and ViT-L/14 architectures.

Additionally, in~\Cref{tab:hoyer-thres-l}, \Cref{tab:reweight-l}, \Cref{tab:k-heads-l}, and~\Cref{tab:ablation-l}, we present detailed analyses of the hoyer threshold parameter, the spatial-semantic reweighting configuration, the number of selected attention heads, and the ablation analysis of individual components, respectively, for the ViT-L/14 model. Consistent with the ViT-B/16 results, the optimal configuration is achieved with a moderate sparsity level ($\tau=0.4$), a reweighting coefficient $\alpha=0.1$ applied to layers 17–23, and the top-$k$ heads with $k=30$. Therefore, we adopt these settings as fixed hyperparameters for ViT-L/14 across all benchmarks, without any dataset-specific tuning. Moreover, the ablation study confirms the effectiveness of each individual component and their complementary effect. As shown in~\Cref{tab:ablation-l}, their combination leads to a substantial improvement of 5.1 mIoU, achieving a final score of 30.1 mIoU across the four benchmark datasets.
\begin{table*}[t]
\centering
\caption{Performance comparison of our approach with other methods on eight semantic segmentation benchmarks following the evaluation protocol in \Cref{sec:experiment}. Our results are marked in \graycell{gray}.}
\label{tab:ovss-results-l}
\footnotesize
\tabcolsep=0.22em
\begin{tabular}{llccccccccc}
\toprule
\multirow{2}{*}{\textbf{Methods}} & \multirow{2}{*}{\textbf{Training}} & \multicolumn{3}{c}{With a background class} & \multicolumn{5}{c}{Without background class} & \multirow{2}{*}{Avg.} \\ \cmidrule(lr){3-5}\cmidrule(lr){6-10}
                        &                                & VOC21       & C60       & Object      & VOC20   & City  & C59  & ADE  & Stuff  &                       \\
 \midrule
CLIP                    & \multicolumn{1}{c}{\ding{55}}           & 9.8        & 4.2             & 3.9         & 18.8    & 4.0  & 5.6       & 2.1 & 2.8   & 6.4\\
MaskCLIP & \multicolumn{1}{c}{\ding{55}} & 24.4 & 10.0 & 9.9 & 30.0 & 11.8 &12.6 & 7.8 & 10.1 & 14.6 \\     
CLIPSurgery & \multicolumn{1}{c}{\ding{55}} & 47.7 & 27.2 & 27.5 & 80.5 & 32.1 &31.5 & 16.4 & 17.3 & 35.0 \\ 
SCLIP & \multicolumn{1}{c}{\ding{55}} &44.4 & 22.3 & 24.9 & 60.3 &32.2 &20.5 &7.1 & 13.1 & 28.1\\
\midrule
ClearCLIP &  \multicolumn{1}{c}{\ding{55}} &48.7 & 28.3 & 29.7 & 80.0 &27.9 &29.6 & 15.0 & 19.9 & 34.9\\
\rowcolor{gray!20}
+LHT-CLIP (ours)  & \multicolumn{1}{c}{\ding{55}} &62.8 & 33.9 & 33.5 & 85.9 &40.5 &39.0&20.3 & 26.1 &  42.7 \green{(+7.8)}\\
NACLIP~\cite{hajimiri2025pay} & \multicolumn{1}{c}{\ding{55}} &52.2 & 28.7 & 30.4 & 78.7 &31.4 & 32.1 &17.3 & 21.4 & 36.5\\
\rowcolor{gray!20}
+LHT-CLIP (ours)  & \multicolumn{1}{c}{\ding{55}} &63.4 & 34.0 & 33.4 & 84.6 &40.8 &39.0 & 20.5 & 25.9&42.7 \green{(+6.2)}\\
ResCLIP~\cite{yang2024resclip} & \multicolumn{1}{c}{\ding{55}} &55.7 & 29.7 & 30.8 & 81.3 &32.8 & 33.6 &17.9 & 22.9 & 38.1\\
\rowcolor{gray!20}
+LHT-CLIP (ours)  & \multicolumn{1}{c}{\ding{55}} &63.4 & 34.0 & 33.4 & 85.9 &40.9 &39.0&20.5 & 26.0 & 42.9 \green{(+4.8)}\\
\bottomrule 
\end{tabular}
\vspace{-.1in}
\end{table*}

\begin{table}[t]  
\noindent
\begin{minipage}{0.48\linewidth}
\centering
\footnotesize
\captionof{table}{Study of hoyer sparsity threshold $\tau$.}
\label{tab:hoyer-thres-l}
\begin{tabular}{cccccc}
\toprule
$\tau$ & C60 & Stuff & C59 & ADE & Avg \\
\midrule
$\tau = 0.2$ & 2.4 & 2.8 & 3.8 & 2.3 & 2.8\\
\rowcolor{gray!20}
$\tau = 0.4$ & 29.7 & 22.5 & 33.6 & 17.8 & 25.9 \\
$\tau = 0.5$ &   29.2 & 22.2 & 33.3 & 17.6 & 25.6  \\
$\tau = 0.8$ &   28.9 & 22.0 & 33.0 & 17.4 & 25.3   \\
$\tau = 0.9$ &   28.9 & 22.0 & 33.0 & 17.4 & 25.3   \\
baseline & 28.6 & 21.7 & 32.5 & 17.3 & 25.0  \\
\bottomrule
\end{tabular}
\end{minipage}%
\hfill
\begin{minipage}{0.51\linewidth}
\centering
\footnotesize
\captionof{table}{Study of $(l_{\text{start}}, l_{\text{end}}, \alpha)$ in SSR module.}
\label{tab:reweight-l}
\begin{tabular}{cccccc}
\toprule
$(l_{\text{start}}, l_{\text{end}}, \alpha)$ & C60 & Stuff & C59 & ADE & Avg \\
\midrule
baseline & 28.6 & 21.7 & 32.5 & 17.3 & 25.0 \\
(14, 23, 0.1) & 30.8 & 21.2 & 33.9 & 16.9 & 25.7 \\
\rowcolor{gray!20}
(17, 23, 0.1)  & 31.4 & 23.1 & 34.7 & 17.8 & 26.8 \\
(19, 23, 0.1)  & 30.9 & 23.0 & 34.2 & 17.9 & 26.5 \\
(17, 23, 0.05) & 30.3 & 22.5 & 33.6 & 17.5 & 26.0 \\
(17, 23, 0.2) & 31.6 & 22.1 & 34.8 & 17.8 & 26.6 \\
\bottomrule
\end{tabular}
\end{minipage}
\end{table}

\begin{table}[t]  
\noindent
\begin{minipage}{0.50\linewidth}
\centering
\footnotesize
\captionof{table}{Study of number of selected heads $k$.}
\label{tab:k-heads-l}
\begin{tabular}{cccccc}
\toprule
$k$ & C60 & Stuff & C59 & ADE & Avg \\
\midrule
baseline & 28.6 & 21.7 & 32.5 & 17.3 & 25.0
\\
$k=1$ &  32.0 & 24.3 & 35.0 & 18.6 & 27.5 \\
$k=10$ &  32.8 & 24.6 &  36.0 & 19.2 & 28.2 \\
\rowcolor{gray!20}
$k=30$ & 33.1 & 24.9 & 36.3 & 19.5 & 28.5 \\
\rowcolor{gray!20}
$k=60$ & 33.1 & 24.9 & 36.3 & 19.5 & 28.5 \\
$k=100$ & 33.0 & 24.8 & 36.4 & 19.5 & 28.4 \\
\bottomrule
\end{tabular}
\end{minipage}%
\hfill
\begin{minipage}{0.48\linewidth}
\centering
\footnotesize
\captionof{table}{Combination of three strategies.}
\label{tab:ablation-l}
\begin{tabular}{lccccc}
\toprule
\textbf{Methods} & \multicolumn{3}{c}{\textbf{Module}} & \textbf{mIoU} & $\Delta$ \\
\cmidrule(lr){2-4}
 & \textit{ATR} & \textit{SSR} &\textit{SHE} & & \\
\midrule
baseline & -- & -- & -- & 25.0 & -- \\
 & \checkmark & \checkmark & -- & 28.1 & \textcolor{green!60!black}{+3.1} \\
 & -- & \checkmark & \checkmark & 29.2 & \textcolor{green!60!black}{+4.2} \\
 & \checkmark & -- & \checkmark & 29.0 & \textcolor{green!60!black}{+4.0} \\
\rowcolor{gray!10} 
Ours & \checkmark & \checkmark& \checkmark & \textbf{30.1} & \textcolor{green!60!black}{\textbf{+5.1}} \\
\bottomrule
\end{tabular}
\end{minipage}
\end{table}

\begin{table}[t]
\centering
\caption{Efficiency comparison of individual strategies.}
\label{tab:efficiency}
\begin{tabular}{lccc}
\toprule
\textbf{Models} & \textbf{FLOPs(G) ↓} & \textbf{Params(M) ↓} & \textbf{Speed(FPS) ↑} \\
\midrule
CLIP       & 106.10 & 149.6 & 13.7 \\
ResCLIP  & 141.34 & 149.6 & 3.0 \\
\midrule
 baseline       & 100.70 & 149.6 & 13.9 \\
 +ATR           & 100.88 & 149.6 & 12.9 \\
 +SSR       & 100.88 & 149.6 & 11.7 \\
 +SHE    & 102.65 & 149.6 & 8.2 \\
\bottomrule
\end{tabular}
\end{table}

\subsubsection{Qualitative results}
\begin{figure}[b]
    \centering
    \includegraphics[width=0.98\linewidth]{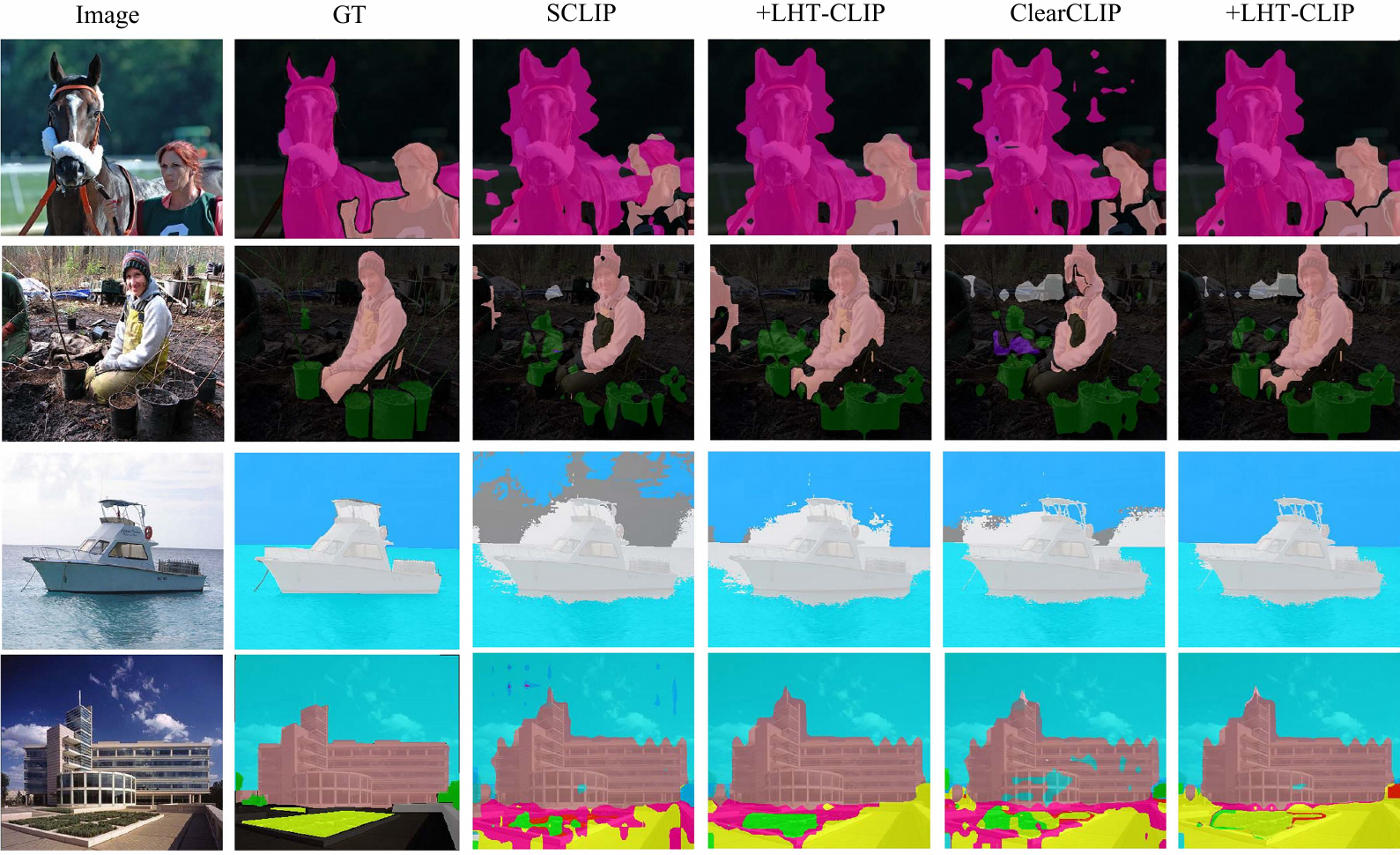}
    \caption{Qualitative comparison between CLIP-based training-free baseline methods and their counterparts integrated with LHT-CLIP.}
    \label{fig:vis}
\end{figure}
In~\Cref{fig:vis}, we present a qualitative comparison between CLIP-based training-free baseline methods and their counterparts integrated with LHT-CLIP. As shown in the figure, the integration of LHT-CLIP consistently leads to more accurate and visually coherent segmentation results. The enhanced models produce cleaner segmentation maps with reduced noise, improved spatial consistency among same-category objects, and better delineation of object boundaries, all while preserving the original text-image alignment capabilities of the baseline methods.

\subsubsection{Efficiency Comparison}
In~\Cref{tab:efficiency}, we report efficiency results on the Context59 dataset using an A5000 GPU with the ViT-B/16 backbone, taking ClearCLIP as the baseline. We further analyze the computational overhead introduced by each strategy. Compared to the prior state-of-the-art ResCLIP, our method removes the costly PAMR post-processing step in its Semantic Feedback Refinement module while simultaneously improving segmentation performance. Consequently, inference speed increases from 3.0 to 8.2 FPS, and computational cost decreases from 141.34G to 102.65G FLOPs.

\begin{table}[b]  
\centering
\footnotesize
\captionof{table}{Performance of the ViT-B variant as the SigLIP vision encoder. The best results are obtained when combined with our proposed LHT-CLIP method, as highlighted in \graybox{gray}.}
\label{tab:siglip}
\begin{tabular}{ccccccc}
\toprule
Method & VOC20 & City & C59 & ADE & Stuff & Avg \\
\midrule
SigLIP & 48.0 & 20.5 & 18.6 & 11.5 & 12.1 & 22.1\\
SigLIP+ClearCLIP & 5.7 & 2.9 & 1.7 & 0.6 & 1.6 & 2.3 \\
\rowcolor{gray!20}
SigLIP+LHT-CLIP & 59.1 & 23.2 & 23.1 & 13.8 & 15.3& 26.9 \\
\bottomrule
\end{tabular}
\hfill
\end{table}
\subsection{Discussion}\label{sec:discussion}
\subsubsection{Hyperparameter Selection}
While LHT-CLIP introduces several hyperparameters, these are guided by universal patterns consistently observed across datasets, such as the sparsity of abnormal tokens, the trade-off between visual discriminability and semantic alignment, and the presence of shared visually discriminative heads. Building on these consistent phenomena, we emphasize three key points:

\begin{itemize}[leftmargin=0.13in,topsep=0.2em,itemsep=0.11em]
    \item [1.]\textbf{Generalizable.} The hyperparameters are generalizable across datasets. In practice, we found that values selected on a small subset of training data (e.g., 1,000 randomly selected training samples from \{Context, ADE, COCO-Stuff\}) also perform well on other datasets, indicating that they are not overfitted to specific datasets. As shown in~\Cref{sec:results-vit-b} and~\Cref{sec:results-vit-l}, we apply the same hyperparameters to all datasets without specific tuning, underscoring their strong generalizability.
     \item [2.]\textbf{Robustness.} The method is robust to a wide range of hyperparameter values, which means that precise tuning is not required to achieve strong performance. This is demonstrated in our ablation studies~\Cref{sec:ablation-b} and~\Cref{sec:results-vit-l}, where the results remain consistent in various hyperparameter settings.
     \item [3.]\textbf{Reliable initialization.} Additionally, our analysis provides interpretable heuristics for selecting hyperparameters. For example, in the case of an SSR starting layer, one can use the observed turning point in visual discriminability or semantic alignment as a reliable initialization.
\end{itemize}
Taken together, while several hyperparameters are introduced, we believe these properties support the practical usability of our method without requiring extensive or fragile hyperparameter tuning.

\subsubsection{Applicability beyond CLIP}
To evaluate the generalizability of our approach beyond the CLIP model, we further assessed it on the SigLIP model~\cite{zhai2023sigmoid}. In particular, we employed the ViT-B variant as the SigLIP vision encoder and conducted experiments on five datasets without a background category, as summarized in~\Cref{tab:siglip}. Consistent with our findings for CLIP, the vanilla SigLIP model performs poorly on these datasets, yielding an average score of only 22.1. Interestingly, we observed that prior work such as ClearCLIP~\cite{lan2024clearclip}, which removes the FFN and residual connections while altering the attention mechanism in the final transformer layer, performs even worse than the vanilla SigLIP model. We believe that this degradation arises from architectural differences between CLIP and SigLIP, particularly the replacement of CLIP’s linear projector with the AttentionPool projector in SigLIP. In this setting, modifying the final transformer layer, as done in ClearCLIP, may substantially disrupt the input distribution to the AttentionPool projector, thereby hindering performance. For this reason, we adopt the vanilla SigLIP model as our baseline.

In contrast to prior approaches that focus exclusively on modifying the final transformer layer, our method is guided by a fine-grained analysis of visual discriminability and semantic alignment across token, head, and layer levels. Rather than restricting improvements to the last stage, we introduce targeted modifications at earlier layers of the model, leading to more robust and generalizable inference. As shown in~\Cref{tab:siglip}, our approach delivers a notable gain of 4.8 points in average segmentation performance, improving from 22.1 to 26.9.

\subsubsection{Comparison between SSR and direct skip methods}
\begin{table}[t]  
\centering
\footnotesize
\captionof{table}{Comparison between SSR and direct skip methods using the ClearCLIP as the baseline method and ViT-L/14 as the CLIP vision encoder. The best results are obtained when combined with our proposed LHT-CLIP method, as highlighted in \graybox{gray}.}
\label{tab:skip}
\begin{tabular}{cccccccccc}
\toprule
Method & VOC21 & C60 & Obj &VOC20& City& C59 & ADE & Stuff & Avg \\
\midrule
ClearCLIP & 48.7 & 28.3 & 29.7 & 80.0 &27.9 &29.6 & 15.0 & 19.9 & 34.9\\
ClearCLIP+direct-skip & 60.3 & 33.7 & 32.1 & 77.3 & 40.8 & 38.8 & 20.1 & 23.7 & 40.8 \\
\rowcolor{gray!20}
ClearCLIP+LHT-CLIP & 62.8 & 33.9 & 33.5 & 85.9 &40.5 &39.0&20.3 & 26.1 &  42.7 \\
\bottomrule
\end{tabular}
\hfill
\end{table}
Recent work on the Perception Encoder~\cite{bolya2025perception} highlights the effectiveness of directly selecting an earlier layer’s output for segmentation. This approach is conceptually related to our spatial–semantic reweighting (SSR) strategy, which emphasizes residual pathways by amplifying their contribution to counteract the dominance of deeper layers. In~\Cref{tab:skip}, we present a systematic comparison between our SSR method and the direct-skip strategy, using ViT-L/14 as the CLIP vision encoder and adopting ClearCLIP as the baseline.

Although directly leveraging earlier layer outputs provides some improvement over baseline CLIP performance, our experiments show that SSR consistently achieves superior results. For example, when augmented with ATR and SHE, directly skipping layers 20–23 of the ViT-L encoder and feeding the output of layer 19 into layer 24 yields an average performance of 40.8, higher than the baseline ClearCLIP score of 34.9, but still below the performance of our full LHT-CLIP method (42.7), as reported in~\Cref{tab:skip}.

\subsubsection{Comparison of Norm-Based and Sparsity-Based Criteria for ATR}\label{sec:norm-sparse-compare}
We regard ATR as a flexible framework that functions by detecting and replacing emergent abnormal tokens. As discussed in~\Cref{sec:analysis}, these tokens exhibit two distinctive characteristics: unusually large activation magnitudes and sparse activation patterns. To evaluate the effectiveness of a magnitude-based criterion for abnormal token detection, we conduct additional experiments using ViT-B/16 as the vision encoder, with ClearCLIP as the baseline method.

The norm-based approach identifies abnormal patches by computing the activation norm of each visual patch, classifying those above a threshold $\gamma$ as abnormal. Empirically, we find that this method achieves its best average performance of 27.9 across the four datasets used in~\Cref{tab:norm-thres} when the threshold is set to $\gamma=14$, which is comparable to the performance of our sparsity-based ATR method (28.0). However, the norm-based method is more sensitive to threshold selection. For example, the overall performance drops to 26.8 when $\gamma=12$ and 27.6 when $\gamma=16$, whereas the sparsity-based approach remains more stable under hyperparameter variations in~\Cref{tab:hoyer-thres}.
\begin{table}[t]  
\centering
\footnotesize
\captionof{table}{Study of norm threshold $\gamma$. The best results and corresponding $\gamma$ are highlighted in \graybox{gray}.}
\label{tab:norm-thres}
\begin{tabular}{cccccccccc}
\toprule
$\tau$ & C60 & Stuff & C59 & ADE & Avg \\
\midrule
$\gamma = 12$ & 31.7 & 23.0 & 35.7 & 16.9 & 26.8\\
$\gamma = 13$ & 32.7 & 23.9 & 36.6 & 17.5 & 27.7 \\
\rowcolor{gray!20}
$\gamma = 14$ &   32.8 & 24.2 & 36.8 & 17.8 & 27.9  \\
$\gamma = 15$ &   32.6 & 24.3 & 36.7 & 17.8 & 27.8   \\
$\gamma = 16$ &   32.2 & 24.1 & 36.6 & 17.6 & 27.6   \\
baseline & 32.4 & 24.0 & 36.0 & 17.6 & 27.5  \\
\bottomrule
\end{tabular}
\hfill
\end{table}
\end{document}